\documentclass[10pt,twocolumn,letterpaper]{article}

\usepackage[pagenumbers]{wacv} 

\usepackage{multirow}

\definecolor{wacvblue}{rgb}{0.21,0.49,0.74}
\usepackage[pagebackref,breaklinks,colorlinks,allcolors=wacvblue]{hyperref}

\def\wacvPaperID{1346} 
\def\confName{WACV}
\def\confYear{2027}

\title{EarthScape: A Multimodal Benchmark for Geologic Mapping and Surface-Aware Learning under Geographic Domain Shift}

\author{
    Matthew A. Massey\\
    {\tt\small matthew.massey@uky.edu}\\
    \and
    Nusrat Munia \\
    {\tt\small Nusrat.Munia@uky.edu} \\
    \and
    Abdullah-Al-Zubaer Imran\\
    {\tt\small aimran@uky.edu} \\ 
    \and
    University of Kentucky, Lexington, KY, USA
}

\begin{document}
\maketitle
\begin{abstract}
Understanding natural Earth surface processes requires integrating topography, imagery, hydrology, and human infrastructure while remaining robust to geographic variation, yet few benchmarks support both challenges. We introduce EarthScape, a multimodal benchmark for geologic mapping and surface-aware learning under geographic domain shift. EarthScape combines aerial imagery, LiDAR-derived elevation, multi-scale terrain derivatives, and contextual vector data into 37 co-registered input channels per sample, together with pixel-level geologic masks. Its geographically separated regions share a common label space, enabling controlled evaluation of in-domain and cross-region generalization under realistic class imbalance and spatial complexity. Benchmark experiments show that engineered terrain representations transfer more robustly than raw elevation, while spatial support and multimodal terrain inputs strongly influence performance. Fusion effectiveness depends on backbone and evaluation domain, with more complex cross-attention providing no consistent advantage over simpler strategies. These representation trends extend across geospatial foundation models and semantic segmentation, while class-level results show that no single terrain feature or scale is optimal across geologic classes. Together, EarthScape provides a benchmark for studying representation design, multimodal fusion, and geographic robustness in surface-aware geospatial learning.
Our source code is available at \href{https://github.com/masseygeo/earthscape}{https://github.com/masseygeo/earthscape} and the dataset at \href{https://doi.org/10.13023/kgs.data.05.01.2025}{https://doi.org/10.13023/kgs.data.05.01.2025}.
\end{abstract}
    
\section{Introduction}

Vision models for Earth observation are increasingly expected to operate beyond the regions in which they are trained, yet performance often degrades under geographic domain shift \citep{marsocci2023geomultitasknet, al2025analysing, ekim2025distribution}. This issue is particularly relevant for tasks that depend on physical surface structure, where predictions may rely on terrain morphology, hydrologic context, and human modification, rather than spectral appearance alone \citep{wu2025integrating, ruiz2023deep, florinsky2016digital}. Although multiple geospatial modalities are often available \citep{li2022deep}, it remains unclear which representations best transfer across regions, how derived features compare with raw sensor inputs, and whether multimodal fusion actually improves geographic robustness.

Existing remote-sensing benchmarks have driven progress in land-cover mapping, object detection, and urban scene understanding, but many emphasize land-cover or land-use categories rather than physical surface structure \citep{van2018spacenet, lam2018xview, christie2018functional, sumbul2019bigearthnet, demir2018deepglobe, schmitt2019sen12ms}. As a result, they are not designed to systematically isolate how topography, representation choice, and geographic variation interact in multimodal learning. A benchmark for this setting should provide aligned heterogeneous inputs, a shared label space across regions, and evaluation protocols that separate in-domain performance from cross-region robustness.

Surficial geologic mapping provides a natural stress test for these questions. Surficial geologic classes correspond to surface materials and landforms shaped by erosion, transport, deposition, weathering, and human modification \citep{Compton1985, Lisle2011} (Fig. \ref{fig:dataset}). Their visual appearance may vary across regions, while their spatial organization is strongly tied to terrain shape, drainage structure, and scale-dependent geomorphic patterns \citep{odeh1991elucidation, schomberg2005evaluating, brigham2022new}. This makes the task challenging for appearance-dominated models and useful for evaluating whether multimodal systems learn physically meaningful, geographically transferable representations.

We introduce EarthScape, a multimodal benchmark for surface-aware representation learning and controlled cross-region transfer, using surficial geologic mapping as a testbed for \underbar{Earth} surface and land\underbar{scape} analysis. EarthScape combines raw sensor data, derived terrain representations, and contextual layers. Each patch contains 37 co-registered channels of RGB+NIR imagery, elevation, multi-scale terrain derivatives, hydrology and infrastructure vector layers, along with surficial geologic map masks. EarthScape uses a shared geologic label space across two geographically separated regions, enabling controlled in- and cross-domain evaluation of representation choice, spatial scale, and multimodal fusion. Our contributions are:

\begin{itemize}
\item We introduce EarthScape as a diagnostic benchmark for multimodal, surface-aware representation learning and cross-region transfer under a shared label space.

\item We define standardized cross-domain evaluation protocols for comparing raw sensors, derived terrain representations, multi-scale features, and fusion strategies.

\item We provide benchmark results showing that engineered terrain representations improve geographic robustness, multi-scale and multimodal terrain representations provide the strongest overall performance, and fusion effectiveness depends on backbone and evaluation domain.

\end{itemize}

\section{Related Work}

\subsection{Geospatial Benchmarks and Domain Shift}
Remote-sensing benchmarks such as SpaceNet \citep{van2018spacenet}, xView \citep{lam2018xview}, and Functional Map of the World \citep{christie2018functional} have enabled progress on scene classification and object detection using high-resolution satellite imagery, while BigEarthNet \citep{sumbul2019bigearthnet}, DeepGlobe \citep{demir2018deepglobe}, and SEN12MS \citep{schmitt2019sen12ms} support land-cover classification and segmentation using multispectral or synthetic aperture radar (SAR) imagery. These resources provide foundational benchmarks for geospatial vision, but they largely emphasize land-cover, land-use, or object-centric categories rather than Earth surface structure. Recent work has highlighted geographic domain shift as a central challenge for Earth observation models, but fewer benchmarks isolate how modality choice, surface representation, and fusion strategy affect transfer between regions with a shared label space \citep{al2025analysing, ekim2025distribution, marsocci2023geomultitasknet, wang2021loveda}. EarthScape specifically addresses this gap by evaluating these implementation choices under controlled geographic shift.

\subsection{Multimodal and Surface-Aware Learning} 
Multimodal learning is increasingly important in geospatial computer vision, where optical imagery, SAR, elevation, and vector data can provide complementary information for recognition tasks \citep{astruc2024omnisat, bi2022vision, jain2022multimodal, han2024bridging, li2022deep}. Many geospatial tasks also require hierarchical spatial reasoning, since local texture, object-scale structure, terrain morphology, and regional context may each provide useful cues \citep{behrens2018multi, liao2026segformer}. This has motivated architectures that capture multi-scale dependencies, including hierarchical vision transformers and segmentation models, as well as feature representations computed across multiple spatial scales \citep{liu2021swin, xie2021segformer, fan2021multiscale, niu2021review, hassanin2024visual}. Prior work has also explored fusion at different stages, including early, feature-level, and decision-level fusion, with mechanisms ranging from simple channel stacking to more complex attention-based integration \citep{baltruvsaitis2018multimodal, steyaert2023multimodal, li2024crossfuse}. However, these studies do not identify a universally robust fusion strategy, and comparisons under geographic shift remain difficult across datasets.

\subsection{Surficial Geologic Mapping}
Surficial geologic mapping aims to delineate near-surface materials shaped by weathering, erosion, transport, deposition, and human modification \citep{Compton1985}. Automated approaches to related geologic surface-mapping problems have mostly focused on single-target geologic hazard tasks, such as landslide and sinkhole identification \citep{kirkwood2016machine, zhu2016applying, crawford2021using, prakash2021new, ji2020landslide, liu2023feature, zhou2023hyper, rafique2022automatic}. A smaller body of work addresses multiclass geologic mapping, using either traditional machine-learning approaches with handcrafted predictors \citep{cracknell2014geological, latifovic2018assessment, johnson2025machine}, or with newer deep learning models that rely on specialized inputs \citep{wang2021lithological, zhou2023hyper, liu2024deep}. Although these studies demonstrate the potential of automated geologic mapping, many are evaluated in local areas, use private or specialized datasets, define task-specific label spaces, and do not provide common train/validation/test splits. Reported models therefore remain difficult to compare and, when evaluated across regions, often show limited transfer beyond their study areas \citep{latifovic2018assessment, rafique2022automatic, johnson2025machine}. This leaves a benchmark gap for evaluating geologic mapping under shared protocols that explicitly test representation choice, fusion strategy, and robustness.

\begin{figure*}[!t]
\centering
\includegraphics[width=\linewidth]{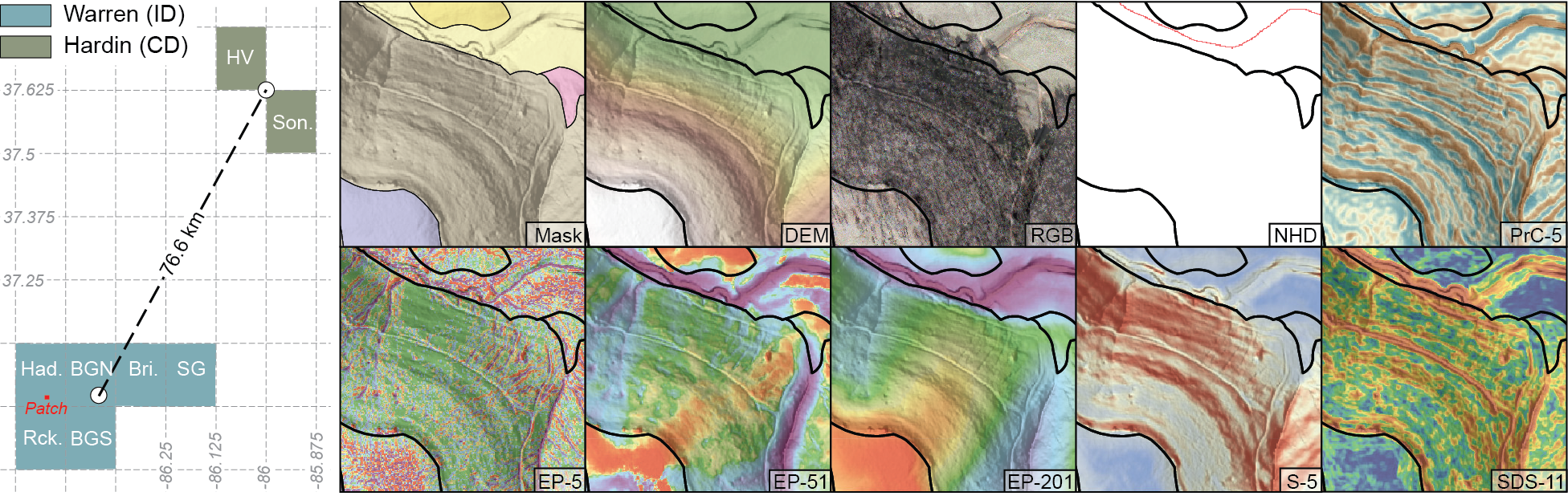}
\caption{EarthScape study regions and selected modalities. Left: in-domain (ID) Warren and cross-domain (CD) Hardin subsets with latitude--longitude grid lines and centroid distance between areas. Right: selected channels from one $256\times256$ patch (location shown in red on map), including the geologic mask, raw DEM and RGB inputs, hydrography (NHD), and DEM-derived terrain features at multiple spatial supports. Each patch is overlaid on a hillshade image for visualizing surface morphology, and geologic class boundaries (black lines) are shown for reference.}
\label{fig:dataset}
\end{figure*}

\section{EarthScape Benchmark Design}

\subsection{Dataset Composition}

\noindent\textbf{Surficial Geologic Maps: }
Detailed 1:24,000 scale surficial geologic mapping is sparse, with less than 14\% of the United States currently mapped \citep{usgs_ngmdb_2025}. Suitable benchmark targets are further constrained by the need for compatible map-unit definitions, mapping conventions and interpretations, and validated digital access. EarthScape uses eight 1:24,000-scale surficial geologic maps from two regions in the central United States selected for compatibility \citep{surficial_rockfield, surficial_hadley, surficial_bgn, surficial_bgs, surficial_bristow, surficial_smithsgrove, surficial_howevalley, surficial_sonora} (Fig. \ref{fig:dataset}). The original vector polygon maps are rasterized to produce pixel-level masks and patch-level labels. Across both regions, EarthScape defines a shared label space of seven mutually exclusive geologic materials, including fluvial deposits (Qal, alluvium; Qat, terrace deposits), debris-flow deposits (Qaf, alluvial fans), hillslope materials (Qc, colluvium; Qca, colluvial aprons), in-situ weathering products (Qr, residuum), and anthropogenic modification (af1, artificial fill). The shared target space in EarthScape enables in-domain and cross-domain evaluation, while retaining labels grounded in real surface processes. See Supp. \ref{supp-a:surficial} for more information.

\medskip
\noindent\textbf{Aerial Imagery and Elevation: }
EarthScape includes aerial imagery and digital elevation models (DEMs) as its core raw sensor modalities \citep{kyfromabove_imagery} (Fig. \ref{fig:dataset}). Optical imagery has a ground sampling distance (GSD) of 6 in ($\approx$ 0.15 m) and includes red (R), green (G), blue (B), and near-infrared (NIR) channels. The DEM is produced from airborne LiDAR at 5 ft GSD ($\approx$ 1.52 m) and provides raw elevation.

\medskip
\noindent\textbf{DEM-derived Terrain Features: }
EarthScape includes five DEM-derived terrain features widely used in geomorphometry \citep{florinsky2016digital}: slope (S), profile curvature (PrC), planform curvature (PlC), elevation percentile (EP), and standard deviation of slope (SDS) (Fig. \ref{fig:dataset}). These features quantify complementary aspects of surface morphology, including local steepness, surface curvature, relative elevation, and terrain roughness. To represent spatial hierarchies and scale relationships in surface structure, terrain features are computed at six spatial supports. S, PrC, and PlC are calculated from the native DEM and five additional DEMs resampled from 5 ft ($\approx$1.52 m) GSD to 10 ft ($\approx$3.05 m), 20 ft ($\approx$6.10 m), 50 ft ($\approx$15.24 m), 100 ft ($\approx$30.48 m), and 200 ft ($\approx$60.96 m) GSD. EP and SDS are computed as neighborhood statistics on the native DEM using kernels with matching approximate spatial supports. These features are included because geomorphometric structure and spatial scale relationships are central to surficial mapping \citep{Compton1985, Lisle2011, Pavlis2017, florinsky2016digital}, and they also enable benchmark comparisons between raw elevation, explicit terrain morphology, and multi-scale surface representations. Individual terrain features are denoted by the spatial support used during their computation. EP and SDS use kernel size (e.g., EP-51 indicates EP computed with a $51 \times 51$ kernel), whereas S, PrC, and PlC use the GSD of the source DEM (e.g., PlC-5 indicates computation from the 5 ft GSD DEM). Multi-scale stacks contain all supports for a feature and are denoted with the suffix -ms, such as EP-ms. See Supp. \ref{supp-b} for more details.

\medskip
\noindent\textbf{Hydrography and Infrastructure: }
Contextual GIS layers for surface hydrography and human infrastructure are included in EarthScape (Fig. \ref{fig:dataset}). Hydrographic features consist of stream centerlines and waterbody polygons from the U.S. Geological Survey National Hydrography Dataset (NHD) \citep{nhd_hr_usgs}, while infrastructure features include road and railway centerlines from OpenStreetMap (OSM) \citep{osm_roads_rail}. These vector layers are rasterized into binary images that represent drainage-network and built-environment context.

\subsection{Benchmark Properties}

\noindent\textbf{Overview and Structure: }
EarthScape spans two disjoint regions in Kentucky, USA, consisting of 23,566 patches from the Warren area and 7,452 patches from the Hardin area, which are separated by nearly 77 km (Fig. \ref{fig:dataset}). Each patch is 256$\times$256 pixels with 50\% overlap, aligned to a common 5 ft ($\approx$ 1.52 m) GSD grid, and contains 37 co-registered predictive channels, including RGB+NIR imagery, DEM, multi-scale terrain derivatives, and binary hydrography and infrastructure layers (Fig. \ref{fig:dataset}). Each patch also includes a co-registered pixel-level surficial geologic mask, proportional class-area summary, and multi-hot label vector, enabling semantic segmentation, multilabel classification, class-area regression, and multitask configurations. The benchmark is extensible to additional modalities, geographic regions, map scales, and label ontologies.

\begin{figure}[!t]
\centering
\includegraphics[width=\linewidth]{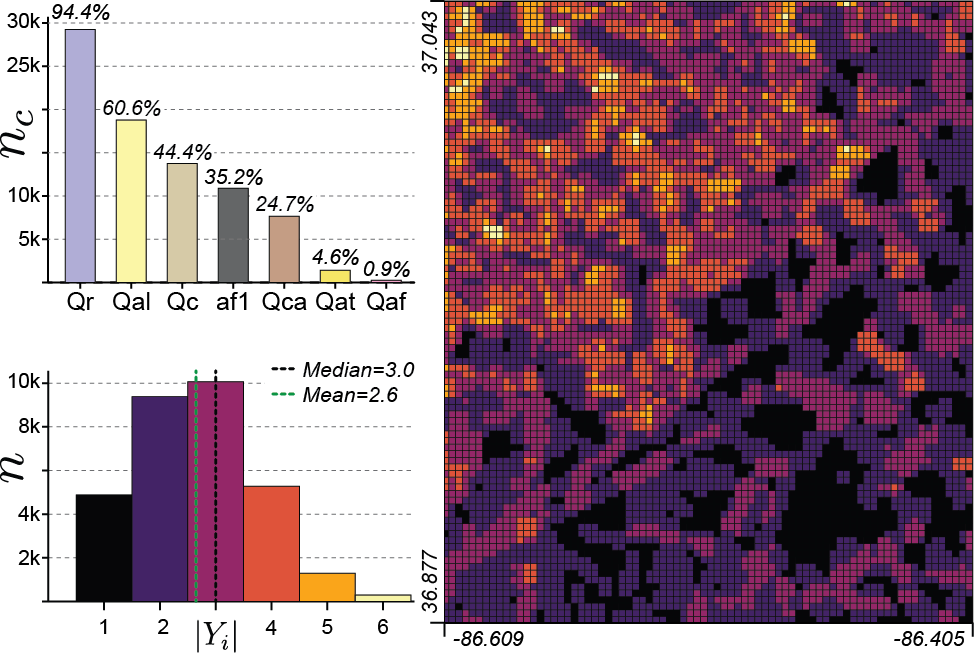}
\caption{EarthScape dataset complexity. Top left: class frequency ($n_c$) with percentages indicating the relative proportion of patches containing each class. Bottom left: sample frequency ($n$) of label cardinality ($|Y_i|$). Right: spatial distribution of $|Y_i|$ illustrating geospatial variation in patch complexity.}
\label{fig:stats}
\end{figure}

\begin{table}[!t]
\centering
\caption{Label statistics and imbalance metrics for EarthScape, including global frequency ($n_c$), mean class-area proportion, majority area rate (MAR), effective number of samples (ENS) \citep{cui2019classbalancedlossbasedeffective}, and the imbalance ratio per label (IRLbl) \citep{charte2013first}.}
\label{tab:stats}
\scriptsize
\begin{tabular}{l c cc cc}
\toprule
\textbf{Class} & \textbf{$n_c$} & \textbf{Mean Area} & \textbf{MAR} & \textbf{ENS} & \textbf{IRLbl} \\
\midrule
Qr  & 29271	& 0.651 & 0.702 & 9464.6 & 1.0\\
Qal & 18801 & 0.089 & 0.058 & 8474.4 & 1.6\\
Qc  & 13768 & 0.142 & 0.148 & 7476.3 & 2.1\\
af1 & 10910 & 0.051 & 0.035 & 6641.4 & 2.7\\
Qca & 7669  & 0.061 & 0.054 & 5355.7 & 3.8\\
Qat & 1435  & 0.006 & 0.004 & 1336.9 & 20.4\\
Qaf & 270   & 0.000 & 0.000 & 266.4  & 108.4\\
\bottomrule
\end{tabular}
\end{table}

\medskip
\noindent\textbf{Class Distribution and Imbalance: }
EarthScape exhibits a pronounced long-tailed distribution across its seven geologic classes (Table \ref{tab:stats}; Fig. \ref{fig:stats}). Qr appears in 94.4\% of patches, whereas Qat and Qaf appear in only 4.6\% and 0.9\%, respectively. The effective number of samples ranges from 9,464 for Qr to 266 for Qaf, and the imbalance ratio per label spans more than two orders of magnitude (1.0--108.4), reflecting severe class imbalance. Beyond global frequency, EarthScape also exhibits substantial intra-patch mixing. Patch-level class frequencies show that many patches contain multiple geologic classes, and the mean class-area proportions demonstrate uneven contributions of spatial footprints. The majority-area rate indicates that Qr occupies the largest area in more than 70\% of patches, while rare classes almost never occupy the largest fraction. Patch-level class frequencies also vary across space, highlighting the benchmark challenge of learning rare and spatially mixed surface-material classes (Fig. \ref{fig:stats}).

\medskip
\noindent\textbf{Geographic Domain Shift: }
The two regions were selected so that geographic variation could be evaluated while retaining the same target ontology and mapping framework. We compute maximum mean discrepancy (MMD) \citep{gretton2012kernel} to quantify distributional differences between patch-level feature summaries (P5, P10, P25, P50, P75, P90, P95) of selected input modalities from each region (Supp. Table~\ref{supp-c:tab_mmd}). We observe measurable domain shift, including MMD values of 0.136 for RGB, 0.273 for DEM, and 0.074 for the multi-scale terrain stack $\mathrm{\text{EP-ms}}+\mathrm{\text{S-ms}}+\mathrm{\text{SDS-ms}}$. The shared label space allows EarthScape to isolate regional input shift without changing the prediction task, providing a controlled setting for studying how modality choice and surface representation affect cross-region transfer. See Supp. \ref{supp-c:mmd} for more details.


\begin{figure*}[!t]
\centering
\includegraphics[width=\linewidth]{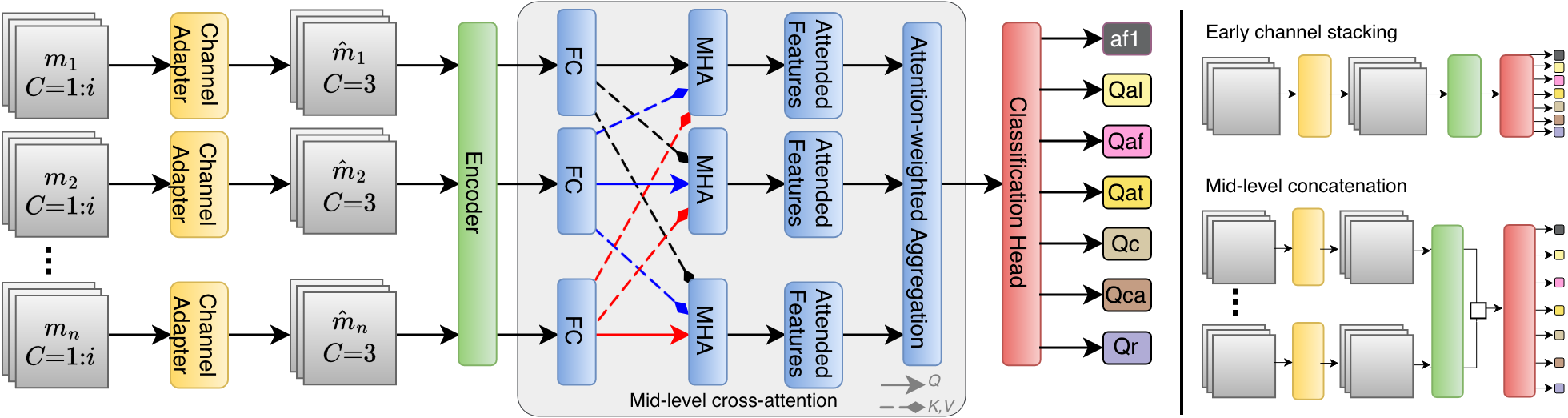}
\caption{SGMap-Net and fusion baselines. Left: SGMap-Net accepts an arbitrary set of modality inputs, adapts each branch to three channels, encodes them with a shared backbone, and fuses modality features using multi-head attention and attention-weighted aggregation before multilabel classification. Right: simplified baselines used to compare fusion strategy, including early channel stacking before encoding and mid-level concatenation after modality-specific encoding.}
\label{fig:model}
\end{figure*}

\section{Methods}

\subsection{Benchmark Task}
We evaluate EarthScape using multilabel classification over multimodal geospatial patches. Each sample is a $256 \times 256$ patch with co-registered input modalities and a target vector indicating which classes are present within the corresponding pixel-level mask. Let $\mathcal{D}=\{(x_i,y_i)\}_{i=1}^{N}$ denote the dataset, where $x_i=\{m_1,m_2,\dots,m_n\}$ is a collection of modality-specific input tensors and each modality $m_j$ may contain one or more channels, $C_i$, such as a single DEM channel, an RGB+NIR image, or a multi-scale stack of terrain features. The target $y_i \in \{0,1\}^{K}$ is a binary vector over $K=7$ geologic classes. A class is marked positive if any pixel of that class intersects the patch. This inclusive criterion preserves rare and spatially narrow classes, including units partially intersecting patch boundaries, while avoiding an arbitrary area threshold for class presence. It also requires models to identify classes from potentially limited spatial evidence rather than only dominant units. Models learn a mapping $f:x_i \rightarrow [0,1]^K$ that predicts per-class probabilities for each patch. We use this task as a tractable diagnostic protocol for comparing modality combinations, derived surface representations, fusion strategies, and cross-region robustness.

\subsection{SGMap-Net and Fusion Baselines} 
We use SGMap-Net as a lightweight benchmark model for comparing inputs, fusion strategies, and backbone architectures. SGMap-Net is not intended as an architectural contribution, but its simple modular design provides a controlled framework for isolating how raw sensor inputs, derived surface representations, and multi-scale features contribute to in-domain and cross-domain performance. 

Figure \ref{fig:model} illustrates the architecture of SGMap-Net, which consists of three key components: a channel adapter module, a feature extractor, and a classification head. For each modality, we first stack its channels and then apply a $1\times1$ convolution followed by batch normalization and ReLU activation to adapt the input to a common channel dimension of $C=3$, 
\begin{equation}
    \hat{m_i} = \mathrm{ReLU}(\mathrm{BN}(\mathrm{Conv1\times1}(m_i))).
\end{equation}
This ensures compatibility with a shared encoder, while preserving modality-specific convolutions. 

Each adapted modality $\hat m_i$ is passed through a shared encoder to extract spatial features $F_i=\mathrm{Encoder}(\hat m_i)$; we use ResNeXt-50 \citep{xie2017aggregated} and Vision Transformer (ViT-B/16) \citep{dosovitskiy2020image} backbones. For cross-attention, encoder features are represented as spatial tokens and projected to dimension $d$, with a learnable modality embedding $e_i$ added to obtain $Z_i=\mathrm{FC}(F_i)+e_i$. For each modality $m_i$, multi-head attention \citep{vaswani2017attention} is computed using $Z_i$ as the query ($Q$), and tokens from all other modalities as keys ($K$) and values ($V$):
\begin{equation}
    A_i = \mathrm{MHA}(Q=Z_i, K=[Z_j]_{j\neq i}, V=[Z_j]_{j\neq i}).
\end{equation}
Each attended token sequence $A_i$ is mean-pooled to $a_i$. The resulting modality representations are concatenated as $a=[a_i]$, and a learnable linear projection $v$ followed by Softmax produces attention weights $w=\mathrm{Softmax}(v^T a)$. The final fused representation is computed as $z_{\mathrm{fused}}=\sum_{i=1}^{N}w_i a_i$, adaptively emphasizing informative modalities for each sample. The fused embedding is then passed through a classification head consisting of two fully connected layers to predict the geologic class logits $\hat{y}$.

In addition to attention-based fusion, we evaluate (1) early channel stacking, and (2) mid-level feature concatenation, where modalities are independently adapted and encoded using the shared encoder (Fig. \ref{fig:model}). For single inputs and early stacking, spatial features are globally pooled and passed directly to the classification head. For concatenation, modality-specific spatial features are independently pooled, concatenated, and then passed to the classification head. These variants allow us to test whether more expressive modality-aware fusion improves robustness over simpler fusion strategies under the same benchmark protocol.

\subsection{Data Splits} 
We define benchmark splits to support complementary in-domain and cross-domain evaluation. The larger Warren subset (23,566 candidate patches) is used for training, validation, and in-domain testing, while the Hardin subset (7,452 patches) is reserved for cross-domain testing (Fig.~\ref{fig:dataset} and Supp. Fig.~\ref{supp-c:fig_splits}). Within Warren, we first randomly sample 1,536 patches for the in-domain test set, followed by 768 non-overlapping patches for validation; the training set consists of the remaining 8,416 patches that do not overlap either set. A separate cross-domain test set of 1,536 patches is randomly sampled from Hardin (Fig.~\ref{fig:dataset} and Supp. Fig.~\ref{supp-c:fig_splits}).

The in-domain split is designed to evaluate generalization across the regional data distribution while retaining the natural spatial autocorrelation of geologic units and preventing patch overlap between training, validation, and test sets. Distributed sampling avoids restricting evaluation to a geographically localized subpopulation of Warren's geology and yields similar class distributions across splits (Supp. Fig.~\ref{supp-c:fig_splits}). In contrast, the Hardin test set evaluates transfer to a distinct geographic domain. All primary experiments use fixed benchmark splits for reproducibility; sensitivity to alternative random partitions is evaluated separately (Table~\ref{supp-c:tab_split_variance}). See Supp.~\ref{supp-c:patch_selection} and \ref{supp-c:split_sensitivity} for additional details.

\begin{table*}
\centering
\caption{Representative SGMap-Net classification results for ResNeXt and ViT backbones. Inputs were selected by the highest combined ID$+$CD F1 within the single modality, multi-scale, and multimodal experiments; all fusion strategies are shown for the multi-scale and multimodal selections. Values report mean and standard deviation (SD) across five random seeds; $\Delta=\bar{\mathrm{F1}}_{\mathrm{ID}}-\bar{\mathrm{F1}}_{\mathrm{CD}}$ and $\Delta\%=100 (\bar{\mathrm{F1}}_{\mathrm{ID}}-\bar{\mathrm{F1}}_{\mathrm{CD}})/\bar{\mathrm{F1}}_{\mathrm{ID}}$ denote absolute and relative cross-domain degradation, respectively.}
\label{tab2:results}
\scriptsize
\setlength{\tabcolsep}{5pt}
\begin{tabular}{l lcccc c lcccc}
\toprule
\multirow{2}{*}{\textbf{Fusion}} & \multicolumn{5}{c}{\textbf{ResNeXt}} && \multicolumn{5}{c}{\textbf{ViT}} \\
\cmidrule{2-6} \cmidrule{8-12}
& \textbf{Input} & $\bar{\text{F1}}_{\text{ID}}\pm\text{SD}$ & $\bar{\text{F1}}_{\text{CD}}\pm\text{SD}$ & $\Delta$ & $\Delta\%$ && \textbf{Input} & $\bar{\text{F1}}_{\text{ID}}\pm\text{SD}$ & $\bar{\text{F1}}_{\text{CD}}\pm\text{SD}$ & $\Delta$ & $\Delta\%$ \\
\midrule
- & DEM & 0.682$\pm$0.003 & 0.526$\pm$0.026 & 0.156 & 22.9 && DEM & 0.655$\pm$0.015 & 0.524$\pm$0.007 & 0.131 & 20.0 \\
- & S-10 & 0.653$\pm$0.015 & 0.587$\pm$0.014 & 0.066 & 10.1 && S-5 & 0.642$\pm$0.010 & 0.587$\pm$0.015 & 0.055 & 8.6 \\
Stacking & PrC-ms & 0.654$\pm$0.014 & 0.593$\pm$0.018 & 0.061 & 9.3 && S-ms & 0.650$\pm$0.008 & 0.583$\pm$0.021 & 0.067 & 10.3 \\
Concat. & PrC-ms & 0.606$\pm$0.011 & 0.509$\pm$0.027 & 0.097 & 16.0 && S-ms & 0.659$\pm$0.010 & 0.569$\pm$0.016 & 0.090 & 13.7 \\
CA & PrC-ms & 0.612$\pm$0.020 & 0.536$\pm$0.024 & 0.076 & 12.4 && S-ms & 0.636$\pm$0.010 & 0.577$\pm$0.016 & 0.059 & 9.3 \\
Stacking & PrC-ms+S-ms & 0.673$\pm$0.003 & 0.609$\pm$0.008 & 0.064 & 9.5 && EP-ms+PrC-ms+S-ms & 0.670$\pm$0.006 & 0.605$\pm$0.024 & 0.065 & 9.7 \\
Concat. & PrC-ms+S-ms & 0.644$\pm$0.007 & 0.580$\pm$0.022 & 0.064 & 9.9 && EP-ms+PrC-ms+S-ms & 0.677$\pm$0.005 & 0.596$\pm$0.020 & 0.081 & 12.0 \\
CA & PrC-ms+S-ms & 0.649$\pm$0.012 & 0.587$\pm$0.018 & 0.062 & 9.6 && EP-ms+PrC-ms+S-ms & 0.667$\pm$0.008 & 0.584$\pm$0.019 & 0.083 & 12.4 \\
\bottomrule
\end{tabular}
\end{table*}

\subsection{Training and Evaluation} 
Each modality was normalized using channel-specific means and standard deviations computed from the training set. Data augmentation consisted of random flips and $90^\circ$ rotations, which preserve the orientation-invariant patch labels while avoiding potential label mismatch introduced by arbitrary-angle rotations. A common loss function was used across all classification experiments. We evaluated multiple class-balancing strategies and focal-loss parameterizations across representative input configurations, and selected focal loss \citep{lin2017focal} with $\gamma=0.5$ as the fixed benchmark loss. All models used ImageNet-pretrained encoders and were fully fine-tuned under a common training protocol. Models were optimized using AdamW with a learning rate of $3\times10^{-4}$ and batch size of 16, with early stopping after five epochs without improvement in validation loss. The checkpoint with the lowest validation loss was retained for evaluation. Label-wise decision thresholds were selected using the validation set and subsequently applied unchanged to both the in-domain and cross-domain test sets. Performance is reported using per-class and macro-averaged precision, recall, F1, average precision (AP), and area under the receiver operating characteristic curve (AUC). Additional details are provided in Supp.~\ref{supp-c:loss} and \ref{supp-c:hardware_training}.

\section{Benchmark Experiments and Results}


\subsection{Modality and Representation}
Input representation strongly influenced both in-domain performance and cross-domain transfer (Table~\ref{tab2:results}; Supp. Fig.~\ref{supp-d:fig_id_cd_summary}; Supp. Tables~\ref{supp-d:tab_single_f1_map}--\ref{supp-d:tab_single_p_r}, \ref{supp-d:tab_ms_f1_map}--\ref{supp-d:tab_ms_p_r}, \ref{supp-d:tab_mm_f1_map}--\ref{supp-d:tab_mm_p_r}). Raw elevation provided strong in-domain performance, but degraded substantially across regions. DEM achieved an in-domain F1 of 0.682$\pm$0.003 with ResNeXt, but decreased to 0.526$\pm$0.026 in cross-domain, corresponding to 22.9\% relative degradation. In contrast, S-10 achieved slightly lower in-domain performance of 0.653$\pm$0.015, but substantially higher cross-domain F1 of 0.587$\pm$0.014, reducing degradation to 10.1\%. The same pattern occurred with ViT, where DEM decreased from 0.655$\pm$0.015 in-domain to 0.524$\pm$0.007 cross-domain (20.0\% drop), whereas S-5 achieved 0.642$\pm$0.010 in-domain and 0.587$\pm$0.015 cross-domain with only 8.6\% degradation. Thus, explicitly encoding surface morphology produced substantially more transferable representations than raw elevation, despite being derived from the same underlying DEM.

Terrain performance also depended on spatial support, with considerable variation among scales of the same engineered feature (Table~\ref{tab2:results}; Supp. Tables~\ref{supp-d:tab_single_f1_map}--\ref{supp-d:tab_single_p_r}, \ref{supp-d:tab_ms_f1_map}--\ref{supp-d:tab_ms_p_r}). Multi-scale aggregation provided additional information in some cases, although the benefit depended on representation, backbone, and fusion strategy. With ResNeXt, multi-scale PrC stacking achieved 0.654$\pm$0.014 in-domain and 0.593$\pm$0.018 cross-domain F1 (9.3\% degradation), modestly improving over the best single-scale terrain representation. With ViT, S-ms did not improve cross-domain performance over S-5, although stacking increased in-domain F1 from 0.642$\pm$0.010 to 0.650$\pm$0.008. The strong scale dependence of individual features and variable gains from multi-scale aggregation indicate that spatial support is itself an important dimension of terrain representation.

Multimodal gains depended on the information represented rather than the number of input modalities (Table~\ref{tab2:results}; Supp. Tables~\ref{supp-d:tab_mm_f1_map}--\ref{supp-d:tab_mm_p_r}). Combining raw and contextual inputs (DEM, RGB, NHD, and OSM) did not consistently improve upon DEM alone, whereas combinations of engineered terrain representations produced the strongest overall performances. With ResNeXt, PrC-ms+S-ms stacking achieved 0.673$\pm$0.003 in-domain and 0.609$\pm$0.008 cross-domain F1, while ViT EP-ms+PrC-ms+S-ms stacking achieved 0.670$\pm$0.006 and 0.605$\pm$0.024, respectively. Engineered terrain representations also improved combinations containing raw modalities. For example, DEM+RGB performance improved from 0.642 in-domain and 0.519 cross-domain, to 0.656 and 0.547 when EP-ms+S-ms+SDS-ms was added. This suggests that explicit surface morphology provides complementary information that is not consistently recovered by simply combining raw inputs.

\subsection{Geographic Robustness and Distribution Shift}
The cross-domain degradation patterns (Table~\ref{tab2:results}; Supp. Fig.~\ref{supp-d:fig_id_cd_summary}; Supp. Tables~\ref{supp-d:tab_single_f1_map}--\ref{supp-d:tab_single_p_r}, \ref{supp-d:tab_ms_f1_map}--\ref{supp-d:tab_ms_p_r}, \ref{supp-d:tab_mm_f1_map}--\ref{supp-d:tab_mm_p_r}) were broadly consistent with the magnitude of distribution shift measured by MMD (Supp. Table~\ref{supp-c:tab_mmd}). DEM exhibited high MMD (0.273) and degraded by 22.9\% with ResNeXt and 20.0\% with ViT. In contrast, the more robust terrain representations exhibited substantially lower MMD: S-10 had an MMD of 0.035 and degraded by 10.1\% with ResNeXt, while S-5 had an MMD of 0.030 and degraded by 8.6\% with ViT. This pattern extended to richer representations, where PrC-ms had an MMD of 0.057 and degraded by 9.3\% with ResNeXt stacking, while PrC-ms+S-ms had an MMD of 0.072 and degraded by 9.5\%. 

Statistical analysis supported observations (Table~\ref{tab3:mmd_stats}; Supp. Fig.~\ref{supp-c:fig_mmd_delta}; Supp. Table~\ref{supp-c:tab_mmd_delta_stats}). For ViT, MMD was significantly correlated with F1 degradation for single inputs and all fusion strategies. The association was moderate for single inputs ($R^2=0.358$, $r=0.598$, $\rho=0.513$, $\tau=0.371$) and substantially stronger for fused representations ($R^2=0.729$--$0.824$, $r=0.854$--$0.908$, $\rho=0.729$--$0.854$, $\tau=0.543$--$0.695$). ResNeXt showed weaker linear associations ($R^2=0.062$--$0.247$), but stacking retained significant rank correlations ($\rho=0.639$, $p_{\mathrm{Holm}}=0.041$; $\tau=0.505$, $p_{\mathrm{Holm}}=0.033$), indicating a monotonic relationship despite weak linear correspondence. These transfer patterns remained stable across five independently generated geographic splits (Supp. Table~\ref{supp-c:tab_split_variance}). Collectively, the results show that reduced covariate shift was generally associated with improved geographic robustness, but the strength and form of this relationship depended on the backbone and fusion strategy. See Supp.~\ref{supp-c:stat_analysis} for more details.

\begin{table}
\centering
\caption{Statistical association between cross-domain F1 degradation ($\Delta=\mathrm{ID}-\mathrm{CD}$) and maximum mean discrepancy (MMD) for unimodal, and multiscale $+$ multimodal experiments stratified by encoder and fusion method. PCC ($r$): Pearson's correlation coefficient; SRCC ($\rho$): Spearman's rank correlation coefficient. All $p$-values were Holm-corrected for multiple comparisons. Significant correlations marked in \textbf{bold}.}
\label{tab3:mmd_stats}
\scriptsize
\begin{tabular}{llc cc c cc}
\toprule
\multirow{2}{*}{\textbf{Encoder}} & \multirow{2}{*}{\textbf{Fusion}} & \multirow{2}{*}{\textbf{$n$}} & \multicolumn{2}{c}{\textbf{PCC}} && \multicolumn{2}{c}{\textbf{SRCC}} \\
\cmidrule{4-5} \cmidrule{7-8}
&&& $r$ & $p_\text{Holm}$ && $\rho$ & $p_\text{Holm}$ \\
\midrule
\multirow{4}{*}{ResNeXt}
& - & 35 & 0.248 & 0.451 && 0.368 & 0.090 \\
& Stacking & 15 & 0.369 & 0.451 && \textbf{0.639} & \textbf{0.041} \\
& Concat. & 15 & 0.497 & 0.237 && 0.454 & 0.179 \\
& CA & 15 & 0.286 & 0.451 && 0.318 & 0.248 \\
\midrule
\multirow{4}{*}{ViT}
& - & 35 & \textbf{0.598} & \textbf{0.001} && \textbf{0.513} & \textbf{0.010} \\
& Stacking & 15 & \textbf{0.859} & \textbf{0.000} && \textbf{0.729} & \textbf{0.010} \\
& Concat. & 15 & \textbf{0.854} & \textbf{0.000} && \textbf{0.807} & \textbf{0.002} \\
& CA & 15 & \textbf{0.908} & \textbf{0.000} && \textbf{0.854} & \textbf{$<$0.001} \\
\bottomrule
\end{tabular}
\end{table}

\subsection{Fusion Strategy}
Fusion performance varied with backbone and evaluation domain, but increased fusion complexity did not provide a systematic advantage. Across the 15 multi-scale and multimodal configurations, ResNeXt favored stacking in-domain, with a mean F1 of 0.653 compared with 0.595 for concatenation and 0.605 for cross-attention. These differences were significant for stacking versus concatenation ($p_{\mathrm{Holm}}<0.001$) and stacking versus cross-attention ($p_{\mathrm{Holm}}<0.001$), while concatenation and cross-attention did not differ significantly (Table~\ref{tab4:fusion_stats}; Supp. Fig.~\ref{supp-c:fig_fusion_f1_comparisons}; Supp. Table \ref{supp-c:tab_fusion_stats}). Under cross-domain evaluation, the mean ResNeXt F1 decreased to 0.559, 0.521, and 0.544 for stacking, concatenation, and cross-attention, respectively, but none of the pairwise differences were significant. ViT showed no significant differences among fusion strategies in-domain, where mean F1 ranged from 0.651 to 0.660. Cross-domain performance, however, was significantly lower with cross-attention than with either stacking ($p_{\mathrm{Holm}}=0.001$) or concatenation ($p_{\mathrm{Holm}}=0.003$), while stacking and concatenation did not differ significantly. These trends were observed across five training seeds and five independently generated geographic splits (Supp. Tables~\ref{supp-c:tab_seed_variance}--\ref{supp-c:tab_split_variance}). Overall, fusion effectiveness depended on backbone and evaluation domain, but cross-attention provided no measurable advantage over simpler strategies and, for ViT cross-domain transfer, significantly reduced performance. See Supp.~\ref{supp-c:stat_analysis} for more details.

\begin{table}
\centering
\caption{Statistical comparisons of stacking, concatenation, and cross-attention (CA) fusion paired by input configurations, stratified by encoder and evaluation domain (ID, in-domain; CD, cross-domain). All $p$-values were Holm-corrected within each encoder--domain family. Significant comparisons marked in \textbf{bold}.}
\label{tab4:fusion_stats}
\scriptsize
\begin{tabular}{lllc cc}
\toprule
\textbf{Encoder} & \textbf{Domain} & \textbf{Comparison} & \textbf{$n$} & \textbf{$t$} & \textbf{$p_{\mathrm{Holm}}$} \\
\midrule
\multirow{6}{*}{ResNeXt}
& \multirow{3}{*}{ID} 
& Stacking--Concat. & 15 & \textbf{8.045} & \textbf{$<$0.001} \\
&& Stacking--CA & 15 & \textbf{10.209} & \textbf{$<$0.001} \\
&& Concat.--CA & 15 & -1.019 & 0.326 \\
\cmidrule{2-6}
& \multirow{3}{*}{CD} 
& Stacking--Concat. & 15 & 2.167 & 0.144 \\
&& Stacking--CA & 15 & 0.939 & 0.364 \\
&& Concat.--CA & 15 & -1.871 & 0.165 \\
\midrule
\multirow{6}{*}{ViT}
& \multirow{3}{*}{ID} 
& Stacking--Concat. & 15 & -1.121 & 0.409 \\
&& Stacking--CA & 15 & 1.331 & 0.409 \\
&& Concat.--CA & 15 & 2.025 & 0.187 \\
\cmidrule{2-6}
& \multirow{3}{*}{CD} 
& Stacking--Concat. & 15 & 1.609 & 0.130 \\
&& Stacking--CA & 15 & \textbf{4.521} & \textbf{0.001} \\
&& Concat.--CA & 15 & \textbf{3.951} & \textbf{0.003} \\
\bottomrule
\end{tabular}
\end{table}

\subsection{Foundation Model Comparisons}
We used compatible EarthScape channels with existing foundation models Copernicus-FM \citep{wang2025towards}, DOFA \citep{xiong2024neural}, Galileo \citep{tseng2025galileolearninggloballocal}, Panopticon \citep{waldmann2025panopticon}, and TerraMind \citep{jakubik2025TerraMind}, and results support the importance of terrain representation (Table~\ref{tab5:foundation_models}; Supp. Table~\ref{supp-c:tab_foundation_models}). Among models evaluated with optical inputs, Panopticon achieved the strongest performance, reaching 0.632 in-domain and 0.507 cross-domain F1. In contrast, models capable of accepting terrain inputs showed substantial gains. Galileo improved from 0.548 in-domain and 0.286 cross-domain with RGB+NIR to 0.593 and 0.508 with DEM+S-5, while TerraMind improved from 0.590 and 0.414 with RGB to 0.629 and 0.511 with DEM. Multimodal inputs did not improve performance. Galileo performance decreased to 0.548 in-domain and 0.272 cross-domain with RGB+NIR+DEM+S-5, and TerraMind similarly performed worse with DEM+RGB. This behavior was also observed with SGMap-Net, where adding RGB to DEM did not improve overall performance. These results reinforce the value of explicit terrain morphology information for surface-aware learning and demonstrate that multimodal performance depends on the task-specific utility of the included modalities rather than modality count alone. See Supp.~\ref{supp-c:foundation_models} for more details.

\begin{table}[!t]
\centering
\caption{Foundation model classification performance using compatible EarthScape inputs. In-domain (ID) and cross-domain (CD) macro-averaged F1 for the best-performing learning rates, with absolute ($\Delta$) and relative ($\Delta\%$) cross-domain degradation.}
\label{tab5:foundation_models}
\scriptsize
\setlength{\tabcolsep}{5pt}
\begin{tabular}{llcccc}
\toprule
\textbf{Model} & \textbf{Input} & \textbf{${\text{F1}}_{\text{ID}}$} & \textbf{${\text{F1}}_{\text{CD}}$} & $\Delta$ & $\Delta \%$ \\
\midrule
Copernicus-FM & RGB+NIR & 0.592 & 0.341 & 0.251 & 42.4 \\
DOFA & RGB+NIR & 0.602 & 0.386 & 0.216 & 35.9 \\
Panopticon & RGB+NIR & 0.632 & 0.507 & 0.125 & 19.8 \\
\midrule
\multirow{3}{*}{Galileo}
& RGB+NIR & 0.548 & 0.286 & 0.262 & 47.8 \\
& DEM+S-5 & 0.593 & 0.508 & 0.085 & 14.3 \\
& RGB+NIR+DEM+S-5 & 0.548 & 0.272 & 0.276 & 50.4 \\
\midrule
\multirow{3}{*}{TerraMind} 
& RGB & 0.590 & 0.414 & 0.176 & 29.8 \\ 
& DEM & 0.629 & 0.511 & 0.118 & 18.8 \\
& DEM+RGB & 0.604 & 0.384 & 0.220 & 36.4 \\
\bottomrule
\end{tabular}
\end{table}

\subsection{Class-Level Behavior}
Class-level performance broadly reflected the long-tailed label distribution, but class frequency alone did not determine prediction quality (Supp. Tables~\ref{supp-d:tab_class_f1_id}--\ref{supp-d:tab_class_auc_cd}, \ref{supp-d:tab_ms_class_f1_id}--\ref{supp-d:tab_ms_class_auc_cd}, \ref{supp-d:tab_mm_class_f1_id}--\ref{supp-d:tab_mm_class_auc_cd}). Common classes generally achieved higher F1, although notable departures from this trend indicated differences in class separability. For example, Qc consistently outperformed the more prevalent Qal, while Qca generally outperformed af1 across both backbones and evaluation domains. Performance was also dependent on terrain representation and spatial support, with different classes favoring different engineered features and scales. These patterns were broadly consistent between ResNeXt and ViT, suggesting that class-specific behavior was not unique to a particular backbone. Rare classes were particularly susceptible to geographic transfer degradation. Qaf frequently produced near-zero or zero cross-domain F1, and Qat also showed substantial degradation across many configurations. However, cross-domain AUC remained relatively high for many configurations with zero F1. For example, ResNeXt S-50 produced zero Qaf F1, but an AUC of 0.878. Similar discrepancies occurred across single, multi-scale, and multimodal inputs, indicating that models often retained the ability to rank rare-class samples despite failing to produce effective final classifications. Together, these results show that class imbalance strongly influences performance, particularly under geographic transfer, while AUC indicates that rare classes can remain discriminable even when absent from thresholded predictions. Class-specific input configuration preferences further show that no single representation is optimal across the label space.

\subsection{Semantic Segmentation}
Semantic segmentation evaluated whether the trends identified through multilabel classification extended to pixel-level mapping. Results were broadly consistent across DeepLabV3+ \citep{chen2018encoder} and SegFormer \citep{xie2021segformer} (Supp. Fig.~\ref{supp-d:fig_id_cd_summary}; Supp. Tables~\ref{supp-d:seg_iou_dice}--\ref{supp-d:seg_class_dice_cd}). DEM provided strong in-domain performance, but degraded sharply in cross-domain evaluation, with mIoU decreasing from 0.486 to 0.356 for DeepLabV3+ and from 0.512 to 0.412 for SegFormer. Single-scale terrain features generally improved cross-domain performance relative to DEM. S-5 achieved cross-domain mIoU scores of 0.384 and 0.430 for DeepLabV3+ and SegFormer, respectively. Multi-scale terrain representations provided further gains, with S-ms reaching 0.485 in-domain and 0.406 cross-domain mIoU with DeepLabV3+, and 0.497 and 0.441 with SegFormer. Combining multiple terrain representations produced the strongest segmentation performance, where EP-ms+PrC-ms+S-ms achieved 0.500 in-domain and 0.406 cross-domain mIoU with DeepLabV3+, and 0.520 and 0.450 with SegFormer. Class-level segmentation results were similarly consistent with classification, with performance varying substantially among geologic units and rare classes remaining particularly challenging. See Supp.~\ref{supp-c:segmentation} for more details.

\section{Conclusions}
We introduced EarthScape, a multimodal benchmark for evaluating surface-aware representation learning under geographic domain shift. Using surficial geologic mapping as a testbed, EarthScape enables controlled comparisons among raw spectral and elevation modalities, engineered terrain representations, multi-scale features, fusion strategies, and backbone architectures. Our experiments show that engineered terrain features improve geographic robustness relative to raw elevation, that spatial scale is an important component of surface representation, and that multimodal terrain inputs provide the strongest overall performance. Fusion effectiveness depends on backbone and evaluation domain, with more complex cross-attention providing no systematic advantage over simpler stacking or concatenation. Geographic degradation is associated with representation-level covariate shift, although the strength and form depend on backbone and fusion strategy. These trends were consistent across compatible foundation models, and extended from multilabel classification to semantic segmentation, demonstrating that the observed effects were not specific to a single model or prediction task. EarthScape is geographically compact, spanning two regions within a shared geologic setting, but its dense modality stack and shared label space provide a controlled benchmark for studying cross-region transfer. The current scope reflects the limited availability of detailed, digitally validated surficial mapping with compatible label definitions and interpretation, while providing a controlled setting for cross-region evaluation. EarthScape is designed to expand to additional regions, modalities, map scales, and label ontologies.

\section{Acknowledgements}
This work was partially supported by a grant funded by the National Science Foundation (NSF) under Grant number 2344533 and the Kentucky Geological Survey.

{
    \small
    \bibliographystyle{ieeenat_fullname}
    \bibliography{main}

@String(ICCV= {Int. Conf. Comput. Vis.})

@String(ICCV  = {ICCV})

@article{surficial_rockfield,
  title={Surficial Geologic Map of the Rockfield 7.5-Minute Quadrangle, Warren, Logan, and Simpson Counties, Kentucky},
  author={Buchanan, Wes and Swallom, Meredith and Bottoms, Antonia and Massey, Matthew and Hodelka, Bailee Nicole and Morris, Emily}, 
  year={2023},
  journal={Kentucky Geological Survey Contract Report},
  volume={13}, 
  number={57}
}

@article{surficial_hadley,
  title={Surficial Geologic Map of the Hadley 7.5-Minute Quadrangle, Warren County, Kentucky},
  author={Massey, Matthew and Swallom, Meredith and Bottoms, Antonia and Buchanan, Wes and Hodelka, Bailee Nicole and Morris, Emily},
  year={2023},
  journal={Kentucky Geological Survey Contract Report},
  volume={13}, 
  number={56}
}

@article{surficial_bgn,
  title={Surficial Geologic Map of the Bowling Green North 7.5-Minute Quadrangle, Warren County, Kentucky},
  author={Swallom, Meredith and Massey, Matthew and Buchanan, Wes and Hodelka, Bailee Nicole and Hayes, Hannah and Wells III, Charles and Morris, Emily},
  year={2023},
  journal={Kentucky Geological Survey Contract Report},
  volume={13}, 
  number={55}
}

@article{surficial_howevalley,
  title={Surficial geologic map of the Howe Valley 7.5-minute quadrangle, central Kentucky},
  author={Bottoms, Antonia and Hammond, Max and Massey, Matthew and Morris, Emily and McHugh, Michelle},
  year={2021},
  journal={Kentucky Geological Survey Contract Report},
  volume={13}, 
  number={43}
}

@article{surficial_sonora,
  title={Surficial geologic map of the Sonora 7.5-minute quadrangle, central Kentucky},
  author={Massey, Matthew and Bottoms, Antonia and Hammond, Max and Morris, Emily and McHugh, Michelle},
  year={2021},
  journal={Kentucky Geological Survey Contract Report},
  volume={13}, 
  number={44}
}

@article{surficial_bgs,
  author={Massey, Matthew and Swallom, Meredith and Hodelka, Bailee and Hayes, Hannah and Wells, Charles and Martin, Steve and Morris, Emily},
  title={Surficial geologic map of the Bowling Green South 7.5-minute quadrangle, Kentucky},
  year={2024},
  journal={Kentucky Geological Survey Contract Report},
  volume={14}, 
  number={2}
}

@article{surficial_bristow,
  author={Hodelka, Bailee and Massey, Matthew and Swallom, Meredith and Martin, Steve and Wells, Charles and Morris, Emily},
  title={Surficial geologic map of the Bristow 7.5-minute quadrangle, Kentucky},
  year={2024},
  journal={Kentucky Geological Survey Contract Report},
  volume={14}, 
  number={3}
}

@article{surficial_smithsgrove,
  author={Swallom, Meredith and Hodelka, Bailee and Massey, Matthew and Hayes, Hannah and Wells, Charles and Morris, Emily},
  title={Surficial geologic map of the Smiths Grove 7.5-minute quadrangle, Kentucky},
  year={2024},
  journal={Kentucky Geological Survey Contract Report},
  volume={14}, 
  number={4}
}

@article{zhou2023hyper,
  title={A Hyper-pixel-wise Contrastive Learning Augmented Segmentation Network for Old Landslide Detection Using High-Resolution Remote Sensing Images and Digital Elevation Model Data},
  author={Zhou, Yiming and Peng, Yuexing and Li, Wei and Yu, Junchuan and Ge, Daqing and Xiang, Wei},
  journal={arXiv preprint arXiv:2308.01251},
  year={2023}
}

@article{prakash2021new,
  title={A new strategy to map landslides with a generalized convolutional neural network},
  author={Prakash, Nikhil and Manconi, Andrea and Loew, Simon},
  journal={Scientific reports},
  volume={11},
  number={1},
  pages={9722},
  year={2021},
  publisher={Nature Publishing Group UK London}
}

@article{wang2021lithological,
  title={Lithological mapping based on fully convolutional network and multi-source geological data},
  author={Wang, Ziye and Zuo, Renguang and Liu, Hao},
  journal={Remote Sensing},
  volume={13},
  number={23},
  pages={4860},
  year={2021},
  publisher={MDPI}
}

@article{liu2023feature,
  title={Feature-fusion segmentation network for landslide detection using high-resolution remote sensing images and digital elevation model data},
  author={Liu, Xinran and Peng, Yuexing and Lu, Zili and Li, Wei and Yu, Junchuan and Ge, Daqing and Xiang, Wei},
  journal={IEEE Transactions on Geoscience and Remote Sensing},
  volume={61},
  pages={1--14},
  year={2023},
  publisher={IEEE}
}

@article{rafique2022automatic,
  title={Automatic segmentation of sinkholes using a convolutional neural network},
  author={Rafique, Muhammad Usman and Zhu, Junfeng and Jacobs, Nathan},
  journal={Earth and Space Science},
  volume={9},
  number={2},
  pages={e2021EA002195},
  year={2022},
  publisher={Wiley Online Library}
}

@article{latifovic2018assessment,
  title={Assessment of convolution neural networks for surficial geology mapping in the South Rae geological region, Northwest Territories, Canada},
  author={Latifovic, Rasim and Pouliot, Darren and Campbell, Janet},
  journal={Remote sensing},
  volume={10},
  number={2},
  pages={307},
  year={2018},
  publisher={Multidisciplinary Digital Publishing Institute}
}

@article{liu2024deep,
  title={Deep learning for geological mapping in the overburden area},
  author={Liu, Yao and Cheng, Jianyuan and L{\"u}, Qingtian and Liu, Zaibin and Lu, Jingjin and Fan, Zhenyu and Zhang, Lianzhi},
  journal={Frontiers in Earth Science},
  volume={12},
  pages={1407173},
  year={2024},
  publisher={Frontiers Media SA}
}

@article{behrens2018multi,
  title={Multi-scale digital soil mapping with deep learning},
  author={Behrens, Thorsten and Schmidt, Karsten and MacMillan, Robert A and Viscarra Rossel, Raphael A},
  journal={Scientific reports},
  volume={8},
  number={1},
  pages={15244},
  year={2018},
  publisher={Nature Publishing Group UK London}
}

@article{baltruvsaitis2018multimodal,
  title={Multimodal machine learning: A survey and taxonomy},
  author={Baltru{\v{s}}aitis, Tadas and Ahuja, Chaitanya and Morency, Louis-Philippe},
  journal={IEEE transactions on pattern analysis and machine intelligence},
  volume={41},
  number={2},
  pages={423--443},
  year={2018},
  publisher={IEEE}
}

@article{steyaert2023multimodal,
  title={Multimodal data fusion for cancer biomarker discovery with deep learning},
  author={Steyaert, Sandra and Pizurica, Marija and Nagaraj, Divya and Khandelwal, Priya and Hernandez-Boussard, Tina and Gentles, Andrew J and Gevaert, Olivier},
  journal={Nature machine intelligence},
  volume={5},
  number={4},
  pages={351--362},
  year={2023},
  publisher={Nature Publishing Group UK London}
}

@article{li2024crossfuse,
  title={CrossFuse: A novel cross attention mechanism based infrared and visible image fusion approach},
  author={Li, Hui and Wu, Xiao-Jun},
  journal={Information Fusion},
  volume={103},
  pages={102147},
  year={2024},
  publisher={Elsevier}
}

@article{dosovitskiy2020image,
  title={An image is worth 16x16 words: Transformers for image recognition at scale},
  author={Dosovitskiy, Alexey},
  journal={arXiv preprint arXiv:2010.11929},
  year={2020}
}

@article{niu2021review,
  title={A review on the attention mechanism of deep learning},
  author={Niu, Zhaoyang and Zhong, Guoqiang and Yu, Hui},
  journal={Neurocomputing},
  volume={452},
  pages={48--62},
  year={2021},
  publisher={Elsevier}
}

@article{hassanin2024visual,
  title={Visual attention methods in deep learning: An in-depth survey},
  author={Hassanin, Mohammed and Anwar, Saeed and Radwan, Ibrahim and Khan, Fahad Shahbaz and Mian, Ajmal},
  journal={Information Fusion},
  volume={108},
  pages={102417},
  year={2024},
  publisher={Elsevier}
}

@inproceedings{fan2021multiscale,
  title={Multiscale vision transformers},
  author={Fan, Haoqi and Xiong, Bo and Mangalam, Karttikeya and Li, Yanghao and Yan, Zhicheng and Malik, Jitendra and Feichtenhofer, Christoph},
  booktitle={Proceedings of the IEEE/CVF international conference on computer vision},
  pages={6824--6835},
  year={2021}
}

@article{lin2017focal,
  title={Focal Loss for Dense Object Detection},
  author={Lin, T},
  journal={arXiv preprint arXiv:1708.02002},
  year={2017}
}

@inproceedings{sumbul2019bigearthnet,
  title={Bigearthnet: A large-scale benchmark archive for remote sensing image understanding},
  author={Sumbul, Gencer and Charfuelan, Marcela and Demir, Beg{\"u}m and Markl, Volker},
  booktitle={IGARSS 2019-2019 IEEE International Geoscience and Remote Sensing Symposium},
  pages={5901--5904},
  year={2019},
  organization={IEEE}
}

@article{van2018spacenet,
  title={Spacenet: A remote sensing dataset and challenge series},
  author={Van Etten, Adam and Lindenbaum, Dave and Bacastow, Todd M},
  journal={arXiv preprint arXiv:1807.01232},
  year={2018}
}

@inproceedings{demir2018deepglobe,
  title={Deepglobe 2018: A challenge to parse the earth through satellite images},
  author={Demir, Ilke and Koperski, Krzysztof and Lindenbaum, David and Pang, Guan and Huang, Jing and Basu, Saikat and Hughes, Forest and Tuia, Devis and Raskar, Ramesh},
  booktitle={Proceedings of the IEEE conference on computer vision and pattern recognition workshops},
  pages={172--181},
  year={2018}
}

@article{ji2020landslide,
  title={Landslide detection from an open satellite imagery and digital elevation model dataset using attention boosted convolutional neural networks},
  author={Ji, Shunping and Yu, Dawen and Shen, Chaoyong and Li, Weile and Xu, Qiang},
  journal={Landslides},
  volume={17},
  pages={1337--1352},
  year={2020},
  publisher={Springer}
}

@article{lam2018xview,
  title={xview: Objects in context in overhead imagery},
  author={Lam, Darius and Kuzma, Richard and McGee, Kevin and Dooley, Samuel and Laielli, Michael and Klaric, Matthew and Bulatov, Yaroslav and McCord, Brendan},
  journal={arXiv preprint arXiv:1802.07856},
  year={2018}
}

@inproceedings{christie2018functional,
  title={Functional map of the world},
  author={Christie, Gordon and Fendley, Neil and Wilson, James and Mukherjee, Ryan},
  booktitle={Proceedings of the IEEE Conference on Computer Vision and Pattern Recognition},
  pages={6172--6180},
  year={2018}
}

@article{schmitt2019sen12ms,
  title={SEN12MS--A curated dataset of georeferenced multi-spectral Sentinel-1/2 imagery for deep learning and data fusion},
  author={Schmitt, Michael and Hughes, Lloyd Haydn and Qiu, Chunping and Zhu, Xiao Xiang},
  journal={arXiv preprint 1906.07789},
  year={2019}
}

@book{Compton1985,
  author    = {Compton, Robert R.},
  title     = {Geology in the Field},
  publisher = {John Wiley \& Sons},
  address   = {New York},
  year      = {1985},
}

@book{Lisle2011,
  author    = {Lisle, Richard J. and Brabham, Peter and Barnes, John W.},
  title     = {Basic Geological Mapping},
  edition   = {5th},
  publisher = {John Wiley \& Sons},
  address   = {Chichester, UK},
  year      = {2011},
  ISBN      = {9780470686348}
}

@article{Pavlis2017,
  author  = {Pavlis, Terry L. and Mason, Kelsey A.},
  title   = {The New World of 3D Geologic Mapping},
  journal = {GSA Today},
  volume  = {27},
  number  = {9},
  pages   = {4--10},
  year    = {2017},
  doi     = {10.1130/GSATG313A.1},
}

@article{odeh1991elucidation,
  title={Elucidation of soil-landform interrelationships by canonical ordination analysis},
  author={Odeh, IOA and Chittleborough, DJ and McBratney, AB},
  journal={Geoderma},
  volume={49},
  number={1-2},
  pages={1--32},
  year={1991},
  publisher={Elsevier}
}

@article{schomberg2005evaluating,
  title={Evaluating the influence of landform, surficial geology, and land use on streams using hydrologic simulation modeling},
  author={Schomberg, Jesse D and Host, George and Johnson, Lucinda B and Richards, Carl},
  journal={Aquatic Sciences},
  volume={67},
  pages={528--540},
  year={2005},
  publisher={Springer}
}

@article{brigham2022new,
  title={A new metric for morphologic variability using landform shape classification via supervised machine learning},
  author={Brigham, Cassandra AP and Crider, Juliet G},
  journal={Geomorphology},
  volume={399},
  pages={108065},
  year={2022},
  publisher={Elsevier}
}

@article{johnson2025machine,
  title={Machine learning for surficial geologic mapping},
  author={Johnson, Sarah E and Haneberg, William C},
  journal={Earth Surface Processes and Landforms},
  volume={50},
  number={1},
  pages={e6032},
  year={2025},
  publisher={Wiley Online Library}
}

@article{cracknell2014geological,
  title={Geological mapping using remote sensing data: A comparison of five machine learning algorithms, their response to variations in the spatial distribution of training data and the use of explicit spatial information},
  author={Cracknell, Matthew J and Reading, Anya M},
  journal={Computers \& Geosciences},
  volume={63},
  pages={22--33},
  year={2014},
  publisher={Elsevier}
}

@article{kirkwood2016machine,
  title={A machine learning approach to geochemical mapping},
  author={Kirkwood, Charlie and Cave, Mark and Beamish, David and Grebby, Stephen and Ferreira, Antonio},
  journal={Journal of Geochemical Exploration},
  volume={167},
  pages={49--61},
  year={2016},
  publisher={Elsevier}
}

@article{zhu2016applying,
  title={Applying a weighted random forests method to extract karst sinkholes from LiDAR data},
  author={Zhu, Junfeng and Pierskalla Jr, William P},
  journal={Journal of Hydrology},
  volume={533},
  pages={343--352},
  year={2016},
  publisher={Elsevier}
}

@article{crawford2021using,
  title={Using landslide-inventory mapping for a combined bagged-trees and logistic-regression approach to determining landslide susceptibility in eastern Kentucky, USA},
  author={Crawford, Matthew M and Dortch, Jason M and Koch, Hudson J and Killen, Ashton A and Zhu, Junfeng and Zhu, Yichuan and Bryson, Lindsey S and Haneberg, William C},
  journal={Quarterly Journal of Engineering Geology and Hydrogeology},
  volume={54},
  number={4},
  pages={qjegh2020--177},
  year={2021},
  publisher={The Geological Society of London}
}

@inproceedings{astruc2024omnisat,
  title={Omnisat: Self-supervised modality fusion for earth observation},
  author={Astruc, Guillaume and Gonthier, Nicolas and Mallet, Clement and Landrieu, Loic},
  booktitle={European Conference on Computer Vision},
  pages={409--427},
  year={2024},
  organization={Springer}
}

@article{bi2022vision,
  title={Vision transformer with contrastive learning for remote sensing image scene classification},
  author={Bi, Meiqiao and Wang, Minghua and Li, Zhi and Hong, Danfeng},
  journal={IEEE Journal of Selected Topics in Applied Earth Observations and Remote Sensing},
  volume={16},
  pages={738--749},
  year={2022},
  publisher={IEEE}
}

@article{jain2022multimodal,
  title={Multimodal contrastive learning for remote sensing tasks},
  author={Jain, Umangi and Wilson, Alex and Gulshan, Varun},
  journal={arXiv preprint arXiv:2209.02329},
  year={2022}
}

@inproceedings{han2024bridging,
  title={Bridging remote sensors with multisensor geospatial foundation models},
  author={Han, Boran and Zhang, Shuai and Shi, Xingjian and Reichstein, Markus},
  booktitle={Proceedings of the IEEE/CVF Conference on Computer Vision and Pattern Recognition},
  pages={27852--27862},
  year={2024}
}

@misc{kyfromabove_imagery,
  author={{Commonwealth of Kentucky}},
  title={KYFromAbove: Kentucky's Elevation Data \& Aerial Photography Program},
  year={2024},
  url={https://kyfromabove.ky.gov},
  note={Aerial RGB+NIR imagery and DEM. Accessed: 2024-08-01}
}

@misc{nhd_hr_usgs,
  author       = {{U.S. Geological Survey}},
  title        = {National Hydrography Dataset (NHD) – High Resolution},
  year         = {2024},
  howpublished = {\url{https://www.usgs.gov/national-hydrography}},
  note         = {Stream centerlines and waterbody polygons. Accessed: 2024-08-01}
}

@misc{osm_roads_rail,
  author       = {{OpenStreetMap contributors}},
  title        = {OpenStreetMap Road and Railway Centerlines},
  year         = {2024},
  howpublished = {\url{https://www.openstreetmap.org}},
  note         = {Road and railway centerlines. Accessed: 2024-08-01}
}

@book{florinsky2016digital,
  title={Digital terrain analysis in soil science and geology},
  author={Florinsky, Igor},
  year={2016},
  publisher={Academic Press}
}

@inproceedings{cui2019class,
  title={Class-balanced loss based on effective number of samples},
  author={Cui, Yin and Jia, Menglin and Lin, Tsung-Yi and Song, Yang and Belongie, Serge},
  booktitle={Proceedings of the IEEE/CVF conference on computer vision and pattern recognition},
  pages={9268--9277},
  year={2019}
}

@inproceedings{xie2017aggregated,
  title={Aggregated residual transformations for deep neural networks},
  author={Xie, Saining and Girshick, Ross and Doll{\'a}r, Piotr and Tu, Zhuowen and He, Kaiming},
  booktitle={Proceedings of the IEEE conference on computer vision and pattern recognition},
  pages={1492--1500},
  year={2017}
}

@article{vaswani2017attention,
  title={Attention is all you need},
  author={Vaswani, Ashish and Shazeer, Noam and Parmar, Niki and Uszkoreit, Jakob and Jones, Llion and Gomez, Aidan N and Kaiser, {\L}ukasz and Polosukhin, Illia},
  journal={Advances in neural information processing systems},
  volume={30},
  year={2017}
}

@misc{usgs_ngmdb_2025,
  author       = {{U.S. Geological Survey}},
  title        = {National Geologic Map Database (NGMDB)},
  year         = {2025},
  howpublished = {\url{https://ngmdb.usgs.gov}},
  note         = {Accessed May 2025}
}

@article{xiong2024neural,
  title={Neural plasticity-inspired multimodal foundation model for earth observation},
  author={Xiong, Zhitong and Wang, Yi and Zhang, Fahong and Stewart, Adam J and Hanna, Joelle and Borth, Damian and Papoutsis, Ioannis and Saux, Bertrand Le and Camps-Valls, Gustau and Zhu, Xiao Xiang},
  journal={arXiv preprint arXiv:2403.15356},
  year={2024}
}

@inproceedings{waldmann2025panopticon,
  title={Panopticon: Advancing any-sensor foundation models for earth observation},
  author={Waldmann, Leonard and Shah, Ando and Wang, Yi and Lehmann, Nils and Stewart, Adam and Xiong, Zhitong and Zhu, Xiao Xiang and Bauer, Stefan and Chuang, John},
  booktitle={Proceedings of the Computer Vision and Pattern Recognition Conference},
  pages={2204--2214},
  year={2025}
}

@misc{cui2019classbalancedlossbasedeffective,
      title={Class-Balanced Loss Based on Effective Number of Samples}, 
      author={Yin Cui and Menglin Jia and Tsung-Yi Lin and Yang Song and Serge Belongie},
      year={2019},
      eprint={1901.05555},
      archivePrefix={arXiv},
      primaryClass={cs.CV},
      url={https://arxiv.org/abs/1901.05555}, 
}

@inproceedings{charte2013first,
  title={A first approach to deal with imbalance in multi-label datasets},
  author={Charte, Francisco and Rivera, Antonio and del Jesus, Mar{\'\i}a Jos{\'e} and Herrera, Francisco},
  booktitle={International conference on hybrid artificial intelligence systems},
  pages={150--160},
  year={2013},
  organization={Springer}
}

@article{gretton2012kernel,
  title={A kernel two-sample test},
  author={Gretton, Arthur and Borgwardt, Karsten M and Rasch, Malte J and Sch{\"o}lkopf, Bernhard and Smola, Alexander},
  journal={The journal of machine learning research},
  volume={13},
  number={1},
  pages={723--773},
  year={2012},
  publisher={JMLR. org}
}

@inproceedings{marsocci2023geomultitasknet,
  title={GeoMultiTaskNet: Remote sensing unsupervised domain adaptation using geographical coordinates},
  author={Marsocci, Valerio and Gonthier, Nicolas and Garioud, Anatol and Scardapane, Simone and Mallet, Cl{\'e}ment},
  booktitle={Proceedings of the IEEE/CVF Conference on Computer Vision and Pattern Recognition},
  pages={2075--2085},
  year={2023}
}

@article{al2025analysing,
  title={Analysing satellite imagery classification under spatial domain shift across geographic regions},
  author={Al-Emadi, Sara A and Yang, Yin and Ofli, Ferda},
  journal={International Journal of Computer Vision},
  volume={133},
  number={11},
  pages={7672--7709},
  year={2025},
  publisher={Springer}
}

@inproceedings{ekim2025distribution,
  title={Distribution shifts at scale: Out-of-distribution detection in earth observation},
  author={Ekim, Burak and Tadesse, Girmaw Abebe and Robinson, Caleb and Hacheme, Gilles and Schmitt, Michael and Dodhia, Rahul and Ferres, Juan M Lavista},
  booktitle={Proceedings of the Computer Vision and Pattern Recognition Conference},
  pages={2265--2274},
  year={2025}
}

@article{li2022deep,
  title={Deep learning in multimodal remote sensing data fusion: A comprehensive review},
  author={Li, Jiaxin and Hong, Danfeng and Gao, Lianru and Yao, Jing and Zheng, Ke and Zhang, Bing and Chanussot, Jocelyn},
  journal={International Journal of Applied Earth Observation and Geoinformation},
  volume={112},
  pages={102926},
  year={2022},
  publisher={Elsevier}
}

@article{ruiz2023deep,
  title={Deep learning methods applied to digital elevation models: state of the art},
  author={Ruiz-Lend{\'\i}nez, Juan J and Ariza-L{\'o}pez, Francisco J and Reinoso-Gordo, Juan F and Ure{\~n}a-C{\'a}mara, Manuel A and Quesada-Real, Francisco J},
  journal={Geocarto International},
  volume={38},
  number={1},
  pages={2252389},
  year={2023},
  publisher={Taylor \& Francis}
}

@article{wu2025integrating,
  title={Integrating DEM and Deep Learning for Forested Terrain Analysis: Enhancing Fire Risk Assessment Through Mountain Peak and Water System Extraction in Chongli District},
  author={Wu, Yihui and Sun, Xueying and Qi, Liang and Xu, Jiang and Gao, Demin and Zhu, Zhengli},
  journal={Forests},
  volume={16},
  number={4},
  pages={692},
  year={2025},
  publisher={MDPI}
}

@inproceedings{liu2021swin,
  title={Swin transformer: Hierarchical vision transformer using shifted windows},
  author={Liu, Ze and Lin, Yutong and Cao, Yue and Hu, Han and Wei, Yixuan and Zhang, Zheng and Lin, Stephen and Guo, Baining},
  booktitle={Proceedings of the IEEE/CVF international conference on computer vision},
  pages={10012--10022},
  year={2021}
}

@article{xie2021segformer,
  title={SegFormer: Simple and efficient design for semantic segmentation with transformers},
  author={Xie, Enze and Wang, Wenhai and Yu, Zhiding and Anandkumar, Anima and Alvarez, Jose M and Luo, Ping},
  journal={Advances in neural information processing systems},
  volume={34},
  pages={12077--12090},
  year={2021}
}

@article{liao2026segformer,
  title={SegFormer-based segmentation approach for landscape planning and overhead remote sensing image analysis},
  author={Liao, Sha},
  journal={Scientific Reports},
  year={2026},
  publisher={Nature Publishing Group UK London}
}

@article{wang2021loveda,
  title={LoveDA: A remote sensing land-cover dataset for domain adaptive semantic segmentation},
  author={Wang, Junjue and Zheng, Zhuo and Ma, Ailong and Lu, Xiaoyan and Zhong, Yanfei},
  journal={arXiv preprint arXiv:2110.08733},
  year={2021}
}

@book{bridge2008earth,
  title={Earth surface processes, landforms and sediment deposits},
  author={Bridge, John S and Demicco, Robert V},
  year={2008},
  publisher={Cambridge University Press}
}

@article{bloom1998geomorphology,
  title={Geomorphology: a systematic analysis of late Cenozoic landforms},
  author={Bloom, Arthur Leroy},
  journal={(No Title)},
  year={1998}
}

@book{neuendorf2005glossary,
  title={Glossary of geology},
  author={Neuendorf, Klaus KE},
  year={2005},
  publisher={Springer Science \& Business Media}
}

@techreport{weary2014karst,
  title={Karst in the United States: a digital map compilation and database},
  author={Weary, David J and Doctor, Daniel H},
  year={2014},
  institution={US Geological Survey}
}

@book{wilson2000terrain,
  title={Terrain analysis: principles and applications},
  author={Wilson, John P and Gallant, John C},
  year={2000},
  publisher={John Wiley \& Sons}
}

@article{pike2009geomorphometry,
  title={Geomorphometry: a brief guide},
  author={Pike, Richard J and Evans, IS and Hengl, Tomislav},
  journal={Developments in soil science},
  volume={33},
  pages={3--30},
  year={2009},
  publisher={Elsevier}
}

@article{lindsay2016whitebox,
  title={Whitebox GAT: A case study in geomorphometric analysis},
  author={Lindsay, John B},
  journal={Computers \& Geosciences},
  volume={95},
  pages={75--84},
  year={2016},
  publisher={Elsevier}
}

@inproceedings{jakubik2025terramind,
  title={Terramind: Large-scale generative multimodality for earth observation},
  author={Jakubik, Johannes and Yang, Felix and Blumenstiel, Benedikt and Scheurer, Erik and Sedona, Rocco and Maurogiovanni, Stefano and Bosmans, Jente and Dionelis, Nikolaos and Marsocci, Valerio and Kopp, Niklas and others},
  booktitle={2025 IEEE/CVF International Conference on Computer Vision (ICCV)},
  pages={7383--7394},
  year={2025},
  organization={IEEE}
}

@inproceedings{wang2025towards,
  title={Towards a unified copernicus foundation model for earth vision},
  author={Wang, Yi and Xiong, Zhitong and Liu, Chenying and Stewart, Adam J and Dujardin, Thomas and Bountos, Nikolaos Ioannis and Zavras, Angelos and Gerken, Franziska and Papoutsis, Ioannis and Leal-Taix{\'e}, Laura and others},
  booktitle={2025 IEEE/CVF International Conference on Computer Vision (ICCV)},
  pages={9888--9899},
  year={2025},
  organization={IEEE}
}

@misc{tseng2025galileolearninggloballocal,
      title={Galileo: Learning Global and Local Features in Pretrained Remote Sensing Models},
      author={Gabriel Tseng and Anthony Fuller and Marlena Reil and Henry Herzog and Patrick Beukema and Favyen Bastani and James R. Green and Evan Shelhamer and Hannah Kerner and David Rolnick},
      year={2025},
      eprint={2502.09356},
      archivePrefix={arXiv},
      primaryClass={cs.CV},
      url={https://arxiv.org/abs/2502.09356},
}

@inproceedings{chen2018encoder,
  title={Encoder-decoder with atrous separable convolution for semantic image segmentation},
  author={Chen, Liang-Chieh and Zhu, Yukun and Papandreou, George and Schroff, Florian and Adam, Hartwig},
  booktitle={European conference on computer vision},
  pages={833--851},
  year={2018},
  organization={Springer}
}

@misc{massey2025earthscape_dataset,
    title        = {EarthScape AI Dataset},
    author       = {Massey, Matthew and Imran, Abdullah-Al-Zubaer},
    year         = {2025},
    institution  = {Kentucky Geological Survey, University of Kentucky},
    series       = {Research Data},
    publisher    = {University of Kentucky Libraries},
    doi          = {10.13023/kgs.data.05.01.2025},
    url          = {https://doi.org/10.13023/kgs.data.05.01.2025},
    note         = {Version 1.1}
}
}

\clearpage
\appendix

\maketitlesupplementary

\section*{Availability and Reproducibility Statement}
The EarthScape dataset and benchmark code are publicly available to support reproducibility and future research. The dataset is archived at \url{https://doi.org/10.13023/kgs.data.05.01.2025} and includes multimodal benchmark patches, geologic masks, patch-level labels, train/validation/test split metadata, and preprocessing information \citep{massey2025earthscape_dataset}. The accompanying code repository (\url{https://github.com/masseygeo/earthscape}) provides scripts for data preparation, model training, and evaluation, together with the configurations necessary to reproduce the reported benchmark results.

\begin{figure}[htb]
\centering
\begin{subfigure}[t]{\linewidth}
\centering
\includegraphics[width=\linewidth]{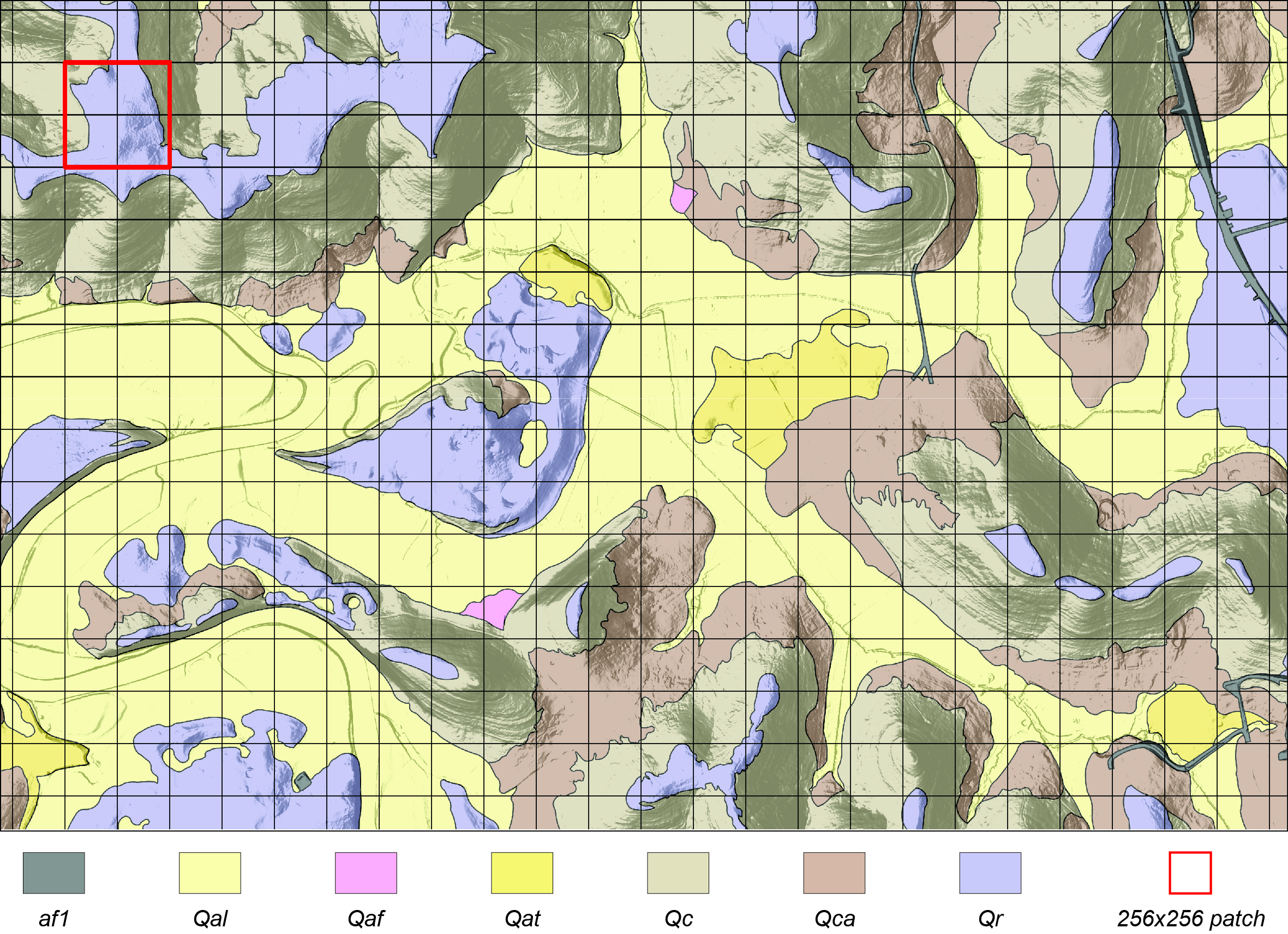}
\caption{Surficial geologic map of part of Warren County.}
\end{subfigure}
\vspace{0.5em}
\begin{subfigure}[htb]{\linewidth}
\centering
\includegraphics[width=\linewidth]{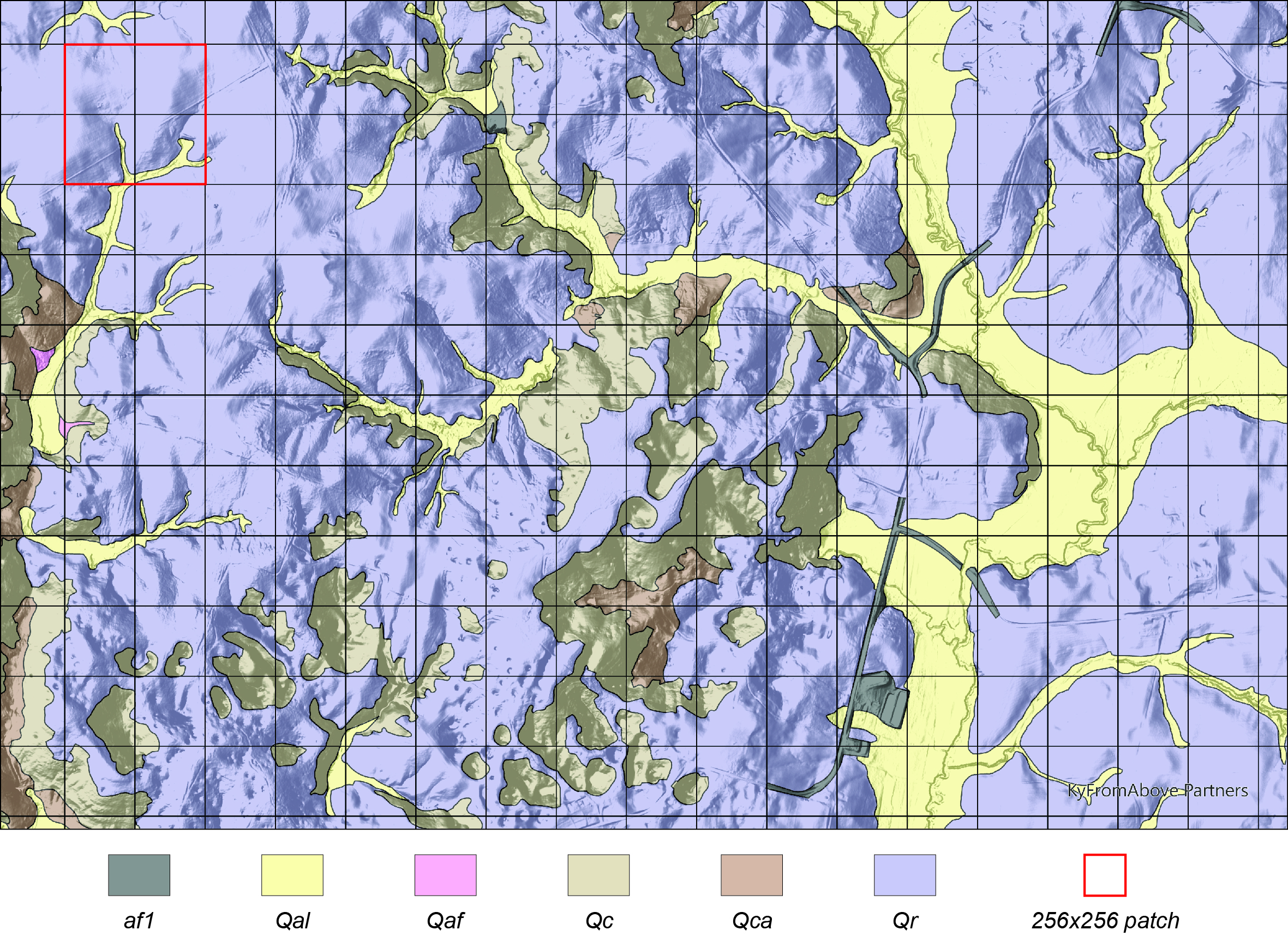}
\caption{Surficial geologic map of part of Hardin County.}
\end{subfigure}
\caption{Example surficial geologic maps overlaid on hillshade images to emphasize topographic context. The spatial correspondence between geologic classes and landscape features is visually apparent. The black grid indicates the layout of EarthScape patches, each measuring $256\times256$ pixels ($390.14\times390.14$ m) with 50\% overlap. Red squares in the upper left corners highlight a single patch.}
\label{supp-a:fig_maps}
\end{figure}

\begin{table*}[htb]
\centering
\caption{Descriptions of surficial geologic units represented in EarthScape.}
\label{supp-a:tab_geologic_unit_descriptions}
\scriptsize
\begin{tabular}{lccl p{0.53\textwidth}}
\toprule
\textbf{Geologic Class} & \textbf{Notation} & \textbf{Encoding} & \textbf{Geomorphic Process} & \textbf{Characteristic Topographic Expression and Visual Cues} \\
\midrule
Artificial fill & af1 & 0 & Anthropogenic & Rectilinear shapes with sharp edges; surface patterns inconsistent with natural terrain. \\
Alluvium & Qal & 1 & Fluvial (active) & Relatively broad, low-relief valleys, channels, and basins. \\
Alluvial fans & Qaf & 2 & Debris Flow & Small, lobate deposits at tributary mouths; wedge-shaped morphology; often isolated. \\
Terrace deposits & Qat & 3 & Fluvial (relict) & Flat benches elevated above active floodplains; pronounced scarps; isolated or paired. \\
Colluvium & Qc & 4 & Gravity-driven (active) &  Steep slopes ($\geq12^\circ$); irregular, rough terrain. \\
Colluvial aprons & Qca & 5 & Gravity-driven (stable) &  Wedge-shaped deposits along slope bases; concave profiles. \\
Residuum & Qr & 6 & In-situ weathering & Broad plains; low-relief; variable surface texture and vegetation. \\
\bottomrule
\end{tabular}
\end{table*}

\section{Geological Background}
\label{supp-a}

\subsection{Surficial Geology}
\label{supp-a:surficial}

EarthScape currently includes seven surficial geologic map classes that represent unique geomorphic surface processes and depositional environments \citep{bridge2008earth, bloom1998geomorphology, neuendorf2005glossary}. These classes can be grouped according to their dominant geomorphic process, including fluvial and debris-flow deposition (Qal, Qaf, and Qat), gravity-driven hillslope transport (Qc and Qca), and in-situ weathering (Qr). Anthropogenic modification (af1) represents an additional human-induced class that has altered or replaced the natural surficial materials. There is strong spatial correspondence between surficial geologic classes and topography (Fig.~\ref{supp-a:fig_maps}). The EarthScape benchmark is designed around this relationship to frame surficial geologic mapping as a vision task, where computer vision models can learn to associate surface patterns with underlying geological processes (Table~\ref{supp-a:tab_geologic_unit_descriptions}). EarthScape adopts the standard surficial geologic map-unit abbreviations used by the Kentucky Geological Survey. Unit symbols follow conventional geologic mapping nomenclature, where the prefix "Q" denotes Quaternary age and the remaining letters identify the mapped surficial unit. The symbols are retained unchanged from the source maps for consistency with the original geologic products.

\begin{itemize}
\smallskip
\item
\textbf{Artificial fill (af1):} Human-made deposits consisting of transported or excavated material placed or removed for engineering, mining, or other anthropogenic structures. Includes road embankments, building pads, quarries, and areas of significant topographic modification. Often exhibits sharp, angular boundaries. The spatial extent of af1 can be below the mapping resolution and inconsistently captured on expert-curated surficial geologic maps.

\medskip
\item
\textbf{Alluvium (Qal):} Unconsolidated sediments, typically consisting of clay-, silt-, sand-, and gravel-sized particles, deposited by modern rivers and streams. Qal is commonly found in active floodplains and valley bottoms and reflects recent sedimentation from overbank flooding and channel migration. These areas are generally flat, vegetated, and hydrologically dynamic.

\medskip
\item
\textbf{Alluvial fans (Qaf):} Fan-shaped deposits formed at the bases of tributaries or drainages, where sediment-laden water rapidly spreads and loses energy. These deposits are typically coarse-grained, poorly sorted, and associated with debris flows or flash floods. Although geologically significant, Qaf deposits are often small, making them inconsistently represented on typical 1:24,000-scale maps.

\medskip
\item
\textbf{Terrace deposits (Qat):} Relict alluvial sediments preserved on elevated flat surfaces above modern stream channels. These deposits reflect former floodplain levels and subsequent stream incision. Compositionally similar to Qal, but usually expressed as distinct landforms above modern floodplains.

\medskip
\item
\textbf{Colluvium (Qc):} Hillslope-derived sediments that accumulate at the bases of slopes due to gravity-driven processes such as soil creep, slopewash, and shallow landslides. Qc deposits are unsorted and variable in thickness, typically found on slopes $>12^\circ$. Qc is considered an active geomorphic unit.

\medskip
\item
\textbf{Colluvial aprons (Qca):} Slope-derived material deposited across lower hillslopes. Qca typically occurs downslope from Qc and is more stable, having accumulated over longer time periods. These deposits may be partially weathered, with poorly defined lower boundaries that grade into Qr due to extended weathering and lower erosion rates.

\medskip
\item
\textbf{Residuum (Qr):} Weathered material formed in place from the physical, chemical, and biological breakdown of underlying bedrock or older unconsolidated deposits. Qr lacks significant sediment transportation and is commonly found in upland areas with minimal active erosion. Qr is commonly gradational and poorly defined where it grades into Qc or Qca, leading to interpretive ambiguity during mapping.
\end{itemize}

\begin{figure}[b]
\centering
\includegraphics[width=\linewidth]{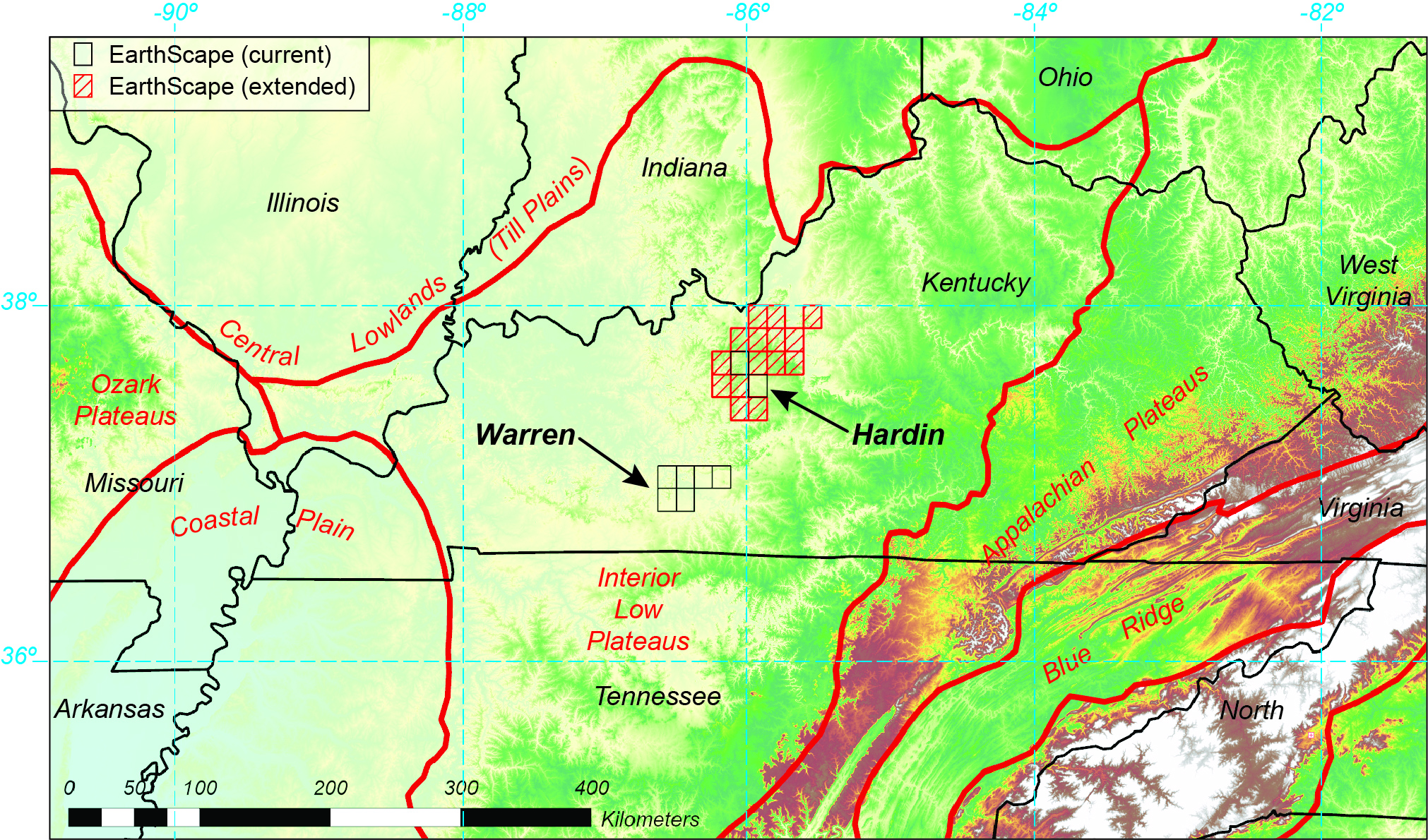}
\caption{Map of the central United States showing the publicly available 1:24,000-scale surficial geologic maps. Red lines show boundaries of major geologic provinces, which provide geological constraints for generalizability. The Interior Low Plateaus and adjacent Appalachian Plateaus may provide suitable settings for geographic transfer based on shared terrain, bedrock, and geomorphic processes. In contrast, the glaciated Central Lowlands and Coastal Plain are characterized by fundamentally different surficial processes and materials.}
\label{supp-a:fig_regional_map}
\end{figure}

\subsection{Geologic Generalization}
\label{suppA2:geo_generalization}

Although EarthScape is geographically limited to Kentucky, the surficial materials and geomorphic processes represented by the dataset are characteristic of many unglaciated temperate landscapes. The benchmark is therefore directly applicable throughout the Interior Low Plateaus and Appalachian Plateaus (Fig.~\ref{supp-a:fig_regional_map}), where similar carbonate bedrock, dissected uplands, and mixed fluvial–colluvial systems occur. Comparable geomorphic settings are also found in portions of the Ozark Plateau (USA), the Carpathians (eastern Europe), the Dinaric Alps (Balkans), and parts of central China and southeastern Australia \citep[e.g., ][]{weary2014karst}. Transferability beyond these environments should be evaluated in the context of the dominant geomorphic processes and surficial materials present. For example, landscapes shaped primarily by Quaternary glaciation contain deposits such as till, outwash, and moraines that are not represented in EarthScape, limiting direct applicability. Consequently, applications to new geographic regions should be guided by geological domain expertise to ensure that the represented surficial processes are appropriate for the intended study area.


\clearpage

\section{Terrain Feature Definitions and Properties}
\label{supp-b}

EarthScape includes five DEM-derived terrain variables computed at six spatial supports to characterize local and regional landform morphology \citep{florinsky2016digital, wilson2000terrain, pike2009geomorphometry, lindsay2016whitebox}. Collectively, these terrain features provide complementary representations of geomorphic form. Terrain features were derived from the source DEM using moving-window neighborhoods of increasing size to capture geomorphic information across multiple spatial scales. All terrain derivatives were computed using WhiteboxTools \citep{lindsay2016whitebox}. Throughout this manuscript, terrain variables are denoted as $X_r^{k \times k}$, where $X$ identifies the terrain variable, $r$ is the raster resolution, and $k \times k$ denotes the moving-window kernel used to compute the feature. Table~\ref{supp-b:tab_terrain_feature_unit_conversions} summarizes the notation, computation inputs, spatial support, and descriptive statistics for every terrain product distributed with EarthScape. 

\begin{itemize}
\smallskip
\item 
\textbf{Elevation percentile (EP):} Neighborhood statistic measuring the relative elevation of a point within a defined neighborhood, expressed as a percentile rank (0--100\%) of the elevation among neighboring values. EP characterizes relative topographic position within a local neighborhood, capturing variations in surface texture and landform morphology independent of absolute elevation.

\begin{equation}
\text{EP} = 100 \cdot \frac{ \left| \left\{ z_i \in Z \mid z_i < z \right\} \right| }{N},
\end{equation}

where $z$ is the elevation at the center cell, $Z$ is the set of elevations in the neighborhood, $z_i$ are the individual neighboring elevations, and $N$ is the total number of neighbors. The numerator counts the number of neighbors with elevation less than $z$.

\medskip
\item
\textbf{Planform curvature (PlC):} Second derivative of elevation measured perpendicular to the direction of the steepest slope. PlC characterizes lateral landform geometry by identifying areas of flow convergence and divergence, which influence erosion, sediment transport, and depositional patterns.

\begin{equation}
\text{PlC} = \frac{q^2 r - 2pq s + p^2 t}{(p^2 + q^2)^{3/2}},
\end{equation}

where \( p = \frac{\partial z}{\partial x} \) and \( q = \frac{\partial z}{\partial y} \) are the first-order partial derivatives of elevation in the x and y directions, and \( r = \frac{\partial^2 z}{\partial x^2} \), \( s = \frac{\partial^2 z}{\partial x \partial y} \), and \( t = \frac{\partial^2 z}{\partial y^2} \) are the corresponding second-order partial derivatives.

\medskip
\item 
\textbf{Profile curvature (PrC):} Second derivative of elevation measured parallel to the direction of the steepest slope. PrC characterizes longitudinal landform geometry by identifying convex, planar, and concave slope profiles that influence flow acceleration, erosion, and sediment deposition.

\begin{equation}
\text{PrC} = \frac{p^2 r + 2pq s + q^2 t}{(p^2 + q^2)^{3/2}},
\end{equation}

where \( p = \frac{\partial z}{\partial x} \) and \( q = \frac{\partial z}{\partial y} \) are the first-order partial derivatives of elevation in the x and y directions, and \( r = \frac{\partial^2 z}{\partial x^2} \), \( s = \frac{\partial^2 z}{\partial x \partial y} \), and \( t = \frac{\partial^2 z}{\partial y^2} \) are the corresponding second-order partial derivatives.

\medskip
\item 
\textbf{Slope (S):} First derivative of elevation, measuring the rate of change of elevation over a horizontal distance. S characterizes local terrain gradient and provides a primary descriptor of hillslope morphology, erosion potential, and gravitational transport.

\begin{equation}
\text{S} = \tan^{-1} \left( \sqrt{ \left( \frac{\partial z}{\partial x} \right)^2 + \left( \frac{\partial z}{\partial y} \right)^2 } \right),
\end{equation}

where \( \frac{\partial z}{\partial x} \) and \( \frac{\partial z}{\partial y} \) are the partial derivatives of elevation in the x and y directions, respectively.

\medskip
\item 
\textbf{Standard deviation of slope (SDS):} Neighborhood statistic that measures the variability in slope angle within a local window. SDS characterizes terrain roughness by quantifying local variability in slope angle, capturing changes in surface texture and topographic complexity.

\begin{equation}
\text{SDS} = \sqrt{ \frac{1}{N} \sum_{i=1}^{N} \left( S_i - \bar{S} \right)^2 },
\end{equation}

where \( S_i \) is the slope angle (in degrees or radians) of the \( i^{th} \) cell in the neighborhood, \( \bar{S} \) is the mean slope within that neighborhood, and \( N \) is the total number of cells used in the calculation window.

\end{itemize}

\begin{table*}[htb]
\centering
\caption{Terrain-feature notation used throughout the manuscript, with the source DEM resolution and kernel size used for computation. Spatial support is provided as an interpretive measure of the effective spatial area. Native processing units are feet, but metric equivalents are reported throughout the manuscript. S, PrC, and PlC are computed from native or resampled DEMs, whereas EP and SDS are computed on the native DEM using square neighborhood windows of varying size.}
\label{supp-b:tab_terrain_feature_unit_conversions}
\scriptsize
\begin{tabular}{llccc}
\toprule
\multirow{2}{*}{\textbf{Terrain Feature}} 
& \multirow{2}{*}{\textbf{Manuscript Notation}} 
& \multicolumn{2}{c}{\textbf{Computational Input}}
& \multirow{2}{*}{\textbf{Spatial Support}} \\
\cmidrule(lr){3-4}
& & \textbf{Source DEM GSD} & \textbf{Kernel Size} & \\
\midrule
\multirow{6}{*}{Elevation Percentile}
& EP-5   & 5 ft ($\approx$1.52 m) & $5 \times 5$     & 625 ft\textsuperscript{2} ($\approx$58 m\textsuperscript{2}) \\
& EP-11  & 5 ft ($\approx$1.52 m) & $11 \times 11$   & 3025 ft\textsuperscript{2} ($\approx$281 m\textsuperscript{2}) \\
& EP-21  & 5 ft ($\approx$1.52 m) & $21 \times 21$   & 11025 ft\textsuperscript{2} ($\approx$1024 m\textsuperscript{2}) \\
& EP-51  & 5 ft ($\approx$1.52 m) & $51 \times 51$   & 65025 ft\textsuperscript{2} ($\approx$6041 m\textsuperscript{2}) \\
& EP-101 & 5 ft ($\approx$1.52 m) & $101 \times 101$ & 255025 ft\textsuperscript{2} ($\approx$23693 m\textsuperscript{2}) \\
& EP-201 & 5 ft ($\approx$1.52 m) & $201 \times 201$ & 1010025 ft\textsuperscript{2} ($\approx$93834 m\textsuperscript{2}) \\
\midrule
\multirow{6}{*}{Planform Curvature}
& PlC-5 & 5 ft ($\approx$1.52 m)    & $5 \times 5$  & 625 ft\textsuperscript{2} ($\approx$58 m\textsuperscript{2}) \\
& PlC-10   & 10 ft ($\approx$3.04 m)   & $5 \times 5$  & 2500 ft\textsuperscript{2} ($\approx$232 m\textsuperscript{2}) \\
& PlC-20   & 20 ft ($\approx$6.10 m)   & $5 \times 5$  & 10000 ft\textsuperscript{2} ($\approx$929 m\textsuperscript{2}) \\
& PlC-50  & 50 ft ($\approx$15.24 m)  & $5 \times 5$  & 62500 ft\textsuperscript{2} ($\approx$5806 m\textsuperscript{2}) \\
& PlC-100  & 100 ft ($\approx$30.48 m) & $5 \times 5$  & 250000 ft\textsuperscript{2} ($\approx$23226 m\textsuperscript{2}) \\
& PlC-200  & 200 ft ($\approx$60.96 m) & $5 \times 5$  & 1000000 ft\textsuperscript{2} ($\approx$92903 m\textsuperscript{2}) \\
\midrule
\multirow{6}{*}{Profile Curvature}
& PrC-5 & 5 ft ($\approx$1.52 m)    & $5 \times 5$  & 625 ft\textsuperscript{2} ($\approx$58 m\textsuperscript{2}) \\
& PrC-10   & 10 ft ($\approx$3.04 m)   & $5 \times 5$  & 2500 ft\textsuperscript{2} ($\approx$232 m\textsuperscript{2}) \\
& PrC-20   & 20 ft ($\approx$6.10 m)   & $5 \times 5$  & 10000 ft\textsuperscript{2} ($\approx$929 m\textsuperscript{2}) \\
& PrC-50  & 50 ft ($\approx$15.24 m)  & $5 \times 5$  & 62500 ft\textsuperscript{2} ($\approx$5806 m\textsuperscript{2}) \\
& PrC-100  & 100 ft ($\approx$30.48 m) & $5 \times 5$  & 250000 ft\textsuperscript{2} ($\approx$23226 m\textsuperscript{2}) \\
& PrC-200  & 200 ft ($\approx$60.96 m) & $5 \times 5$  & 1000000 ft\textsuperscript{2} ($\approx$92903 m\textsuperscript{2}) \\
\midrule
\multirow{6}{*}{Slope}
& S-5   & 5 ft ($\approx$1.52 m)   & $5 \times 5$  & 625 ft\textsuperscript{2} ($\approx$58 m\textsuperscript{2}) \\
& S-10  & 10 ft ($\approx$3.04 m)  & $5 \times 5$  & 2500 ft\textsuperscript{2} ($\approx$232 m\textsuperscript{2}) \\
& S-20  & 20 ft ($\approx$6.10 m)  & $5 \times 5$  & 10000 ft\textsuperscript{2} ($\approx$929 m\textsuperscript{2}) \\
& S-50  & 50 ft ($\approx$15.24 m)  & $5 \times 5$  & 62500 ft\textsuperscript{2} ($\approx$5806 m\textsuperscript{2}) \\
& S-100 & 100 ft ($\approx$30.48 m) & $5 \times 5$  & 250000 ft\textsuperscript{2} ($\approx$23226 m\textsuperscript{2}) \\
& S-200 & 200 ft ($\approx$60.96 m) & $5 \times 5$  & 1000000 ft\textsuperscript{2} ($\approx$92903 m\textsuperscript{2}) \\
\midrule
\multirow{6}{*}{Standard Deviation of Slope}
& SDS-5   & 5 ft ($\approx$1.52 m) & $5 \times 5$     & 625 ft\textsuperscript{2} ($\approx$58 m\textsuperscript{2}) \\
& SDS-11  & 5 ft ($\approx$1.52 m) & $11 \times 11$   & 3025 ft\textsuperscript{2} ($\approx$281 m\textsuperscript{2}) \\
& SDS-21  & 5 ft ($\approx$1.52 m) & $21 \times 21$   & 11025 ft\textsuperscript{2} ($\approx$1024 m\textsuperscript{2}) \\
& SDS-51  & 5 ft ($\approx$1.52 m) & $51 \times 51$   & 65025 ft\textsuperscript{2} ($\approx$6041 m\textsuperscript{2}) \\
& SDS-101 & 5 ft ($\approx$1.52 m) & $101 \times 101$ & 255025 ft\textsuperscript{2} ($\approx$23693 m\textsuperscript{2}) \\
& SDS-201 & 5 ft ($\approx$1.52 m) & $201 \times 201$ & 1010025 ft\textsuperscript{2} ($\approx$93834 m\textsuperscript{2}) \\
\bottomrule
\end{tabular}
\end{table*}

\clearpage
\newpage
\begin{figure*}[h]
\centering
\includegraphics[width=\linewidth]{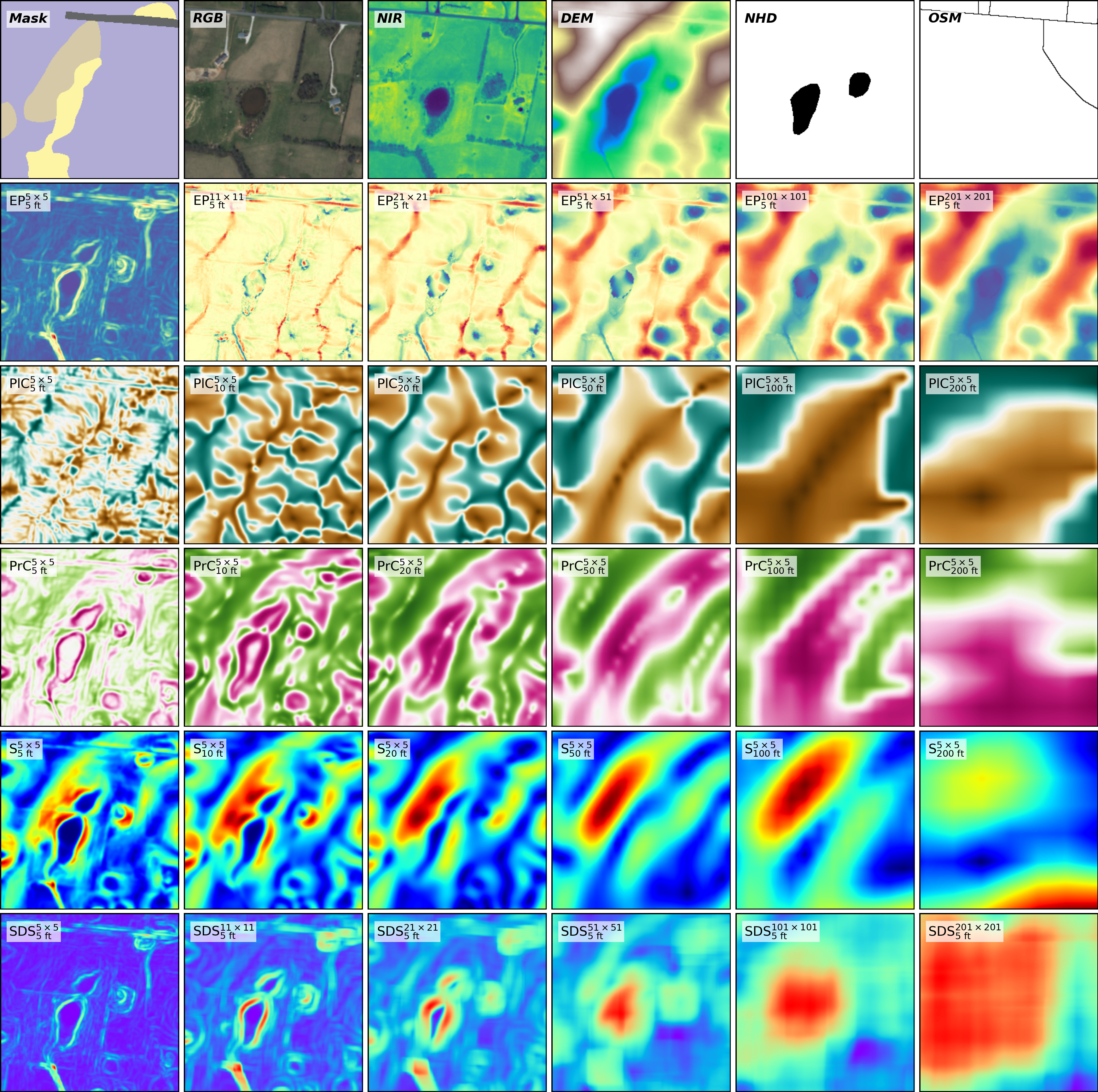}
\caption{Example patch from the Warren County area showcasing the target mask and 37 predictive channels available in EarthScape. Channels are displayed from top left to bottom right: target mask, RGB aerial imagery, NIR aerial imagery, DEM, NHD hydrologic features, OSM infrastructure, six spatial scales of S, PrC, and PlC derived from downsampled DEMs, and multiple scales of SDS and EP calculated using six kernel sizes with the original DEM.}
\label{supp-b:fig_warren_modalities}
\end{figure*}

\clearpage
\newpage
\begin{figure*}[!t]
\centering
\includegraphics[width=\linewidth]{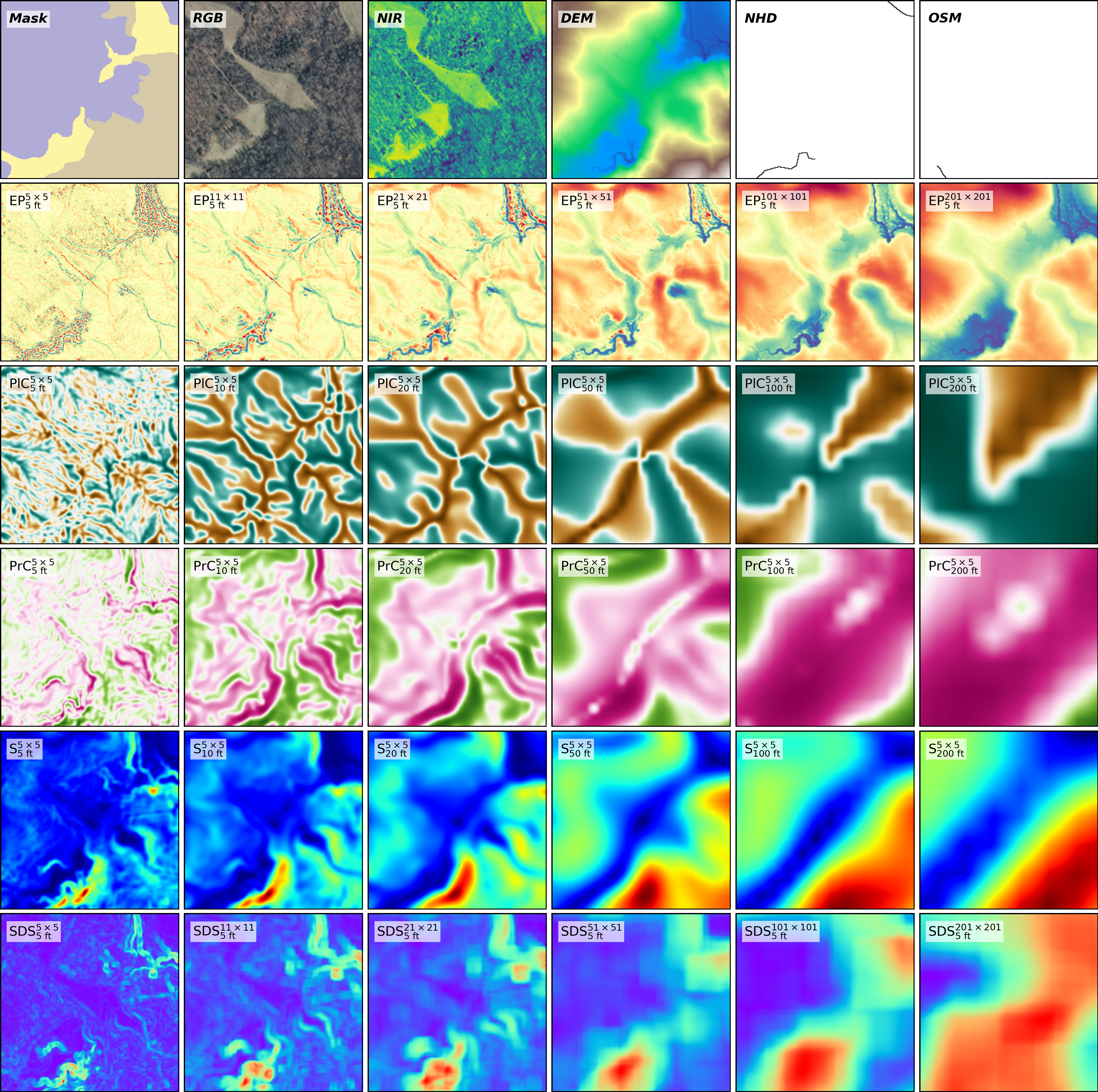}
\caption{Example patch from the Hardin County area showcasing the target mask and 37 predictive channels available in EarthScape. Channels are displayed from top left to bottom right: target mask, RGB aerial imagery, NIR aerial imagery, DEM, NHD hydrologic features, OSM infrastructure, six spatial scales of S, PrC, and PlC derived from downsampled DEMs, and multiple scales of SDS and EP calculated using six kernel sizes with the original DEM.}
\label{supp-b:fig_hardin_modalities}
\end{figure*}


\clearpage
\twocolumn

\section{Additional Experimental Details}

\subsection{Maximum Mean Discrepancy (MMD) Analysis}
\label{supp-c:mmd}
Maximum mean discrepancy (MMD) \citep{gretton2012kernel} was used to characterize differences in the distributions of input modalities between the Warren and Hardin County subsets. For each input representation, 1,000 patches were randomly sampled without replacement from each area. Each raster band within a patch was represented by its 5th, 10th, 25th, 50th, 75th, 90th, and 95th percentiles. Percentile vectors were concatenated for multiband, multi-scale, and multimodal inputs, yielding one feature vector per patch. Before MMD calculation, each vector component was standardized using the mean and standard deviation calculated from the pooled Warren and Hardin samples. Distributional differences were evaluated using an RBF kernel and the unbiased squared-MMD estimator. The reported MMD was obtained as the square root of the squared-MMD estimate. For each input, the kernel parameter was defined using the median heuristic,
\begin{equation}
\gamma =
\frac{1}
{2\,\operatorname{median}_{i<j}
\left\lVert \mathbf{x}_i-\mathbf{x}_j \right\rVert^2},
\end{equation}
where the pairwise distances were calculated from the pooled standardized sample. An input-specific value of $\gamma$ was estimated from an initial random sample and subsequently held fixed across repeated calculations for that input. The analysis was repeated for 50 independently drawn pairs of Warren and Hardin subsamples using a fixed random seed. Statistical significance was evaluated using 1,000 permutations per comparison. Mean MMD, standard deviation, and 95\% confidence intervals are reported in Table~\ref{supp-c:tab_mmd}.

\begin{table}
\centering
\caption{Mean maximum mean discrepancy (MMD), standard deviation (SD), and 95\% confidence interval half widths (CI) for raw feature input representations. Lower MMD values indicate greater distributional similarity between regions, whereas higher values indicate stronger geographic divergence.}
\label{supp-c:tab_mmd}
\scriptsize
\begin{tabular}{l ccc}
\toprule
\textbf{Input} & \textbf{MMD} & \textbf{SD} & \textbf{95\% CI} \\
\midrule
DEM & 0.273 & 0.003 & 0.001 \\
RGB & 0.136 & 0.006 & 0.002 \\
NIR & 0.287 & 0.002 & 0.001 \\
NHD & 0.036 & 0.006 & 0.002 \\
OSM & 0.028 & 0.004 & 0.001 \\
\midrule
EP-5 & 0.056 & 0.006 & 0.002 \\
EP-11 & 0.037 & 0.005 & 0.001 \\
EP-21 & 0.067 & 0.007 & 0.002 \\
EP-51 & 0.099 & 0.006 & 0.002 \\
EP-101 & 0.094 & 0.006 & 0.002 \\
EP-201 & 0.061 & 0.006 & 0.002 \\
\midrule
PlC-5 & 0.142 & 0.006 & 0.002 \\
PlC-10 & 0.117 & 0.005 & 0.002 \\
PlC-20 & 0.116 & 0.005 & 0.001 \\
PlC-50 & 0.081 & 0.006 & 0.002 \\
PlC-100 & 0.026 & 0.005 & 0.002 \\
PlC-200 & 0.007 & 0.005 & 0.001 \\
\midrule
PrC-5 & 0.030 & 0.005 & 0.001 \\
PrC-10 & 0.039 & 0.004 & 0.001 \\
PrC-20 & 0.040 & 0.004 & 0.001 \\
PrC-50 & 0.039 & 0.005 & 0.001 \\
PrC-100 & 0.036 & 0.005 & 0.001 \\
PrC-200 & 0.034 & 0.004 & 0.001 \\
\midrule
S-5 & 0.030 & 0.005 & 0.001 \\
S-10 & 0.035 & 0.004 & 0.001 \\
S-20 & 0.040 & 0.005 & 0.002 \\
S-50 & 0.054 & 0.007 & 0.002 \\
S-100 & 0.067 & 0.007 & 0.002 \\
S-200 & 0.081 & 0.006 & 0.002 \\
\midrule
SDS-5 & 0.031 & 0.006 & 0.002 \\
SDS-11 & 0.027 & 0.005 & 0.002 \\
SDS-21 & 0.030 & 0.005 & 0.001 \\
SDS-51 & 0.036 & 0.005 & 0.001 \\
SDS-101 & 0.044 & 0.006 & 0.002 \\
SDS-201 & 0.057 & 0.007 & 0.002 \\
\midrule
EP-ms & 0.092 & 0.005 & 0.002 \\
PlC-ms & 0.118 & 0.005 & 0.001 \\
PrC-ms & 0.057 & 0.004 & 0.001 \\
S-ms & 0.059 & 0.005 & 0.001 \\
SDS-ms & 0.040 & 0.005 & 0.001 \\
\midrule
DEM+RGB & 0.210 & 0.003 & 0.001 \\
DEM+NHD+OSM+RGB & 0.205 & 0.003 & 0.001 \\
NHD+OSM+RGB+S-ms & 0.105 & 0.004 & 0.001 \\
DEM+S-ms & 0.141 & 0.002 & 0.001 \\
DEM+EP-ms+RGB+S-ms+SDS-ms & 0.133 & 0.002 & 0.001 \\
EP-ms+PrC-ms & 0.088 & 0.004 & 0.001 \\
EP-ms+S-ms & 0.081 & 0.005 & 0.001 \\
PrC-ms+S-ms & 0.072 & 0.004 & 0.001 \\
EP-ms+PrC-ms+S-ms & 0.086 & 0.004 & 0.001 \\
EP-ms+S-ms+SDS-ms & 0.074 & 0.004 & 0.001 \\
\bottomrule 
\end{tabular}
\end{table}

\begin{figure*}[t]
    \centering
    \includegraphics[width=\linewidth]{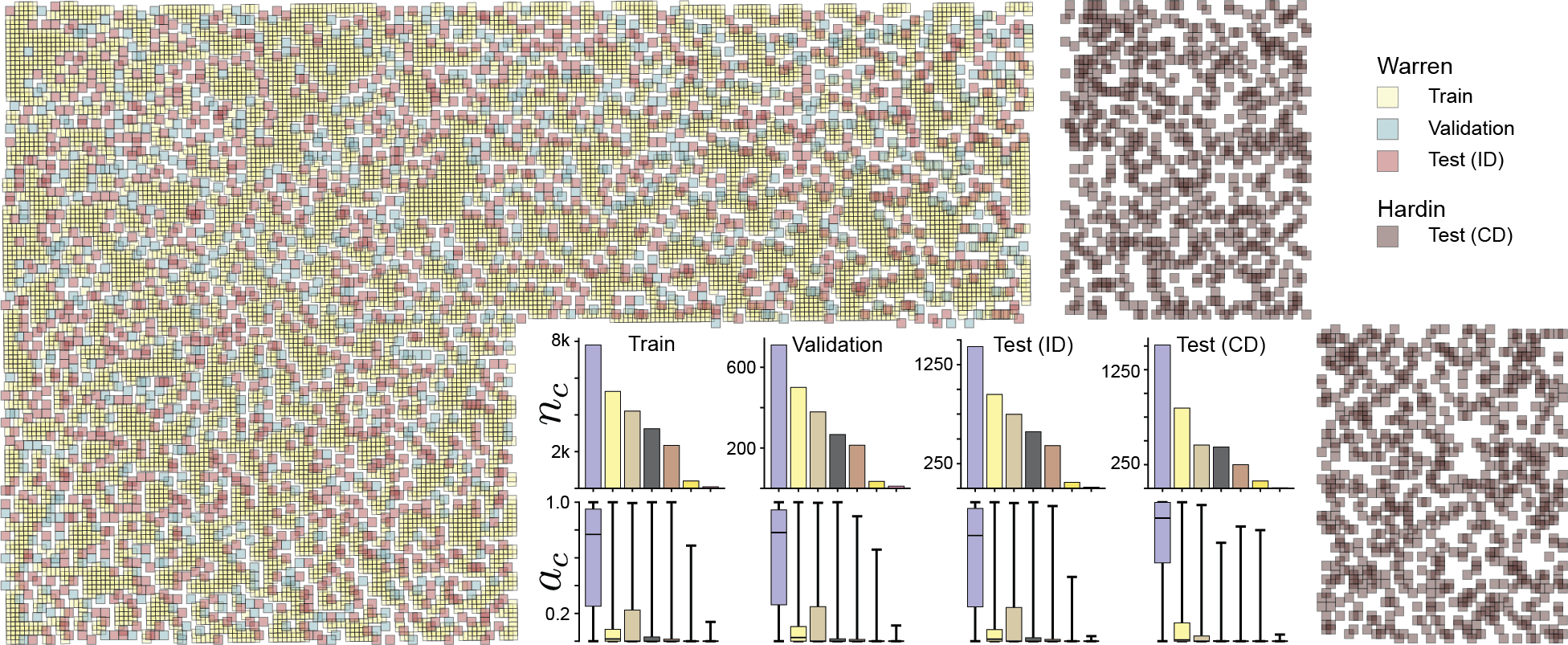}
    \caption{EarthScape benchmark split locations together with class frequency ($n_c$) and intra-patch spatial support ($a_c$) distributions for the Warren (in-domain, ID) and Hardin (cross-domain, CD) regions. Geographic locations are shown in Fig.~\ref{fig:dataset}.}
    \label{supp-c:fig_splits}
\end{figure*}

\subsection{Patch Selection and Experimental Design}
\label{supp-c:patch_selection}
EarthScape contains 31,018 total overlapping candidate patches, but the benchmark experiments use a spatially independent subset. Candidate patches are excluded when necessary to prevent spatial overlap between train, validation, and test splits. Patches may overlap within the same split because of the 50\% stride, but no patch in one benchmark split spatially intersects a patch in another split. Patches were split into spatially independent training, validation, and test sets to ensure robust and fair evaluation (Table~\ref{supp-c:tab_splits}). Warren County was used for in-domain training and evaluation due to its broader spatial coverage. We first randomly selected 1,536 test patches, followed by 768 validation patches that did not spatially intersect with the test set, and then assigned the remaining 8,416 eligible patches to the training set (Fig. \ref{supp-c:fig_splits}).

To assess geographic domain shift, we created a cross-domain test set of 1,536 patches randomly selected from the Hardin area located approximately 77 km from the Warren area (Fig. \ref{supp-c:fig_splits}). This separation enables testing model performance under geographic domain shift. Figure~\ref{supp-c:fig_splits} shows the class distributions for each data split. All subsets reflect the inherent class imbalance typical of surficial geologic mapping, driven by the localized nature of surface processes. Importantly, the class distributions are consistent across the training, validation, and both test sets, reducing the potential for differences in class representation to confound performance comparisons.

\begin{table}[htb]
\centering
\caption{Total candidate patches for the in-domain (ID) Warren and cross-domain (CD) Hardin subsets, and patch counts and proportions for the benchmark splits. Proportions are relative to the total number of selected samples in the ID splits.}
\label{supp-c:tab_splits}
\scriptsize
\setlength{\tabcolsep}{5pt}
\begin{tabular}{lcc lcc}
\toprule
\multirow{2}{*}{\textbf{Region}} & \multirow{2}{*}{\textbf{Domain}} & \textbf{Candidate} & \multirow{2}{*}{\textbf{Split}} & \textbf{Split} & \textbf{ID Split} \\
&& \textbf{Patches} && \textbf{Count} & \textbf{Proportion} \\
\midrule
\multirow{3}{*}{Warren} 
& \multirow{3}{*}{ID}
& \multirow{3}{*}{23566}
& Training & 8416 & 78.5\% \\
&&& Validation & 768 & 7.2\% \\
&&& Testing & 1536 & 14.3\% \\
\midrule
Hardin & CD & 7452 & Testing & 1536 & - \\
\bottomrule
\end{tabular}
\end{table}

\subsection{Loss Ablation}
\label{supp-c:loss}
EarthScape exhibits substantial class imbalance, motivating a staged loss-function ablation to evaluate class weighting and focal modulation independently and in combination. Each loss setting was evaluated using four representative input configurations (DEM, S-5, S-ms, and DEM+S-ms), with performance assessed using both in-domain (ID) and cross-domain (CD) macro-averaged F1. We first established an unweighted binary cross-entropy (BCE) baseline and then evaluated class-balanced BCE using three strategies for computing class-specific positive-class weights $w_c^+$: inverse class frequency (ICF), IRLbl, and the effective-number-of-samples (ENS) formulation of \cite{cui2019class}. We next evaluated focal loss \citep{lin2017focal} by varying the focusing parameter over $\gamma \in {0.5,1,1.5,2,3}$, initially without additional class weighting. Using the best-performing value ($\gamma=0.5$), we then evaluated focal-loss class balancing with $\alpha \in {0.25,0.5,0.75}$. Finally, we combined the best focal configuration with IRLbl, the best-performing positive-class weighting strategy, to test whether focal modulation and positive-class weighting provided complementary benefits. The final loss was selected according to the mean combined ID$+$CD macro-F1 across the four representative input configurations (Table~\ref{supp-c:loss_ablation}). Focal loss with $\gamma=0.5$, without additional $\alpha$ or $w_c^+$, achieved the highest aggregate performance and was therefore used for all subsequent experiments.

\begin{table*}
\centering
\caption{Macro-averaged F1 for in-domain (ID) and cross-domain (CD) performance for a loss-function ablation across four representative input configurations. Loss settings are parameterized by positive-class weight ($w_c^+$), focal focusing parameter ($\gamma$), and focal balancing parameter ($\alpha$); binary cross-entropy corresponds to $w_c^+=1$ and $\gamma=0$. Transfer degradation was computed as $\Delta=\mathrm{ID}-\mathrm{CD}$ and reported both in raw F1 scores and percentages ($\Delta\%$). Italicized values indicate mean ID$+$CD across input configurations; the mean ID$+$CD in \textbf{bold} denotes the selected loss setting.}
\label{supp-c:loss_ablation}
\scriptsize
\setlength{\tabcolsep}{5pt}
\renewcommand{\arraystretch}{0.96}
\setlength{\aboverulesep}{0pt}
\setlength{\belowrulesep}{1pt}
\begin{tabular}{cccc c ccccc}
\toprule
\textbf{$w_c^+$} & \textbf{$\gamma$} & \textbf{$\alpha$} & \textbf{Input} & \textbf{ID} & \textbf{CD} & \textbf{$\Delta$} & \textbf{$\Delta \%$} & \textbf{ID$+$CD} \\
\midrule
\multirow{4}{*}{1} & \multirow{4}{*}{0} & \multirow{4}{*}{-}
& DEM & 0.683 & 0.493 & 0.191 & 27.9 & 1.176 \\
&&& S-5 & 0.639 & 0.601 & 0.038 & 5.9 & 1.240 \\
&&& S-ms & 0.644 & 0.580 & 0.064 & 9.9 & 1.225 \\
&&& DEM+S-ms & 0.655 & 0.585 & 0.071 & 10.8 & 1.240 \\
\cmidrule{9-9}
&&&  &&&&& \textit{1.220} \\
\midrule
\multirow{4}{*}{ICF} & \multirow{4}{*}{0} & \multirow{4}{*}{-}
& DEM & 0.658 & 0.489 & 0.169 & 25.7 & 1.147 \\
&&& S-5 & 0.630 & 0.542 & 0.088 & 14.0 & 1.173 \\
&&& S-ms & 0.608 & 0.549 & 0.059 & 9.8 & 1.157 \\
&&& DEM+S-ms & 0.639 & 0.554 & 0.085 & 13.2 & 1.193 \\
\cmidrule{9-9}
&&&  &&&&& \textit{1.168} \\
\midrule
\multirow{4}{*}{IRLbl} & \multirow{4}{*}{0} & \multirow{4}{*}{-}
& DEM & 0.652 & 0.479 & 0.173 & 26.6 & 1.131 \\
&&& S-5 & 0.637 & 0.595 & 0.042 & 6.6 & 1.233 \\
&&& S-ms & 0.649 & 0.580 & 0.069 & 10.6 & 1.228 \\
&&& DEM+S-ms & 0.659 & 0.575 & 0.084 & 12.7 & 1.235 \\
\cmidrule{9-9}
&&& &&&&& \textit{1.207} \\
\midrule
\multirow{4}{*}{ENS} & \multirow{4}{*}{0} & \multirow{4}{*}{-}
& DEM & 0.669 & 0.509 & 0.161 & 24.0 & 1.178 \\
&&& S-5 & 0.623 & 0.566 & 0.058 & 9.3 & 1.189 \\
&&& S-ms & 0.634 & 0.592 & 0.042 & 6.6 & 1.226 \\
&&& DEM+S-ms & 0.621 & 0.561 & 0.060 & 9.7 & 1.181 \\
\cmidrule{9-9}
&&& &&&&& \textit{1.194} \\
\midrule
\multirow{4}{*}{1} & \multirow{4}{*}{0.5} & \multirow{4}{*}{-}
& DEM & 0.659 & 0.516 & 0.143 & 21.7 & 1.175 \\
&&& S-5 & 0.632 & 0.601 & 0.031 & 4.9 & 1.233 \\
&&& S-ms & 0.645 & 0.614 & 0.031 & 4.8 & 1.259 \\
&&& DEM+S-ms & 0.661 & 0.602 & 0.059 & 9.0 & 1.263 \\
\cmidrule{9-9}
&&& &&&&& \textbf{\textit{1.233}} \\
\midrule
\multirow{4}{*}{1} & \multirow{4}{*}{1.0} & \multirow{4}{*}{-}
& DEM & 0.674 & 0.489 & 0.185 & 27.4 & 1.163 \\
&&& S-5 & 0.630 & 0.604 & 0.026 & 4.2 & 1.234 \\
&&& S-ms & 0.631 & 0.618 & 0.013 & 2.0 & 1.249 \\
&&& DEM+S-ms & 0.654 & 0.574 & 0.080 & 12.2 & 1.228 \\
\cmidrule{9-9}
&&& &&&&& \textit{1.219} \\
\midrule
\multirow{4}{*}{1} & \multirow{4}{*}{1.5} & \multirow{4}{*}{-}
& DEM & 0.657 & 0.465 & 0.191 & 29.1 & 1.122 \\
&&& S-5 & 0.636 & 0.593 & 0.043 & 6.7 & 1.229 \\
&&& S-ms & 0.632 & 0.611 & 0.022 & 3.4 & 1.243 \\
&&& DEM+S-ms & 0.666 & 0.596 & 0.069 & 10.4 & 1.262 \\
\cmidrule{9-9}
&&& &&&&& \textit{1.214} \\
\midrule
\multirow{4}{*}{1} & \multirow{4}{*}{2.0} & \multirow{4}{*}{-}
& DEM & 0.639 & 0.497 & 0.142 & 22.3 & 1.136 \\
&&& S-5 & 0.631 & 0.591 & 0.040 & 6.3 & 1.222 \\
&&& S-ms & 0.647 & 0.606 & 0.041 & 6.4 & 1.252 \\
&&& DEM+S-ms & 0.657 & 0.598 & 0.059 & 9.0 & 1.255 \\
\cmidrule{9-9}
&&& &&&&& \textit{1.216} \\
\midrule
\multirow{4}{*}{1} & \multirow{4}{*}{3.0} & \multirow{4}{*}{-}
& DEM & 0.667 & 0.528 & 0.139 & 20.8 & 1.195 \\
&&& S-5 & 0.638 & 0.603 & 0.035 & 5.5 & 1.241 \\
&&& S-ms & 0.632 & 0.627 & 0.005 & 0.8 & 1.259 \\
&&& DEM+S-ms & 0.669 & 0.555 & 0.114 & 17.0 & 1.224 \\
\cmidrule{9-9}
&&& &&&&& \textit{1.230} \\
\midrule
\multirow{4}{*}{1} & \multirow{4}{*}{0.5} & \multirow{4}{*}{0.25}
& DEM & 0.663 & 0.526 & 0.137 & 20.6 & 1.189 \\
&&& S-5 & 0.636 & 0.594 & 0.042 & 6.6 & 1.230 \\
&&& S-ms & 0.640 & 0.612 & 0.028 & 4.4 & 1.252 \\
&&& DEM+S-ms & 0.656 & 0.565 & 0.091 & 13.9 & 1.221 \\
\cmidrule{9-9}
&&& &&&&& \textit{1.223} \\
\midrule
\multirow{4}{*}{1} & \multirow{4}{*}{0.5} & \multirow{4}{*}{0.5}
& DEM & 0.681 & 0.535 & 0.146 & 21.4 & 1.216 \\
&&& S-5 & 0.637 & 0.605 & 0.032 & 5.1 & 1.242 \\
&&& S-ms & 0.635 & 0.586 & 0.049 & 7.7 & 1.221 \\
&&& DEM+S-ms & 0.652 & 0.586 & 0.065 & 10.0 & 1.238 \\
\cmidrule{9-9}
&&& &&&&& \textit{1.229} \\
\midrule
\multirow{4}{*}{1} & \multirow{4}{*}{0.5} & \multirow{4}{*}{0.75}
& DEM & 0.664 & 0.497 & 0.167 & 25.1 & 1.161 \\
&&& S-5 & 0.637 & 0.597 & 0.040 & 6.3 & 1.233 \\
&&& S-ms & 0.662 & 0.620 & 0.042 & 6.3 & 1.282 \\
&&& DEM+S-ms & 0.656 & 0.587 & 0.069 & 10.5 & 1.242 \\
\cmidrule{9-9}
&&& &&&&& \textit{1.230} \\
\midrule
\multirow{4}{*}{IRLbl} & \multirow{4}{*}{0.5} & \multirow{4}{*}{-}
& DEM & 0.666 & 0.505 & 0.161 & 24.2 & 1.171 \\
&&& S-5 & 0.643 & 0.591 & 0.053 & 8.2 & 1.234 \\
&&& S-ms & 0.664 & 0.588 & 0.076 & 11.5 & 1.252 \\
&&& DEM+S-ms & 0.637 & 0.534 & 0.103 & 16.1 & 1.171 \\
\cmidrule{9-9}
&&& &&&&& \textit{1.207} \\
\bottomrule
\end{tabular}
\end{table*}

\subsection{Hardware, Compute, and Training}
\label{supp-c:hardware_training}
\begin{figure}[t]
    \centering
    \begin{subfigure}{0.98\linewidth}
        \centering
        \includegraphics[width=\linewidth]{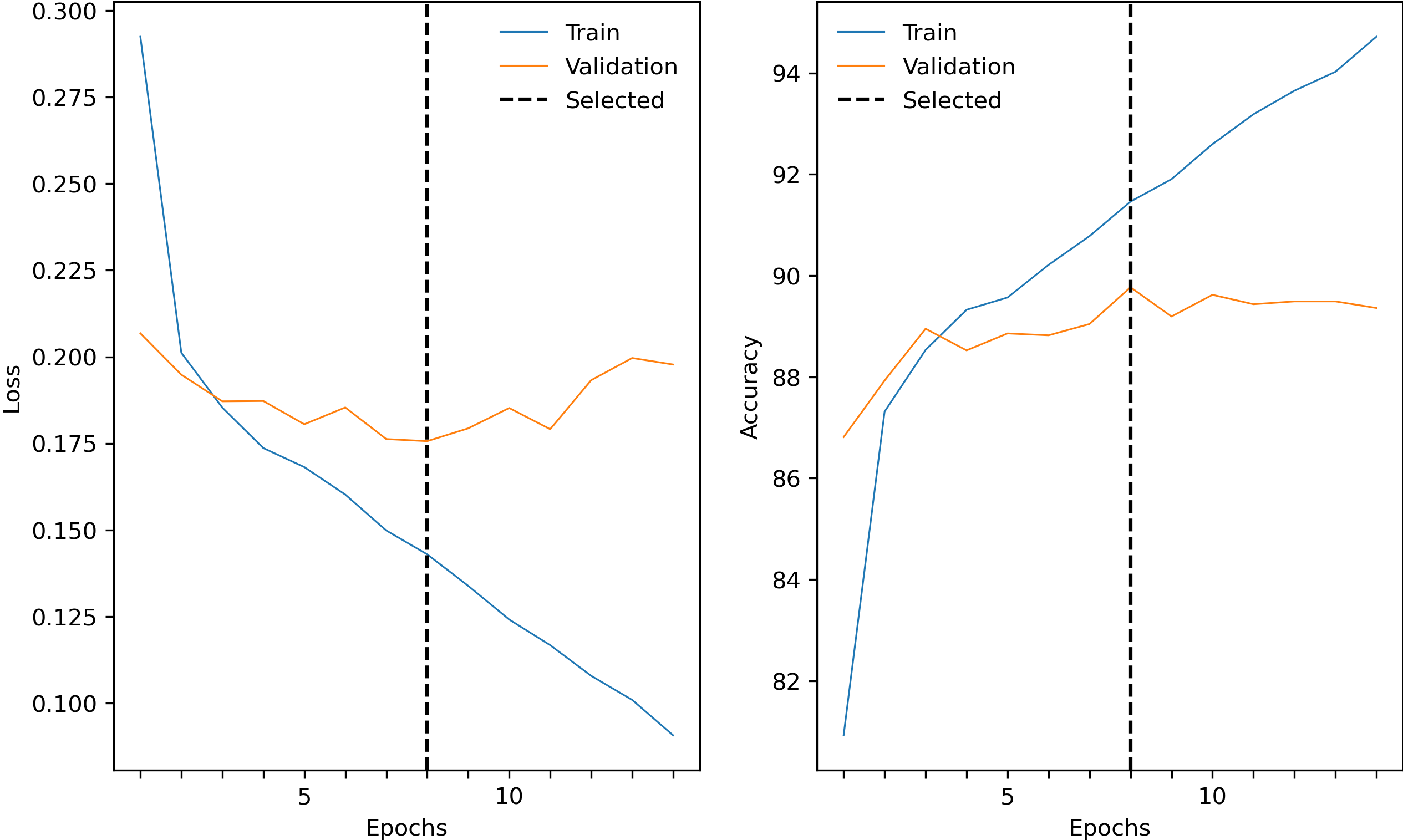}
        \caption{Training and validation curves for the DEM-only ResNeXt model.}
    \end{subfigure}
    \vspace{0.2cm}
    \begin{subfigure}{0.98\linewidth}
        \centering
        \includegraphics[width=\linewidth]{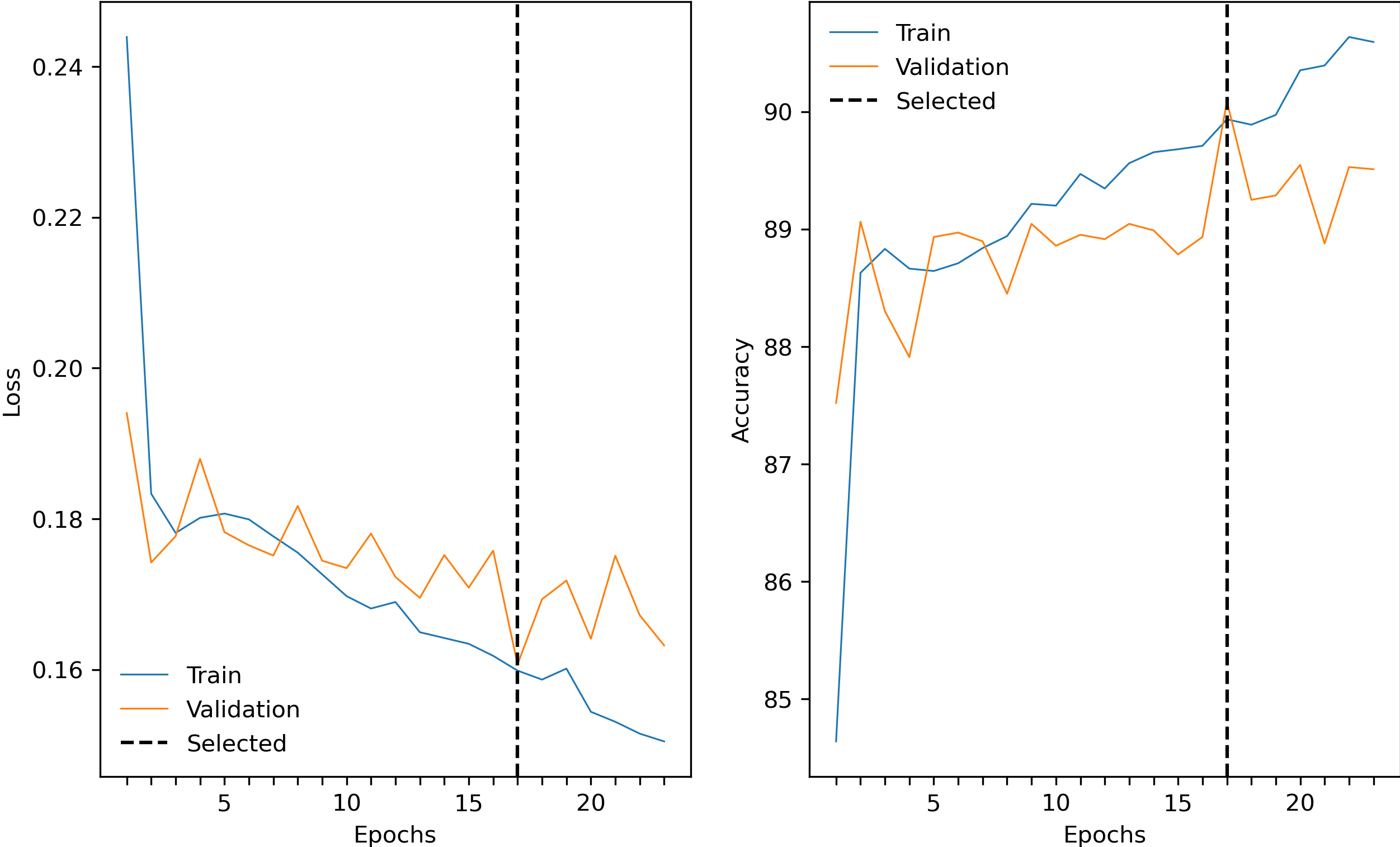}
        \caption{Training and validation curves for the DEM+RGB+EP-ms+S-ms+SDS-ms ViT model using cross-attention fusion.}
    \end{subfigure}
    \caption{Representative training behavior for single-modality and complex multimodal SGMap-Net configurations. Loss and accuracy are shown for the training and validation sets across epochs. Dashed vertical lines indicate the checkpoint selected according to minimum validation loss.}
    \label{supp-c:fig_training_curves}
\end{figure}

All experiments were implemented in Python 3.12 using PyTorch 2.10 and torchvision 0.25 and were conducted on an NVIDIA GeForce RTX 4090 GPU. All models were initialized with ImageNet-pretrained encoder weights and trained with a batch size of 16 using AdamW with a learning rate of $3\times10^{-4}$ with a linear warmup over five epochs. Training used early stopping with a patience of five epochs, where the checkpoint with the lowest validation loss was retained for evaluation. In practice, the minimum validation loss was typically reached within the first 20 epochs (Fig.~\ref{supp-c:fig_training_curves}). Model size depended primarily on the encoder and feature-fusion strategy. For a single-channel DEM input, SGMap-Net contained 24.03 M and 86.20 M trainable parameters with the ResNeXt and ViT encoders, respectively. For the largest evaluated multimodal configuration (DEM+RGB+EP-ms+S-ms+SDS-ms), stacking contained 24.03 M/86.20 M trainable parameters (ResNeXt/ViT), compared with 28.23 M/87.77 M for feature concatenation and 29.57 M/91.73 M for cross-attention. Thus, input-level stacking introduced negligible parameter overhead, whereas encoder-level fusion increased model capacity, particularly for cross-attention. Complete implementation and experiment configurations are provided in the accompanying code repository.

\subsection{Training-Seed Variability}
\label{supp-c:performance_variability}
To quantify performance variability arising from stochastic training, the training procedure was repeated using five independent random seeds while holding the training, validation, and test splits fixed. This analysis was performed for the best-performing unimodal, multi-scale, and multimodal input configurations for each benchmark architecture. Table~\ref{supp-c:tab_seed_variance} reports the individual run results and corresponding summary statistics, providing an estimate of training-seed variability and the reproducibility of benchmark performance.

\begin{table*}
\centering
\caption{Macro-averaged F1 for in-domain (ID) and cross-domain (CD) performance across five training runs with different random seeds for representative benchmark configurations. Run 1* corresponds to the fixed-seed run used for the main experiments, while Runs 2--5 used additional random seeds to quantify performance variability. Mean and standard deviation (SD) were computed across all five runs. Best and second-best scores within each run and summary column are indicated in \textbf{bold} and \underline{underlined}, respectively.}
\label{supp-c:tab_seed_variance}
\scriptsize
\setlength{\tabcolsep}{4pt}
\begin{tabular}{lll cc c cc c cc c cc c cc c cc c cc}
\toprule
\multirow{2}{*}{\textbf{Encoder}} & \multirow{2}{*}{\textbf{Input}} & \multirow{2}{*}{\textbf{Fusion}} & \multicolumn{2}{c}{\textbf{Run 1*}} && \multicolumn{2}{c}{\textbf{Run 2}} && \multicolumn{2}{c}{\textbf{Run 3}} && \multicolumn{2}{c}{\textbf{Run 4}} && \multicolumn{2}{c}{\textbf{Run 5}} && 
\multicolumn{2}{c}{\textbf{Mean$\pm$SD}} \\
\cmidrule{4-5} \cmidrule{7-8} \cmidrule{10-11} \cmidrule{13-14} \cmidrule{16-17} \cmidrule{19-20}
&&& ID & CD && ID & CD && ID & CD && ID & CD && ID & CD && ID & CD \\
\midrule
\multirow{8}{*}{ResNeXt}
& DEM & - & \textbf{0.687} & 0.508 && \textbf{0.680} & 0.543 && \textbf{0.680} & 0.530 && 0.681 & 0.555 && \textbf{0.680} & 0.492 && \textbf{0.682$\pm$0.003} & 0.526$\pm$0.026 \\
& S-10 & - & 0.674 & 0.599 && 0.639 & 0.577 && 0.651 & 0.569 && 0.663 & 0.592 && 0.639 & 0.600 && 0.653$\pm$0.015 & 0.587$\pm$0.014 \\
& PrC-ms & Stacking & 0.664 & 0.614 && 0.631 & 0.582 && 0.652 & 0.604 && 0.658 & 0.594 && 0.664 & 0.569 && 0.654$\pm$0.014 & 0.593$\pm$0.018 \\
& PrC-ms & Concat.  & 0.598 & 0.500 && 0.614 & 0.504 && 0.592 & 0.473 && 0.609 & 0.520 && 0.619 & 0.547 && 0.606$\pm$0.011 & 0.509$\pm$0.027 \\
& PrC-ms & CA       & 0.599 & 0.504 && 0.645 & 0.542 && 0.614 & 0.570 && 0.592 & 0.525 && 0.608 & 0.539 && 0.612$\pm$0.020 & 0.536$\pm$0.024 \\
& PrC-ms+S-ms & Stacking & 0.674 & 0.613 && 0.669 & 0.596 && \underbar{0.675} & \textbf{0.615} && \underbar{0.677} & 0.604 && 0.670 & \textbf{0.615} && 0.673$\pm$0.003 & \textbf{0.609$\pm$0.008} \\
& PrC-ms+S-ms & Concat.  & 0.636 & 0.544 && 0.646 & \underbar{0.599} && 0.641 & 0.577 && 0.655 & 0.593 && 0.644 & 0.589 && 0.644$\pm$0.007 & 0.580$\pm$0.022 \\
& PrC-ms+S-ms & CA       & 0.632 & 0.559 && 0.652 & 0.597 && 0.651 & 0.594 && 0.644 & 0.605 && 0.664 & 0.579 && 0.649$\pm$0.012 & 0.587$\pm$0.018 \\
\midrule
\multirow{8}{*}{ViT}
& DEM & - & 0.674 & 0.517 && 0.636 & 0.536 && 0.659 & 0.526 && 0.660 & 0.521 && 0.646 & 0.521 && 0.655$\pm$0.015 & 0.524$\pm$0.007 \\
& S-5 & - & 0.646 & 0.611 && 0.646 & 0.585 && 0.654 & 0.591 && 0.628 & 0.573 && 0.638 & 0.575 && 0.642$\pm$0.010 & 0.587$\pm$0.015 \\
& S-ms & Stacking   & 0.641 & \underbar{0.617} && 0.645 & 0.565 && 0.651 & 0.588 && 0.663 & 0.582 && 0.651 & 0.565 && 0.650$\pm$0.008 & 0.583$\pm$0.021 \\
& S-ms & Concat.    & 0.656 & 0.558 && 0.648 & 0.560 && 0.665 & 0.569 && 0.653 & 0.563 && 0.674 & 0.597 && 0.659$\pm$0.010 & 0.569$\pm$0.016 \\
& S-ms & CA         & 0.624 & 0.557 && 0.634 & 0.574 && 0.644 & 0.601 && 0.647 & 0.579 && 0.629 & 0.575 && 0.636$\pm$0.010 & 0.577$\pm$0.016 \\
& EP-ms+PrC-ms+S-ms & Stacking & 0.673 & \textbf{0.631} && 0.660 & 0.567 && 0.668 & \underbar{0.603} && 0.673 & \textbf{0.614} && 0.676 & \underbar{0.610} && 0.670$\pm$0.006 & \underbar{0.605$\pm$0.024} \\
& EP-ms+PrC-ms+S-ms & Concat.  & \underbar{0.675} & 0.570 && \underbar{0.670} & \textbf{0.613} && \textbf{0.680} & 0.579 && \textbf{0.683} & \underbar{0.607} && \underbar{0.678} & 0.609 && \underbar{0.677$\pm$0.005} & 0.596$\pm$0.020 \\
& EP-ms+PrC-ms+S-ms & CA       & 0.671 & 0.559 && 0.662 & 0.590 && 0.655 & 0.571 && 0.676 & 0.600 && 0.671 & 0.601 && 0.667$\pm$0.008 & 0.584$\pm$0.019 \\
\bottomrule
\end{tabular}
\end{table*}

\subsection{Data-Split Variability}
\label{supp-c:split_sensitivity}
To evaluate whether benchmark performance was sensitive to the particular patches assigned to each data subset, the selected input configurations were evaluated across five independently generated data splits. Each split was constructed using the same patch-selection procedure described above, with different random seeds used to select the training, validation, and test subsets while maintaining spatial independence between splits. The training seed was held fixed across all experiments to isolate variability associated with data partitioning. Table~\ref{supp-c:tab_split_variance} reports performance for each split together with the corresponding summary statistics, providing an estimate of benchmark sensitivity to the particular spatial partition of the data.

\begin{table*}
\centering
\caption{Macro-averaged F1 for in-domain (ID) and cross-domain (CD) performance across five geographic data splits for representative benchmark configurations. Split 1* corresponds to the official EarthScape benchmark split, while Splits 2--5 were generated with different random seeds to quantify performance variability across geographic splits. Mean and standard deviation (SD) were computed across all five runs. Best and second-best scores within each run and summary column are indicated in \textbf{bold} and \underline{underlined}, respectively.}
\label{supp-c:tab_split_variance}
\scriptsize
\setlength{\tabcolsep}{4pt}
\begin{tabular}{lll cc c cc c cc c cc c cc c cc c cc c cc}
\toprule
\multirow{2}{*}{\textbf{Encoder}} & \multirow{2}{*}{\textbf{Input}} & \multirow{2}{*}{\textbf{Fusion}} & \multicolumn{2}{c}{\textbf{Split 1*}} && \multicolumn{2}{c}{\textbf{Split 2}} && \multicolumn{2}{c}{\textbf{Split 3}} && \multicolumn{2}{c}{\textbf{Split 4}} && \multicolumn{2}{c}{\textbf{Split 5}} && 
\multicolumn{2}{c}{\textbf{Mean$\pm$SD}} \\
\cmidrule{4-5} \cmidrule{7-8} \cmidrule{10-11} \cmidrule{13-14} \cmidrule{16-17} \cmidrule{19-20}
&&& ID & CD && ID & CD && ID & CD && ID & CD && ID & CD && ID & CD \\
\midrule
\multirow{8}{*}{ResNeXt}
& DEM & - & \textbf{0.687} & 0.508 && 0.669 & 0.535 && 0.664 & 0.506 && \underbar{0.693} & 0.562 && \textbf{0.704} & 0.521 && \textbf{0.683$\pm$0.017} & 0.526$\pm$0.023 \\
& S-10 & - & 0.674 & 0.599 && 0.656 & 0.585 && 0.660 & 0.590 && 0.668 & \underbar{0.615} && 0.657 & 0.599 && 0.663$\pm$0.008 & 0.598$\pm$0.011 \\
& PrC-ms & Stacking & 0.664 & 0.614 && 0.665 & 0.611 && 0.663 & 0.578 && 0.670 & 0.607 && 0.643 & 0.597 && 0.661$\pm$0.010 & 0.601$\pm$0.015 \\
& PrC-ms & Concat.  & 0.598 & 0.500 && 0.617 & 0.515 && 0.614 & 0.506 && 0.576 & 0.469 && 0.616 & 0.503 && 0.604$\pm$0.018 & 0.499$\pm$0.017 \\
& PrC-ms & CA       & 0.599 & 0.504 && 0.619 & 0.543 && 0.635 & 0.578 && 0.649 & 0.555 && 0.635 & 0.527 && 0.627$\pm$0.019 & 0.541$\pm$0.028 \\
& PrC-ms+S-ms & Stacking & 0.674 & 0.613 && \underbar{0.671} & \underbar{0.615} && 0.666 & 0.598 && 0.681 & 0.607 && 0.672 & \textbf{0.620} && 0.673$\pm$0.005 & \underbar{0.611$\pm$0.008} \\
& PrC-ms+S-ms & Concat.  & 0.636 & 0.544 && 0.632 & 0.547 && 0.662 & 0.539 && 0.646 & 0.565 && 0.642 & 0.533 && 0.644$\pm$0.012 & 0.546$\pm$0.012 \\
& PrC-ms+S-ms & CA       & 0.632 & 0.559 && 0.625 & 0.578 && 0.656 & 0.556 && 0.648 & 0.564 && 0.639 & 0.586 && 0.640$\pm$0.012 & 0.569$\pm$0.013 \\
\midrule
\multirow{8}{*}{ViT}
& DEM & - & 0.674 & 0.517 && 0.668 & 0.541 && 0.637 & 0.515 && 0.652 & 0.504 && 0.635 & 0.514 && 0.653$\pm$0.018 & 0.518$\pm$0.014 \\
& S-5 & - & 0.646 & 0.611 && 0.638 & 0.579 && 0.653 & 0.585 && 0.652 & 0.596 && 0.658 & 0.579 && 0.649$\pm$0.008 & 0.590$\pm$0.014 \\
& S-ms & Stacking   & 0.641 & \underbar{0.617} && 0.640 & 0.576 && 0.667 & 0.560 && 0.658 & 0.586 && 0.653 & 0.595 && 0.652$\pm$0.011 & 0.587$\pm$0.021 \\
& S-ms & Concat.    & 0.656 & 0.558 && 0.654 & 0.582 && 0.669 & \textbf{0.620} && 0.670 & 0.601 && 0.645 & 0.582 && 0.659$\pm$0.011 & 0.589$\pm$0.023 \\
& S-ms & CA         & 0.624 & 0.557 && 0.656 & 0.571 && 0.621 & 0.544 && 0.635 & 0.565 && 0.620 & 0.546 && 0.631$\pm$0.015 & 0.557$\pm$0.012 \\
& EP-ms+PrC-ms+S-ms & Stacking & 0.673 & \textbf{0.631} && \underbar{0.671} & \textbf{0.619} && \underbar{0.673} & 0.601 && 0.668 & 0.608 && \underbar{0.690} & \underbar{0.610} && 0.675$\pm$0.009 & \textbf{0.614$\pm$0.012} \\
& EP-ms+PrC-ms+S-ms & Concat.  & \underbar{0.675} & 0.570 && \textbf{0.673} & 0.580 && \textbf{0.679} & \underbar{0.608} && \textbf{0.697} & \textbf{0.626} && 0.673 & 0.594 && \underbar{0.679$\pm$0.010} & 0.596$\pm$0.022 \\
& EP-ms+PrC-ms+S-ms & CA       & 0.671 & 0.559 && 0.637 & 0.556 && 0.647 & 0.558 && 0.675 & 0.570 && \underbar{0.690} & 0.572 && 0.664$\pm$0.022 & 0.563$\pm$0.007 \\
\bottomrule
\end{tabular}
\end{table*}

\subsection{Statistical Analyses}
\label{supp-c:stat_analysis}

\begin{figure}[!t]
\centering
\includegraphics[width=1\linewidth]{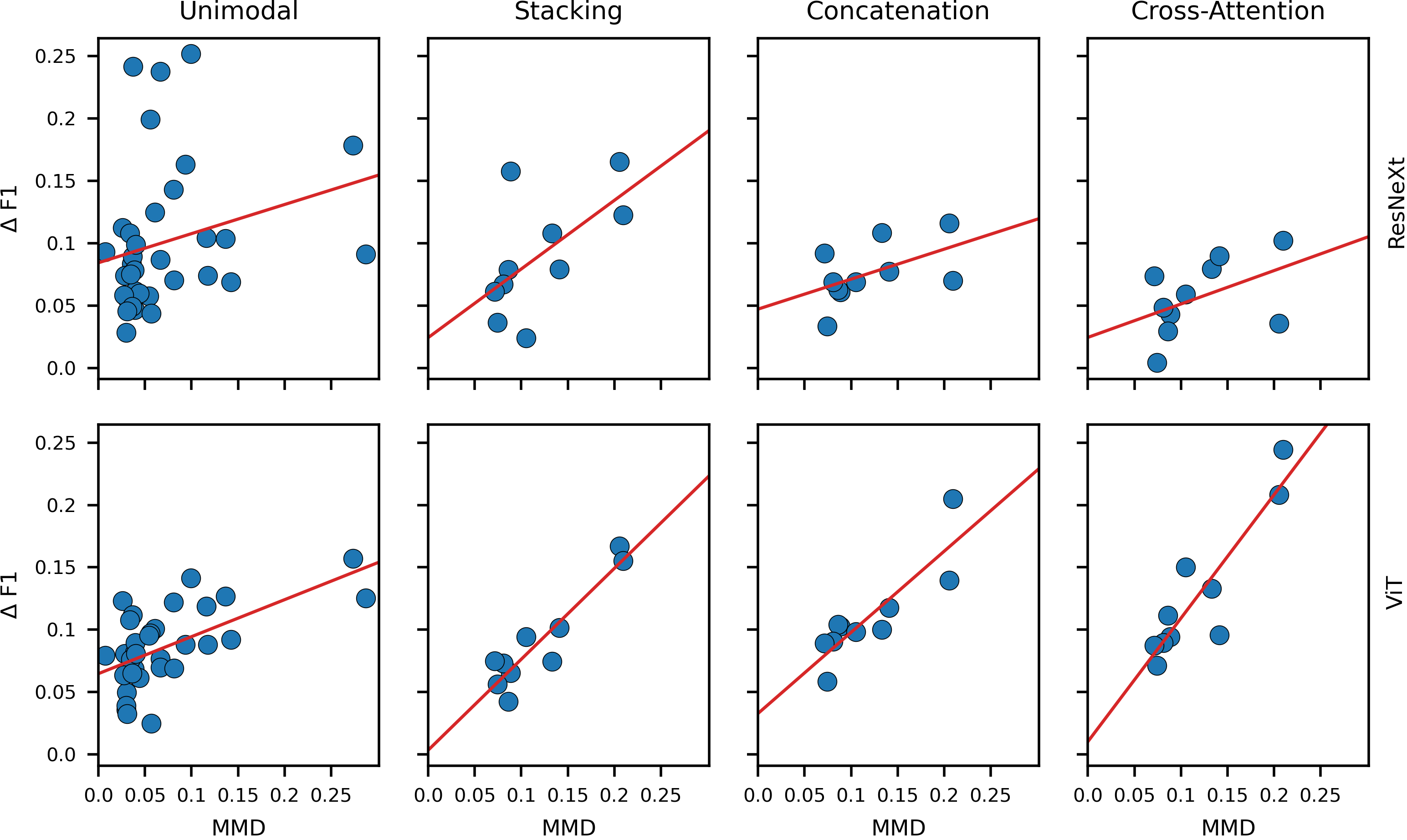}
\caption{Relationship between distribution shift (MMD) and cross-domain performance degradation ($\Delta$F1) across encoders and fusion strategies. Red lines show ordinary least-squares fits. Corresponding statistical analyses are reported in Table~\ref{supp-c:tab_mmd_delta_stats}.}
\label{supp-c:fig_mmd_delta}
\end{figure}

\begin{figure}[t]
\centering
\includegraphics[width=0.98\linewidth]{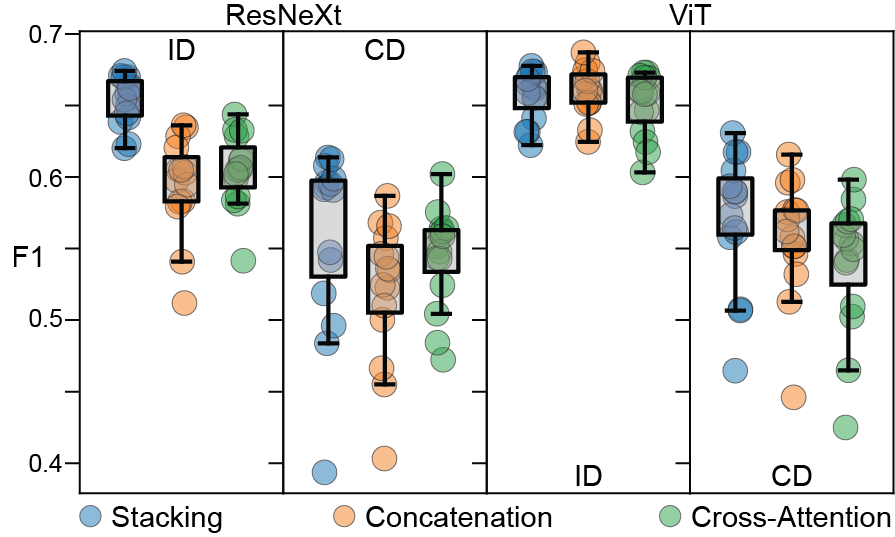}
\caption{Comparison of in-domain (ID) and cross-domain (CD) macro-F1 across stacking, concatenation, and cross-attention fusion strategies for ResNeXt and ViT encoders. Points show individual input configurations and boxplots summarize distributions.}
\label{supp-c:fig_fusion_f1_comparisons}
\end{figure}

Performance degradation was evaluated as a function of distribution shift between the in-domain and cross-domain datasets (Fig.~\ref{supp-c:fig_mmd_delta}, Table~\ref{supp-c:tab_mmd_delta_stats}). Performance degradation was quantified as $\Delta\mathrm{F1}=\mathrm{F1}_{\mathrm{ID}}-\mathrm{F1}_{\mathrm{CD}}$ and related to the corresponding MMD between the in-domain and cross-domain input distributions (Table~\ref{supp-c:tab_mmd}). Ordinary least-squares linear regression was used to quantify the linear relationship between MMD and $\Delta\mathrm{F1}$, with goodness of fit reported using the coefficient of determination ($R^2$). Pearson's correlation coefficient (PCC; $r$), Spearman's rank correlation coefficient (SRCC; $\rho$), and Kendall's rank correlation coefficient (KRCC; $\tau$) were additionally computed to evaluate linear and rank-based associations (Fig.~\ref{supp-c:fig_mmd_delta}, Table~\ref{supp-c:tab_mmd_delta_stats}). For each correlation analysis, the eight comparisons spanning the two encoders and unimodal, stacking, concatenation, and cross-attention configurations were treated as a single family for multiple-testing correction. 

Differences among fusion strategies were evaluated separately for multi-scale and multimodal input configurations using paired $t$-tests of macro-averaged F1, with Wilcoxon signed-rank tests providing a non-parametric comparison (Fig.~\ref{supp-c:fig_fusion_f1_comparisons}, Table~\ref{supp-c:tab_fusion_stats}). Tests were conducted independently for ID and CD performance, with observations matched by input configuration. Cohen's $d_z$ was used to quantify effect size. Multiple-testing families were defined by encoder and evaluation domain (ID or CD).

Within each family of hypotheses, $p$-values were adjusted using the Holm procedure to control the family-wise error rate. The Holm procedure sequentially adjusts ordered $p$-values and is uniformly more powerful than the standard Bonferroni correction while maintaining family-wise error control. Statistical significance was assessed using Holm-adjusted $p<0.05$.

\begin{table*}
\centering
\caption{Statistical association between cross-domain F1 degradation, $\Delta=\mathrm{ID}-\mathrm{CD}$, and maximum mean discrepancy (MMD) for unimodal, and multiscale and multimodal experiments, stratified by encoder and fusion method. Linear association was summarized by coefficient of determination ($R^2$) and Pearson's correlation coefficient (PCC, $r$), with monotonic association assessed using Spearman's rank correlation coefficient (SRCC, $\rho$) and Kendall's rank correlation coefficient (KRCC, $\tau$). All reported $p$-values were Holm-corrected for multiple comparisons.}
\label{supp-c:tab_mmd_delta_stats}
\scriptsize
\begin{tabular}{ll c c cc c cc c cc c cc}
\toprule
\multirow{2}{*}{Encoder} & \multirow{2}{*}{Fusion} & \multirow{2}{*}{$n$} & \multirow{2}{*}{$R^2$} & \multicolumn{2}{c}{PCC} && \multicolumn{2}{c}{SRCC} && \multicolumn{2}{c}{KRCC} \\
\cmidrule{5-6} \cmidrule{8-9} \cmidrule{11-12}
&&&& $r$ & $p_\text{Holm}$ && $\rho$ & $p_\text{Holm}$ && $\tau$ & $p_\text{Holm}$ \\
\midrule
\multirow{4}{*}{ResNeXt}
& - & 35 & 0.062 & 0.248 & 0.451 && 0.368 & 0.090 && 0.234 & 0.145 \\
& Stacking & 15 & 0.136 & 0.369 & 0.451 && 0.639 & 0.041 && 0.505 & 0.033 \\
& Concat. & 15 & 0.247 & 0.497 & 0.237 && 0.454 & 0.179 && 0.333 & 0.185 \\
& CA & 15 & 0.082 & 0.286 & 0.451 && 0.318 & 0.248 && 0.257 & 0.202 \\
\midrule
\multirow{4}{*}{ViT}
& - & 35 & 0.358 & 0.598 & 0.001 && 0.513 & 0.010 && 0.371 & 0.010 \\
& Stacking & 15 & 0.737 & 0.859 & $<$0.001 && 0.729 & 0.010 && 0.543 & 0.021 \\
& Concat. & 15 & 0.729 & 0.854 & $<$0.001 && 0.807 & 0.002 && 0.657 & 0.002 \\
& CA & 15 & 0.824 & 0.908 & $<$0.001 && 0.854 & $<$0.001 && 0.695 & 0.001 \\
\bottomrule
\end{tabular}
\end{table*}

\begin{table*}
\centering
\caption{Paired statistical comparisons of stacking, concatenation, and cross-attention (CA) fusion across matched input configurations, stratified by encoder and evaluation domain (ID, in-domain; CD, cross-domain). $A$ and $B$ denote the paired fusion methods, with their corresponding mean macro-F1 ($\overline{\text{F1}}_A$ and $\overline{\text{F1}}_B$) and number of matched pairs ($n$). Differences were assessed using paired $t$-tests ($t$) and Wilcoxon signed-rank tests ($W$), with paired-sample effect size reported as Cohen's $d_z$. All $p$-values were Holm-corrected within each encoder--domain family.}
\label{supp-c:tab_fusion_stats}
\scriptsize
\begin{tabular}{ll cc cc c cc cc c}
\toprule
\textbf{Encoder} & \textbf{Domain} & \textbf{$A$} & \textbf{$\overline{\text{F1}}_A$} & \textbf{$B$} & \textbf{$\overline{\text{F1}}_B$} & \textbf{$n$} & \textbf{$t$} & \textbf{$p_{\mathrm{Holm}}$} & \textbf{$W$} & \textbf{$p_{\mathrm{Holm}}$} & \textbf{$d_z$} \\
\midrule
\multirow{6}{*}{ResNeXt}
& \multirow{3}{*}{ID} 
& Stacking & 0.653 & Concat. & 0.595 & 15 & 8.045 & $<$0.001 & 0.000 & $<$0.001 & 2.077 \\
&& Stacking & 0.653 & CA & 0.605 & 15 & 10.209 & $<$0.001 & 0.000 & $<$0.001 & 2.636 \\
&& Concat. & 0.595 & CA & 0.605 & 15 & -1.019 & 0.326 & 56.000 & 0.847 & -0.263 \\
\cmidrule{2-12}
& \multirow{3}{*}{CD} 
& Stacking & 0.559 & Concat. & 0.521 & 15 & 2.167 & 0.144 & 22.000 & 0.077 & 0.559 \\
&& Stacking & 0.559 & CA & 0.544 & 15 & 0.939 & 0.364 & 41.000 & 0.303 & 0.242 \\
&& Concat. & 0.521 & CA & 0.544 & 15 & -1.871 & 0.165 & 21.000 & 0.077 & -0.483 \\
\midrule
\multirow{6}{*}{ViT}
& \multirow{3}{*}{ID} 
& Stacking & 0.657 & Concat. & 0.660 & 15 & -1.121 & 0.409 & 38.000 & 0.229 & -0.289 \\
&& Stacking & 0.657 & CA & 0.651 & 15 & 1.331 & 0.409 & 31.000 & 0.214 & 0.344 \\
&& Concat. & 0.660 & CA & 0.651 & 15 & 2.025 & 0.187 & 26.000 & 0.166 & 0.523 \\
\cmidrule{2-12}
& \multirow{3}{*}{CD} 
& Stacking & 0.571 & Concat. & 0.558 & 15 & 1.609 & 0.130 & 32.000 & 0.121 & 0.415 \\
&& Stacking & 0.571 & CA & 0.539 & 15 & 4.521 & 0.001 & 4.000 & 0.001 & 1.167 \\
&& Concat. & 0.558 & CA & 0.539 & 15 & 3.951 & 0.003 & 5.000 & 0.001 & 1.020 \\
\bottomrule
\end{tabular}
\end{table*}

\subsection{Foundation Model Evaluation}
\label{supp-c:foundation_models}
Five geospatial foundation models were evaluated for the benchmark, including Copernicus-FM \citep{wang2025towards}, DOFA \citep{xiong2024neural}, Panopticon \citep{waldmann2025panopticon}, Galileo \citep{tseng2025galileolearninggloballocal}, and TerraMind \citep{jakubik2025TerraMind}. For each model, EarthScape inputs were restricted to modalities compatible with the model's native input specifications, preserving each model's native modality handling. For each model, we performed a learning-rate sweep over $1\times10^{-3}$, $1\times10^{-4}$, and $1\times10^{-5}$ using the same training and validation procedure for each candidate rate. The learning rate producing the lowest validation loss was selected for the final evaluation of each foundation model. Complete results of the learning-rate sweeps are reported in Table~\ref{supp-c:tab_foundation_models}.

\begin{table}[b]
\centering
\caption{Macro-averaged F1 for in-domain (ID) and cross-domain (CD) performance of Copernicus-FM, DOFA, Galileo, Panopticon, and TerraMind foundation models across three learning rates ($\alpha\in\{10^{-3},10^{-4},10^{-5}\}$) and supported input configurations.}
\label{supp-c:tab_foundation_models}
\scriptsize
\setlength{\tabcolsep}{1pt}
\begin{tabular}{ll cc c cc c cc}
\toprule
\multirow{2}{*}{\textbf{Model}} & \multirow{2}{*}{\textbf{Input}} & \multicolumn{2}{c}{$\alpha=0.001$} && \multicolumn{2}{c}{$\alpha=0.0001$} && \multicolumn{2}{c}{$\alpha=0.00001$} \\
\cmidrule{3-4} \cmidrule{6-7} \cmidrule{9-10}
&& ID & CD && ID & CD && ID & CD \\
\midrule
Copernicus-FM & RGB+NIR & 0.558 & 0.316 && 0.592 & 0.341 && 0.591 & 0.357 \\
DOFA & RGB+NIR & 0.573 & 0.374 && 0.602 & 0.386 && 0.601 & 0.359 \\
Galileo & RGB+NIR & 0.545 & 0.260 && 0.548 & 0.286 && 0.542 & 0.249 \\
Panopticon & RGB+NIR & 0.587 & 0.408 && 0.632 & 0.507 && 0.617 & 0.461 \\
TerraMind & RGB & 0.563 & 0.394 && 0.590 & 0.414 && 0.582 & 0.376 \\ 
\midrule
Galileo & DEM+S-5 & 0.584 & 0.493 && 0.593 & 0.508 && 0.580 & 0.512 \\
Galileo & RGB+NIR+DEM+S-5 & 0.532 & 0.277 && 0.548 & 0.272 && 0.543 & 0.249 \\
TerraMind & DEM & 0.609 & 0.518 && 0.629 & 0.511 && 0.597 & 0.469 \\
TerraMind & DEM+RGB & 0.604 & 0.384 && 0.584 & 0.382 && 0.596 & 0.344 \\
\bottomrule
\end{tabular}
\end{table}

\subsection{Semantic Segmentation Evaluation}
\label{supp-c:segmentation}
We conducted additional semantic segmentation experiments to evaluate whether the principal findings from patch-level classification extend to dense pixel-level prediction. Experiments used the same spatially independent training, validation, and test partitions as the classification benchmark and included both in-domain (ID) and cross-domain (CD) evaluation. Rather than reproducing the full classification benchmark and fusion experiments, we manually selected representative input configurations spanning the major input categories, including the raw DEM as a baseline, single-scale terrain features, multi-scale terrain features, and multimodal combinations. Segmentation experiments were conducted using the same SGMap-Net channel adapter module before passing inputs to DeepLabV3+ with a ResNeXt-50 encoder \citep{chen2018encoder} and SegFormer with a MiT-B2 encoder \citep{xie2021segformer}. Models were initialized with ImageNet-pretrained encoder weights and followed the same optimization and training procedure used for the classification experiments (Supp.~\ref{supp-c:hardware_training}), including a batch size of 16, AdamW optimization with a learning rate of $3\times10^{-4}$, five-epoch linear warmup, and early stopping with a patience of five epochs. Cross-entropy loss was used in place of the focal loss used for multilabel classification. The checkpoint attaining the lowest validation loss was retained for evaluation. Performance was evaluated independently on the ID and CD test sets using intersection-over-union (IoU), Dice score, precision, and recall. Metrics were computed both globally and by geologic class to characterize overall segmentation performance and class-specific variation (Tables~\ref{supp-d:seg_iou_dice}--\ref{supp-d:seg_class_dice_cd}).


\clearpage

\begin{figure*}[!t]
\centering
\includegraphics[width=0.95\linewidth]{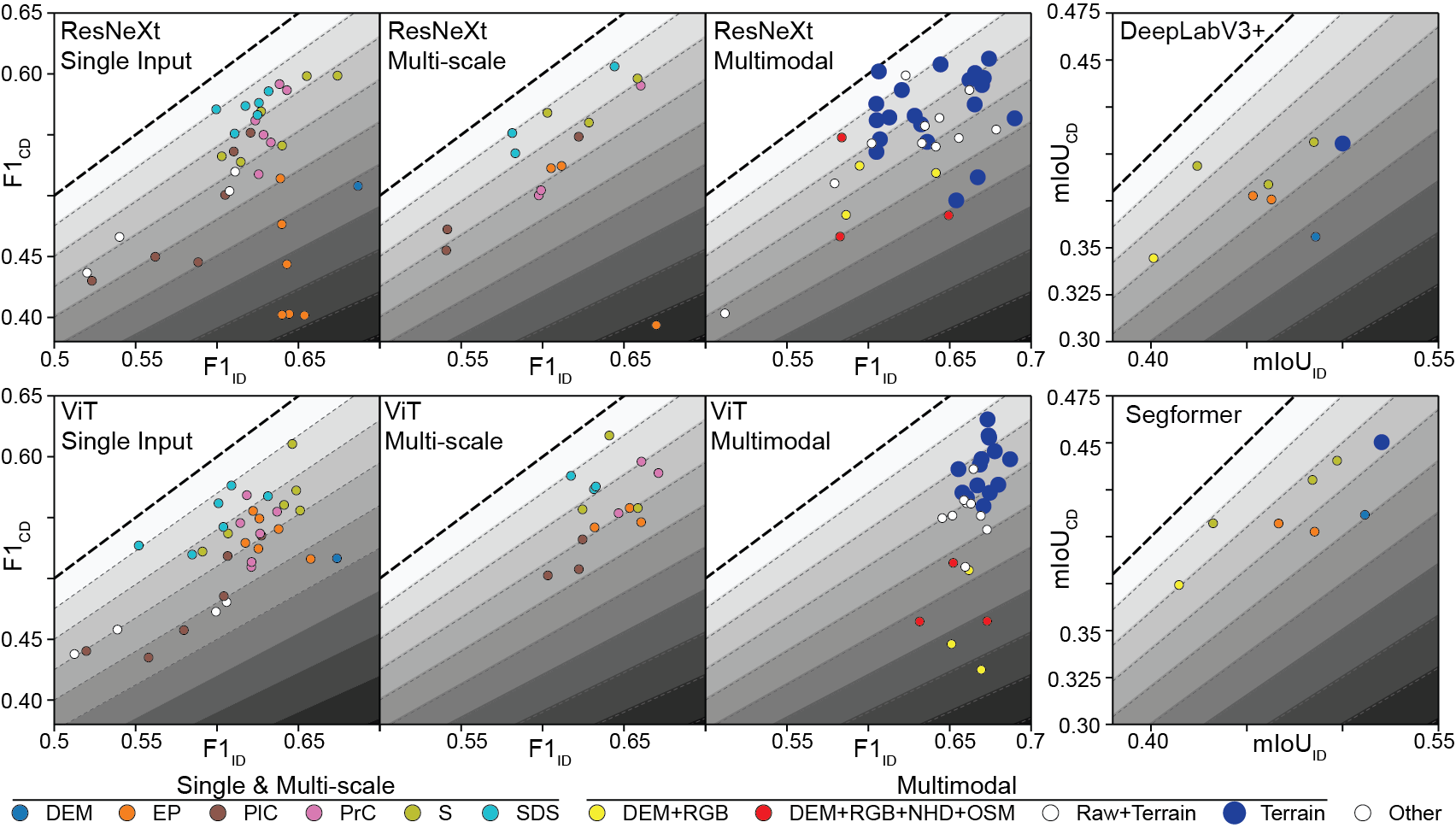}
\caption{Summary of in-domain (ID) and cross-domain (CD) performance across classification and semantic segmentation experiments. Classification results are shown for single-input, multi-scale, and multimodal configurations using ResNeXt and ViT, with segmentation results shown for DeepLabV3+ and SegFormer. The dashed black line indicates equivalent ID and CD performance, and shaded bands denote successive 5\% intervals of relative performance degradation, $(\mathrm{ID}-\mathrm{CD})/\mathrm{ID}\times100$. Point colors distinguish input families.}
\label{supp-d:fig_id_cd_summary}
\end{figure*}


\section{Comprehensive Results}
\label{supp-d}

This section provides comprehensive results for the EarthScape benchmark experiments summarized in the main text and in Figure~\ref{supp-d:fig_id_cd_summary}. Results are reported for both in-domain (ID) and cross-domain (CD) evaluation. Unimodal multilabel classification experiments are reported as global model performance in Tables~\ref{supp-d:tab_single_f1_map}--\ref{supp-d:tab_single_p_r} and class-wise performance in Tables~\ref{supp-d:tab_class_f1_id}--\ref{supp-d:tab_class_auc_cd}. Multi-scale, multilabel classification experiments are reported as global model performance in Tables~\ref{supp-d:tab_ms_f1_map}--\ref{supp-d:tab_ms_p_r} and class-wise performance in Tables~\ref{supp-d:tab_ms_class_f1_id}--\ref{supp-d:tab_ms_class_auc_cd}. Multimodal, multilabel classification experiments are reported as global model performance in Tables~\ref{supp-d:tab_mm_f1_map}--\ref{supp-d:tab_mm_p_r} and class-wise performance in Tables~\ref{supp-d:tab_mm_class_f1_id}--\ref{supp-d:tab_mm_class_auc_cd}. Semantic segmentation experiments are reported as global model performance in Tables~\ref{supp-d:seg_iou_dice}--\ref{supp-d:seg_p_r} and class-wise performance in Tables~\ref{supp-d:seg_class_iou_id}--\ref{supp-d:seg_class_dice_cd}.

\clearpage

\begin{table*}[htb]
\centering
\caption{Macro-averaged F1 and mAP for single input configurations with ResNeXt and ViT backbones. Performance was evaluated in-domain (ID) and cross-domain (CD), with cross-domain degradation computed as $\Delta=\mathrm{ID}-\mathrm{CD}$. ID$+$CD was additionally reported for F1 as a measure of combined performance across domains. Best and second-best scores in each column are indicated in \textbf{bold} and \underline{underlined}, respectively.}
\label{supp-d:tab_single_f1_map}
\scriptsize
\begin{tabular}{l cccc c cccc c ccc c ccc}
    \toprule
    \multirow{2}{*}{\textbf{Input}} & \multicolumn{4}{c}{\textbf{F1 (ResNeXt)}} && \multicolumn{4}{c}{\textbf{F1 (ViT)}} && \multicolumn{3}{c}{\textbf{mAP (ResNeXt)}}  && \multicolumn{3}{c}{\textbf{mAP (ViT)}}\\
    \cmidrule{2-5} \cmidrule{7-10} \cmidrule{12-14} \cmidrule{16-18}
     & \textbf{ID} & \textbf{CD} & \textbf{$\Delta$} & \textbf{ID$+$CD} && \textbf{ID} & \textbf{CD} & \textbf{$\Delta$} & \textbf{ID$+$CD} && \textbf{ID} & \textbf{CD} & \textbf{$\Delta$} && \textbf{ID} & \textbf{CD} & \textbf{$\Delta$} \\
    \midrule
    DEM & \textbf{0.687} & 0.508 & 0.179 & 1.195 && \textbf{0.674} & 0.517 & 0.157 & 1.191 && \textbf{0.736} & 0.565 & 0.171 && \textbf{0.720} & 0.579 & 0.141 \\
    RGB & 0.608 & 0.504 & 0.104 & 1.112 && 0.599 & 0.473 & 0.127 & 1.072 && 0.653 & 0.483 & 0.170 && 0.638 & 0.483 & 0.155 \\
    NIR & 0.611 & 0.520 & 0.091 & 1.131 && 0.606 & 0.481 & 0.125 & 1.087 && 0.647 & 0.534 & 0.113 && 0.616 & 0.464 & 0.151 \\
    NHD & 0.520 & 0.437 & 0.084 & 0.957 && 0.512 & 0.438 & 0.075 & 0.950 && 0.481 & 0.375 & 0.106 && 0.465 & 0.377 & 0.088 \\
    OSM & 0.540 & 0.466 & 0.074 & 1.006 && 0.539 & 0.458 & 0.081 & 0.997 && 0.495 & 0.411 & 0.084 && 0.488 & 0.401 & 0.087 \\

    EP-5 & 0.643 & 0.444 & 0.199 & 1.087 && 0.638 & 0.541 & 0.097 & 1.179 && 0.697 & 0.498 & 0.199 && 0.680 & 0.554 & 0.126 \\
\bottomrule
\end{tabular}
\end{table*}

\begin{table*}[h]
\centering
\caption{Macro-averaged precision and recall for single input configurations with ResNeXt and ViT backbones. Performance was evaluated in-domain (ID) and cross-domain (CD), with cross-domain degradation computed as $\Delta=\mathrm{ID}-\mathrm{CD}$. Best and second-best scores in each column are indicated in \textbf{bold} and \underline{underlined}, respectively.}
\label{supp-d:tab_single_p_r}
\scriptsize
\begin{tabular}{l ccc c ccc c ccc c ccc}
    \toprule
    \multirow{2}{*}{\textbf{Input}} & \multicolumn{3}{c}{\textbf{Precision (ResNeXt)}} && \multicolumn{3}{c}{\textbf{Precision (ViT)}} && \multicolumn{3}{c}{\textbf{Recall (ResNeXt)}}  && \multicolumn{3}{c}{\textbf{Recall (ViT)}}\\
    \cmidrule{2-4} \cmidrule{6-8} \cmidrule{10-12} \cmidrule{14-16}
     & \textbf{ID} & \textbf{CD} & \textbf{$\Delta$} && \textbf{ID} & \textbf{CD} & \textbf{$\Delta$} && \textbf{ID} & \textbf{CD} & \textbf{$\Delta$} && \textbf{ID} & \textbf{CD} & \textbf{$\Delta$} \\
    \midrule
    DEM & \underbar{0.666} & 0.474 & 0.192 && \textbf{0.639} & 0.439 & 0.200 && 0.719 & 0.585 & 0.134 && 0.740 & 0.649 & 0.090 \\
    RGB & 0.569 & 0.436 & 0.134 && 0.578 & 0.423 & 0.155 && 0.654 & 0.615 & 0.040 && 0.668 & 0.649 & 0.019 \\
    NIR & 0.566 & 0.492 & 0.074 && 0.549 & 0.467 & 0.082 && 0.683 & 0.586 & 0.097 && 0.713 & 0.510 & 0.203 \\
    NHD & 0.418 & 0.347 & 0.072 && 0.419 & 0.362 & 0.057 && 0.741 & 0.708 & 0.033 && 0.767 & 0.725 & 0.042 \\
    OSM & 0.454 & 0.378 & 0.076 && 0.449 & 0.373 & 0.076 && \textbf{0.837} & \textbf{0.829} & 0.008 && \textbf{0.841} & 0.820 & 0.021 \\

    EP-5 & 0.623 & 0.495 & 0.127 && 0.660 & 0.488 & 0.172 && 0.670 & 0.415 & 0.255 && 0.691 & 0.615 & 0.076 \\
\bottomrule
\end{tabular}
\end{table*}

\begin{table*}
\centering
\caption{Class-wise F1 for single input configurations with ResNeXt and ViT backbones under in-domain (ID) evaluation. Best and second-best scores in each column are indicated in \textbf{bold} and \underline{underlined}, respectively.}
\label{supp-d:tab_class_f1_id}
\scriptsize
\setlength{\tabcolsep}{5pt}
\begin{tabular}{lc ccccccc c ccccccc}
\toprule
\multirow{2}{*}{\textbf{Input}} & \multirow{2}{*}{\textbf{Fusion}} & \multicolumn{7}{c}{\textbf{ResNeXt}} && \multicolumn{7}{c}{\textbf{ViT}} \\
\cmidrule{3-9} \cmidrule{11-17}
&& \textbf{af1} & \textbf{Qal} & \textbf{Qaf} & \textbf{Qat} & \textbf{Qc} & \textbf{Qca} & \textbf{Qr} && \textbf{af1} & \textbf{Qal} & \textbf{Qaf} & \textbf{Qat} & \textbf{Qc} & \textbf{Qca} & \textbf{Qr} \\
\midrule
DEM & - & 0.703 & 0.813 & \underbar{0.229} & \textbf{0.457} & 0.883 & 0.750 & \textbf{0.972} && 0.659 & \underbar{0.808} & \underbar{0.250} & \textbf{0.354} & 0.901 & \textbf{0.778} & 0.966 \\
RGB & - & 0.719 & 0.790 & 0.000 & 0.262 & 0.837 & 0.680 & 0.966 && 0.695 & 0.777 & 0.088 & 0.209 & 0.813 & 0.652 & 0.961 \\
NIR & - & 0.694 & 0.778 & 0.029 & 0.303 & 0.817 & 0.693 & 0.963 && 0.681 & 0.762 & 0.080 & 0.270 & 0.831 & 0.653 & 0.964 \\
NHD & - & 0.542 & 0.762 & 0.043 & 0.209 & 0.657 & 0.464 & 0.964 && 0.541 & 0.764 & 0.036 & 0.165 & 0.655 & 0.461 & 0.964 \\
OSM & - & 0.677 & 0.773 & 0.030 & 0.111 & 0.721 & 0.502 & 0.965 && 0.671 & 0.770 & 0.033 & 0.128 & 0.711 & 0.494 & 0.966 \\
EP-5 & - & 0.722 & 0.804 & 0.041 & \underbar{0.352} & 0.872 & 0.747 & 0.963 && 0.677 & 0.795 & 0.133 & 0.326 & 0.846 & 0.725 & 0.961 \\
\bottomrule
\end{tabular}
\end{table*}

\begin{table*}
\centering
\caption{Class-wise F1 for single input configurations with ResNeXt and ViT backbones under cross-domain (CD) evaluation. Best and second-best scores in each column are indicated in \textbf{bold} and \underline{underlined}, respectively.}
\label{supp-d:tab_class_f1_cd}
\scriptsize
\setlength{\tabcolsep}{5pt}
\begin{tabular}{lc ccccccc c ccccccc}
\toprule
\multirow{2}{*}{\textbf{Input}} & \multirow{2}{*}{\textbf{Fusion}} & \multicolumn{7}{c}{\textbf{ResNeXt}} && \multicolumn{7}{c}{\textbf{ViT}} \\
\cmidrule{3-9} \cmidrule{11-17}
&& \textbf{af1} & \textbf{Qal} & \textbf{Qaf} & \textbf{Qat} & \textbf{Qc} & \textbf{Qca} & \textbf{Qr} && \textbf{af1} & \textbf{Qal} & \textbf{Qaf} & \textbf{Qat} & \textbf{Qc} & \textbf{Qca} & \textbf{Qr} \\
\midrule
DEM & - & 0.558 & 0.683 & 0.000 & 0.023 & 0.688 & 0.611 & \textbf{0.993} && 0.523 & 0.694 & 0.000 & 0.023 & 0.761 & 0.623 & \textbf{0.992} \\
RGB & - & 0.629 & 0.702 & 0.000 & 0.237 & 0.507 & 0.459 & 0.991 && 0.467 & 0.669 & 0.000 & 0.131 & 0.637 & 0.451 & 0.954 \\
NIR & - & 0.583 & 0.601 & 0.007 & \textbf{0.408} & 0.578 & 0.474 & 0.988 && 0.624 & 0.680 & 0.000 & 0.260 & 0.360 & 0.448 & \textbf{0.992} \\
NHD & - & 0.440 & 0.709 & 0.000 & 0.142 & 0.471 & 0.304 & 0.991 && 0.436 & 0.710 & 0.008 & 0.155 & 0.459 & 0.306 & \underbar{0.991} \\
OSM & - & 0.650 & 0.709 & 0.009 & 0.106 & 0.495 & 0.302 & 0.991 && 0.624 & 0.708 & 0.009 & 0.103 & 0.474 & 0.296 & \textbf{0.992} \\
EP-5 & - & 0.703 & 0.599 & 0.000 & 0.503 & 0.220 & 0.089 & \underbar{0.992} && 0.598 & 0.697 & 0.000 & 0.155 & 0.700 & 0.645 & 0.988 \\
\bottomrule
\end{tabular}
\end{table*}

\begin{table*}
\centering
\caption{Class-wise AUC for single input configurations with ResNeXt and ViT backbones under in-domain (ID) evaluation. Best and second-best scores in each column are indicated in \textbf{bold} and \underline{underlined}, respectively.}
\label{supp-d:tab_class_auc_id}
\scriptsize
\setlength{\tabcolsep}{5pt}
\begin{tabular}{lc ccccccc c ccccccc}
\toprule
\multirow{2}{*}{\textbf{Input}} & \multirow{2}{*}{\textbf{Fusion}} & \multicolumn{7}{c}{\textbf{ResNeXt}} && \multicolumn{7}{c}{\textbf{ViT}} \\
\cmidrule{3-9} \cmidrule{11-17}
&& \textbf{af1} & \textbf{Qal} & \textbf{Qaf} & \textbf{Qat} & \textbf{Qc} & \textbf{Qca} & \textbf{Qr} && \textbf{af1} & \textbf{Qal} & \textbf{Qaf} & \textbf{Qat} & \textbf{Qc} & \textbf{Qca} & \textbf{Qr} \\
\midrule
DEM & - & 0.849 & \underbar{0.839} & 0.893 & \textbf{0.925} & 0.961 & 0.914 & \textbf{0.939} && 0.800 & \textbf{0.836} & \textbf{0.930} & \textbf{0.905} & 0.972 & 0.931 & \textbf{0.936} \\
NIR & - & 0.837 & 0.742 & 0.748 & 0.824 & 0.914 & 0.869 & 0.869 && 0.831 & 0.694 & 0.743 & 0.769 & 0.910 & 0.841 & 0.819 \\
NHD & - & 0.553 & 0.664 & 0.567 & 0.784 & 0.635 & 0.625 & 0.733 && 0.504 & 0.651 & 0.587 & 0.770 & 0.610 & 0.617 & 0.710 \\
OSM & - & 0.823 & 0.561 & 0.673 & 0.635 & 0.723 & 0.661 & 0.649 && 0.806 & 0.582 & 0.733 & 0.688 & 0.712 & 0.654 & 0.594 \\
RGB & - & 0.857 & 0.763 & 0.707 & 0.828 & 0.924 & 0.870 & 0.884 && 0.846 & 0.718 & 0.806 & 0.860 & 0.911 & 0.842 & 0.845 \\
EP-5 & - & 0.857 & 0.829 & 0.773 & 0.880 & 0.950 & 0.914 & 0.917 && 0.829 & 0.804 & 0.832 & 0.890 & 0.938 & 0.907 & 0.926 \\
\bottomrule
\end{tabular}
\end{table*}

\begin{table*}
\centering
\caption{Class-wise AUC for single input configurations with ResNeXt and ViT backbones under cross-domain (CD) evaluation. Best and second-best scores in each column are indicated in \textbf{bold} and \underline{underlined}, respectively.}
\label{supp-d:tab_class_auc_cd}
\scriptsize
\setlength{\tabcolsep}{5pt}
\begin{tabular}{lc ccccccc c ccccccc}
\toprule
\multirow{2}{*}{\textbf{Input}} & \multirow{2}{*}{\textbf{Fusion}} & \multicolumn{7}{c}{\textbf{ResNeXt}} && \multicolumn{7}{c}{\textbf{ViT}} \\
\cmidrule{3-9} \cmidrule{11-17}
&& \textbf{af1} & \textbf{Qal} & \textbf{Qaf} & \textbf{Qat} & \textbf{Qc} & \textbf{Qca} & \textbf{Qr} && \textbf{af1} & \textbf{Qal} & \textbf{Qaf} & \textbf{Qat} & \textbf{Qc} & \textbf{Qca} & \textbf{Qr} \\
\midrule
DEM & - & 0.774 & 0.660 & 0.629 & 0.722 & 0.953 & 0.893 & 0.636 && 0.724 & 0.644 & 0.703 & 0.497 & 0.954 & 0.924 & 0.814 \\
NIR & - & 0.779 & 0.554 & 0.573 & 0.828 & 0.777 & 0.786 & 0.808 && 0.822 & 0.591 & 0.565 & 0.678 & 0.596 & 0.792 & 0.723 \\
NHD & - & 0.524 & 0.633 & 0.693 & 0.725 & 0.487 & 0.518 & 0.543 && 0.500 & 0.641 & 0.592 & 0.701 & 0.494 & 0.518 & 0.538 \\
OSM & - & 0.839 & 0.510 & 0.467 & 0.568 & 0.656 & 0.587 & 0.668 && 0.824 & 0.495 & 0.551 & 0.558 & 0.615 & 0.597 & 0.648 \\
RGB & - & 0.843 & 0.596 & 0.461 & 0.724 & 0.691 & 0.784 & 0.770 && 0.739 & 0.571 & 0.311 & 0.624 & 0.921 & 0.730 & 0.687 \\
EP-5 & - & 0.871 & 0.639 & 0.526 & 0.774 & 0.515 & 0.722 & 0.671 && 0.810 & 0.637 & 0.517 & 0.540 & 0.888 & 0.920 & 0.861 \\
\bottomrule
\end{tabular}
\end{table*}


\clearpage

\begin{table*}[h]
\centering
\caption{Macro-averaged F1 and mAP for multi-scale input configurations with ResNeXt and ViT backbones. Performance was evaluated in-domain (ID) and cross-domain (CD), with cross-domain degradation computed as $\Delta=\mathrm{ID}-\mathrm{CD}$. ID$+$CD was additionally reported for F1 as a measure of combined performance across domains. Best and second-best scores in each column are indicated in \textbf{bold} and \underline{underlined}, respectively.}
\label{supp-d:tab_ms_f1_map}
\scriptsize
\setlength{\tabcolsep}{5pt}
\begin{tabular}{ll cccc c cccc c ccc c ccc}
    \toprule
    \multirow{2}{*}{\textbf{Input}} & \multirow{2}{*}{\textbf{Fusion}} & \multicolumn{4}{c}{\textbf{F1 (ResNeXt)}} && \multicolumn{4}{c}{\textbf{F1 (ViT)}} && \multicolumn{3}{c}{\textbf{mAP (ResNeXt)}}  && \multicolumn{3}{c}{\textbf{mAP (ViT)}}\\
    \cmidrule{3-6} \cmidrule{8-11} \cmidrule{13-15} \cmidrule{17-19}
     && \textbf{ID} & \textbf{CD} & \textbf{$\Delta$} & \textbf{ID$+$CD} && \textbf{ID} & \textbf{CD} & \textbf{$\Delta$} & \textbf{ID$+$CD} && \textbf{ID} & \textbf{CD} & \textbf{$\Delta$} && \textbf{ID} & \textbf{CD} & \textbf{$\Delta$} \\
    \midrule
    \multirow{3}{*}{EP-ms} 
    & Stacking & \textbf{0.670} & 0.394 & 0.276 & 1.064 && 0.656 & 0.558 & 0.098 & 1.214 && \underbar{0.710} & 0.465 & 0.245 && 0.703 & 0.575 & 0.128 \\ &Concat.& 0.605 & 0.523 & 0.082 & 1.128 && 0.660 & 0.546 & 0.114 & 1.206 && 0.610 & 0.511 & 0.099 && 0.702 & 0.579 & 0.123 \\ &CA & 0.612 & 0.524 & 0.088 & 1.136 && 0.632 & 0.542 & 0.090 & 1.174 && 0.655 & 0.537 & 0.118 && 0.674 & 0.560 & 0.114 \\
    \midrule

    \multirow{3}{*}{PlC-ms} 
    & Stacking & 0.620 & 0.542 & 0.078 & 1.162 && 0.622 & 0.508 & 0.114 & 1.130 && 0.664 & 0.57 & 0.094 && 0.661 & 0.531 & 0.130 \\ &Concat.& 0.541 & 0.455 & 0.086 & 0.996 && 0.625 & 0.532 & 0.093 & 1.157 && 0.539 & 0.435 & 0.104 && 0.678 & 0.549 & 0.129 \\ &CA & 0.541 & 0.472 & 0.069 & 1.013 && 0.603 & 0.502 & 0.101 & 1.105 && 0.568 & 0.420 & 0.148 && 0.636 & 0.509 & 0.127 \\
    \midrule

    \multirow{3}{*}{PrC-ms} 
    & Stacking & \underbar{0.664} & \textbf{0.614} & 0.050 & \textbf{1.278} && \textbf{0.671} & 0.587 & 0.084 & \textbf{1.258} && 0.705 & \textbf{0.652} & 0.053 && \textbf{0.724} & 0.620 & 0.104 \\ &Concat.& 0.598 & 0.500 & 0.098 & 1.098 && \underbar{0.661} & \underbar{0.596} & 0.065 & \underbar{1.257} && 0.632 & 0.522 & 0.110 && \underbar{0.713} & \underbar{0.629} & 0.084 \\ &CA & 0.599 & 0.504 & 0.095 & 1.103 && 0.647 & 0.554 & 0.093 & 1.201 && 0.627 & 0.536 & 0.091 && 0.685 & 0.576 & 0.109 \\
    \midrule

    \multirow{3}{*}{S-ms} 
    & Stacking & 0.658 & \underbar{0.596} & 0.062 & \underbar{1.254} && 0.641 & \textbf{0.617} & \textbf{0.024} & \textbf{1.258} && \textbf{0.717} & \underbar{0.649} & 0.068 && 0.691 & \textbf{0.680} & 0.011 \\ &Concat.& 0.603 & 0.568 & \underbar{0.035} & 1.171 && 0.656 & 0.558 & 0.098 & 1.214 && 0.623 & 0.582 & \textbf{0.041} && 0.688 & 0.585 & 0.103 \\ &CA & 0.628 & 0.560 & 0.068 & 1.188 && 0.624 & 0.557 & 0.067 & 1.181 && 0.650 & 0.571 & 0.079 && 0.676 & 0.587 & 0.089 \\
    \midrule

    \multirow{3}{*}{SDS-ms} 
    & Stacking & 0.639 & 0.593 & 0.046 & 1.232 && 0.632 & 0.573 & 0.059 & 1.205 && 0.680 & 0.629 & \underbar{0.051} && 0.663 & 0.610 & \underbar{0.053} \\ &Concat.& 0.583 & 0.535 & 0.048 & 1.118 && 0.633 & 0.576 & 0.057 & 1.209 && 0.597 & 0.527 & 0.070 && 0.673 & 0.594 & 0.079 \\ &CA & 0.581 & 0.551 & \textbf{0.030} & 1.132 && 0.617 & 0.584 & \underbar{0.033} & 1.201 && 0.607 & 0.537 & 0.070 && 0.650 & 0.599 & \textbf{0.051} \\

    \bottomrule
\end{tabular}
\end{table*}

\begin{table*}[h]
\centering
\caption{Macro-averaged precision and recall for multi-scale input configurations with ResNeXt and ViT backbones. Performance was evaluated in-domain (ID) and cross-domain (CD), with cross-domain degradation computed as $\Delta=\mathrm{ID}-\mathrm{CD}$. Best and second-best scores in each column are indicated in \textbf{bold} and \underline{underlined}, respectively.}
\label{supp-d:tab_ms_p_r}
\scriptsize
\begin{tabular}{ll ccc c ccc c ccc c ccc}
    \toprule
    \multirow{2}{*}{\textbf{Input}} & \multirow{2}{*}{\textbf{Fusion}} & \multicolumn{3}{c}{\textbf{Precision (ResNeXt)}} && \multicolumn{3}{c}{\textbf{Precision (ViT)}} && \multicolumn{3}{c}{\textbf{Recall (ResNeXt)}}  && \multicolumn{3}{c}{\textbf{Recall (ViT)}}\\
    \cmidrule{3-5} \cmidrule{7-9} \cmidrule{11-13} \cmidrule{15-17}
     && \textbf{ID} & \textbf{CD} & \textbf{$\Delta$} && \textbf{ID} & \textbf{CD} & \textbf{$\Delta$} && \textbf{ID} & \textbf{CD} & \textbf{$\Delta$} && \textbf{ID} & \textbf{CD} & \textbf{$\Delta$} \\
    \midrule
    \multirow{3}{*}{EP-ms}
    & Stacking & \underbar{0.644} & 0.457 & 0.187 && 0.606 & 0.497 & 0.109 && 0.706 & 0.377 & \textbf{-0.329} && 0.650 & 0.734 & -0.084 \\ & Concat. & 0.540 & 0.472 & 0.068 && \underbar{0.629} & 0.519 & 0.110 && 0.602 & 0.711 & \underbar{-0.109} && 0.586 & 0.713 & \textbf{-0.127} \\ & CA & 0.565 & 0.496 & 0.069 && 0.591 & 0.496 & 0.095 && 0.571 & 0.677 & -0.106 && 0.607 & 0.699 & \underbar{-0.092} \\
    \midrule

    \multirow{3}{*}{PlC-ms}
    & Stacking & 0.625 & 0.486 & 0.139 && 0.574 & 0.439 & 0.135 && 0.653 & 0.632 & 0.021 && 0.632 & 0.695 & -0.063 \\ & Concat. & 0.461 & 0.370 & 0.091 && 0.584 & 0.465 & 0.119 && 0.650 & 0.680 & -0.030 && 0.642 & 0.677 & -0.035 \\ & CA & 0.476 & 0.399 & 0.077 && 0.543 & 0.456 & 0.087 && \underbar{0.755} & \underbar{0.732} & 0.023 && 0.640 & 0.693 & -0.053 \\
    \midrule

    \multirow{3}{*}{PrC-ms}
    & Stacking & 0.639 & \textbf{0.554} & 0.085 && \textbf{0.637} & \underbar{0.526} & 0.111 && 0.702 & 0.714 & -0.012 && 0.672 & 0.715 & -0.043 \\ & Concat. & 0.559 & 0.435 & 0.124 && 0.618 & \textbf{0.52}7 & 0.091 && 0.613 & 0.669 & -0.056 && 0.701 & 0.734 & -0.033 \\ & CA & 0.543 & 0.434 & 0.109 && 0.588 & 0.477 & 0.111 && 0.655 & 0.728 & -0.073 && 0.680 & \textbf{0.779} & -0.099 \\
    \midrule

    \multirow{3}{*}{S-ms}
    & Stacking & \textbf{0.656} & \underbar{0.538} & 0.118 && 0.588 & \textbf{0.527} & \underbar{0.061} && 0.717 & 0.703 & 0.014 && \textbf{0.772} & 0.754 & 0.018 \\ & Concat. & 0.549 & 0.507 & \textbf{0.042} && 0.624 & 0.498 & 0.126 && 0.696 & 0.683 & 0.013 && 0.644 & 0.721 & -0.077 \\ & CA & 0.576 & 0.506 & 0.070 && 0.587 & 0.495 & 0.092 && 0.653 & \textbf{0.751} & -0.098 && 0.653 & 0.674 & -0.021 \\
    \midrule

    \multirow{3}{*}{SDS-ms}
    & Stacking & 0.603 & 0.519 & 0.084 && 0.593 & 0.512 & 0.081 && 0.717 & 0.712 & 0.005 && 0.672 & 0.690 & -0.018 \\ & Concat. & 0.532 & 0.461 & 0.071 && 0.579 & 0.504 & 0.075 && 0.681 & 0.678 & 0.003 && 0.691 & \underbar{0.757} & -0.066 \\ & CA & 0.516 & 0.462 & \underbar{0.054} && 0.564 & 0.506 & \textbf{0.058} && \textbf{0.804} & 0.699 & 0.105 && \underbar{0.709} & 0.746 & -0.037 \\
    
    \bottomrule
\end{tabular}
\end{table*}

\clearpage

\begin{table*}
\centering
\caption{Class-wise F1 for multi-scale input configurations with ResNeXt and ViT backbones under in-domain (ID) evaluation. Best and second-best scores in each column are indicated in \textbf{bold} and \underline{underlined}, respectively.}
\label{supp-d:tab_ms_class_f1_id}
\scriptsize
\setlength{\tabcolsep}{5pt}
\begin{tabular}{lc ccccccc c ccccccc}
\toprule
\multirow{2}{*}{\textbf{Input}} & \multirow{2}{*}{\textbf{Fusion}} & \multicolumn{7}{c}{\textbf{ResNeXt}} && \multicolumn{7}{c}{\textbf{ViT}} \\
\cmidrule{3-9} \cmidrule{11-17}
&& \textbf{af1} & \textbf{Qal} & \textbf{Qaf} & \textbf{Qat} & \textbf{Qc} & \textbf{Qca} & \textbf{Qr} && \textbf{af1} & \textbf{Qal} & \textbf{Qaf} & \textbf{Qat} & \textbf{Qc} & \textbf{Qca} & \textbf{Qr} \\
\midrule
\multirow{3}{*}{EP-ms} 
& Stacking & \textbf{0.724} & \textbf{0.810} & \underbar{0.150} & \textbf{0.420} & 0.869 & 0.751 & 0.966 && 0.696 & 0.807 & 0.122 & \textbf{0.380} & 0.859 & 0.759 & 0.967 \\
& Concat. & 0.707 & 0.783 & 0.078 & 0.265 & 0.761 & 0.677 & 0.966 && \textbf{0.740} & \underbar{0.818} & 0.151 & 0.351 & 0.851 & 0.745 & \underbar{0.968} \\
& CA & 0.700 & 0.779 & 0.133 & 0.243 & 0.759 & 0.703 & 0.965 && 0.700 & 0.805 & \underbar{0.160} & 0.218 & 0.847 & 0.729 & 0.965 \\
\midrule
\multirow{3}{*}{PlC-ms} 
& Stacking & 0.666 & 0.799 & 0.071 & 0.276 & 0.831 & 0.732 & 0.966 && 0.643 & 0.799 & 0.107 & 0.295 & 0.824 & 0.723 & 0.965 \\
& Concat. & 0.507 & 0.771 & 0.000 & 0.205 & 0.711 & 0.626 & 0.966 && 0.631 & 0.806 & 0.114 & 0.301 & 0.836 & 0.726 & 0.958 \\
& CA & 0.561 & 0.764 & 0.069 & 0.105 & 0.717 & 0.608 & 0.965 && 0.602 & 0.798 & 0.043 & 0.305 & 0.816 & 0.692 & 0.967 \\
\midrule
\multirow{3}{*}{PrC-ms} 
& Stacking & \underbar{0.710} & \underbar{0.802} & 0.108 & \underbar{0.355} & 0.886 & \textbf{0.816} & \textbf{0.973} && 0.695 & 0.801 & \textbf{0.258} & 0.293 & 0.886 & 0.793 & 0.972 \\
& Concat. & 0.613 & 0.761 & 0.040 & 0.216 & 0.845 & 0.739 & 0.968 && 0.684 & \textbf{0.819} & 0.081 & \underbar{0.360} & 0.898 & \textbf{0.819} & 0.964 \\
& CA & 0.575 & 0.775 & 0.060 & 0.311 & 0.843 & 0.672 & 0.958 && 0.663 & 0.811 & 0.082 & 0.321 & 0.882 & \underbar{0.800} & 0.966 \\
\midrule
\multirow{3}{*}{S-ms} 
& Stacking & 0.703 & 0.792 & \textbf{0.200} & 0.261 & \textbf{0.912} & \underbar{0.774} & 0.965 && 0.700 & 0.786 & 0.087 & 0.283 & 0.904 & 0.760 & 0.966 \\
& Concat. & 0.609 & 0.767 & 0.091 & 0.184 & 0.890 & 0.711 & \underbar{0.969} && 0.686 & 0.795 & 0.123 & 0.313 & \textbf{0.919} & 0.789 & \textbf{0.969} \\
& CA & 0.602 & 0.774 & 0.110 & 0.333 & 0.893 & 0.719 & 0.967 && 0.704 & 0.773 & 0.000 & 0.241 & \underbar{0.917} & 0.767 & \textbf{0.969} \\
\midrule
\multirow{3}{*}{SDS-ms} 
& Stacking & \underbar{0.710} & 0.775 & 0.074 & 0.312 & \underbar{0.898} & 0.733 & 0.967 && 0.696 & 0.774 & 0.067 & 0.327 & 0.889 & 0.704 & 0.965 \\
& Concat. & 0.683 & 0.758 & 0.108 & 0.176 & 0.817 & 0.576 & 0.965 && 0.702 & 0.770 & 0.076 & 0.291 & 0.882 & 0.738 & \textbf{0.969} \\
& CA & 0.587 & 0.767 & 0.053 & 0.193 & 0.849 & 0.657 & 0.965 && \underbar{0.707} & 0.767 & 0.051 & 0.242 & 0.880 & 0.708 & 0.966 \\
\bottomrule
\end{tabular}
\end{table*}

\begin{table*}
\centering
\caption{Class-wise F1 for multi-scale input configurations with ResNeXt and ViT backbones under cross-domain (CD) evaluation. Best and second-best scores in each column are indicated in \textbf{bold} and \underline{underlined}, respectively.}
\label{supp-d:tab_ms_class_f1_cd}
\scriptsize
\setlength{\tabcolsep}{5pt}
\begin{tabular}{lc ccccccc c ccccccc}
\toprule
\multirow{2}{*}{\textbf{Input}} & \multirow{2}{*}{\textbf{Fusion}} & \multicolumn{7}{c}{\textbf{ResNeXt}} && \multicolumn{7}{c}{\textbf{ViT}} \\
\cmidrule{3-9} \cmidrule{11-17}
&& \textbf{af1} & \textbf{Qal} & \textbf{Qaf} & \textbf{Qat} & \textbf{Qc} & \textbf{Qca} & \textbf{Qr} && \textbf{af1} & \textbf{Qal} & \textbf{Qaf} & \textbf{Qat} & \textbf{Qc} & \textbf{Qca} & \textbf{Qr} \\
\midrule
\multirow{3}{*}{EP-ms} 
& Stacking & 0.623 & 0.692 & 0.000 & 0.301 & 0.114 & 0.033 & \underbar{0.991} && 0.647 & 0.702 & 0.000 & 0.286 & 0.644 & 0.635 & \underbar{0.991} \\
& Concat. & 0.672 & \underbar{0.739} & 0.000 & 0.343 & 0.563 & 0.348 & \textbf{0.992} && \textbf{0.706} & 0.709 & 0.000 & 0.159 & 0.618 & 0.640 & \textbf{0.992} \\
& CA & 0.699 & \textbf{0.758} & 0.000 & \underbar{0.360} & 0.327 & 0.536 & \underbar{0.991} && 0.665 & 0.703 & 0.000 & 0.120 & 0.689 & 0.625 & \underbar{0.991} \\
\midrule
\multirow{3}{*}{PlC-ms} 
& Stacking & 0.626 & 0.710 & 0.000 & 0.209 & 0.674 & 0.584 & \underbar{0.991} && 0.562 & 0.702 & 0.000 & 0.115 & 0.624 & 0.561 & 0.989 \\
& Concat. & 0.433 & 0.707 & 0.000 & 0.057 & 0.531 & 0.466 & \textbf{0.992} && 0.500 & 0.685 & 0.000 & \underbar{0.314} & 0.636 & 0.598 & 0.991 \\
& CA & 0.453 & 0.709 & 0.038 & 0.168 & 0.473 & 0.474 & \underbar{0.991} && 0.484 & 0.708 & 0.000 & 0.239 & 0.568 & 0.525 & \textbf{0.992} \\
\midrule
\multirow{3}{*}{PrC-ms} 
& Stacking & \textbf{0.729} & 0.729 & 0.000 & \textbf{0.398} & 0.783 & \underbar{0.666} & \underbar{0.991} && 0.647 & \underbar{0.718} & 0.000 & 0.159 & 0.825 & \textbf{0.765} & \textbf{0.992} \\
& Concat. & 0.546 & 0.717 & 0.000 & 0.000 & 0.630 & 0.620 & 0.989 && 0.637 & \textbf{0.727} & 0.000 & 0.279 & 0.837 & 0.704 & 0.987 \\
& CA & 0.553 & 0.694 & 0.000 & 0.145 & 0.719 & 0.435 & 0.986 && 0.590 & 0.707 & 0.000 & 0.082 & 0.786 & 0.720 & \underbar{0.991} \\
\midrule
\multirow{3}{*}{S-ms} 
& Stacking & \underbar{0.705} & 0.715 & 0.000 & 0.314 & \textbf{0.836} & 0.620 & 0.983 && 0.629 & 0.687 & \textbf{0.190} & \textbf{0.338} & 0.818 & 0.670 & 0.989 \\
& Concat. & 0.507 & 0.711 & \textbf{0.211} & 0.118 & 0.779 & 0.660 & \textbf{0.992} && 0.610 & 0.712 & 0.000 & 0.000 & \textbf{0.856} & \underbar{0.734} & \textbf{0.992} \\
& CA & 0.536 & 0.726 & 0.000 & 0.194 & 0.781 & \textbf{0.693} & \underbar{0.991} && 0.606 & 0.678 & 0.000 & 0.077 & \underbar{0.853} & 0.691 & \underbar{0.991} \\
\midrule
\multirow{3}{*}{SDS-ms} 
& Stacking & 0.685 & 0.695 & 0.000 & 0.344 & \underbar{0.784} & 0.649 & \textbf{0.992} && 0.682 & 0.699 & 0.000 & 0.214 & 0.780 & 0.648 & 0.990 \\
& Concat. & 0.648 & 0.697 & 0.000 & 0.292 & 0.697 & 0.420 & \underbar{0.991} && \underbar{0.695} & 0.707 & 0.000 & 0.202 & 0.824 & 0.611 & 0.989 \\
& CA & 0.533 & 0.706 & \underbar{0.055} & 0.267 & 0.762 & 0.546 & \textbf{0.992} && 0.653 & 0.700 & 0.000 & 0.305 & 0.803 & 0.639 & 0.990 \\
\bottomrule
\end{tabular}
\end{table*}

\clearpage

\begin{table*}
\centering
\caption{Class-wise AUC for multi-scale input configurations with ResNeXt and ViT backbones under in-domain (ID) evaluation. Best and second-best scores in each column are indicated in \textbf{bold} and \underline{underlined}, respectively.}
\label{supp-d:tab_ms_class_auc_id}
\scriptsize
\setlength{\tabcolsep}{5pt}
\begin{tabular}{lc ccccccc c ccccccc}
\toprule
\multirow{2}{*}{\textbf{Input}} & \multirow{2}{*}{\textbf{Fusion}} & \multicolumn{7}{c}{\textbf{ResNeXt}} && \multicolumn{7}{c}{\textbf{ViT}} \\
\cmidrule{3-9} \cmidrule{11-17}
&& \textbf{af1} & \textbf{Qal} & \textbf{Qaf} & \textbf{Qat} & \textbf{Qc} & \textbf{Qca} & \textbf{Qr} && \textbf{af1} & \textbf{Qal} & \textbf{Qaf} & \textbf{Qat} & \textbf{Qc} & \textbf{Qca} & \textbf{Qr} \\
\midrule
\multirow{3}{*}{EP-ms} 
& Stacking & \textbf{0.865} & \textbf{0.840} & 0.772 & \underbar{0.894} & 0.951 & 0.917 & 0.923 && 0.841 & 0.842 & 0.834 & \textbf{0.916} & 0.944 & 0.925 & 0.915 \\
& Concat. & \underbar{0.859} & 0.754 & 0.717 & 0.755 & 0.838 & 0.856 & 0.778 && \textbf{0.869} & \textbf{0.852} & 0.854 & 0.899 & 0.942 & 0.922 & 0.919 \\
& CA & 0.843 & 0.782 & 0.872 & 0.846 & 0.837 & 0.894 & 0.853 && 0.830 & 0.826 & \underbar{0.881} & 0.857 & 0.931 & 0.914 & 0.882 \\
\midrule
\multirow{3}{*}{PlC-ms} 
& Stacking & 0.806 & 0.810 & 0.691 & 0.880 & 0.914 & 0.896 & 0.902 && 0.797 & 0.804 & 0.825 & 0.902 & 0.915 & 0.907 & 0.900 \\
& Concat. & 0.545 & 0.669 & 0.722 & 0.797 & 0.799 & 0.818 & 0.780 && 0.775 & 0.816 & 0.866 & 0.884 & 0.926 & 0.912 & 0.900 \\
& CA & 0.636 & 0.650 & 0.801 & 0.646 & 0.781 & 0.826 & 0.596 && 0.750 & 0.787 & 0.832 & 0.886 & 0.906 & 0.885 & 0.876 \\
\midrule
\multirow{3}{*}{PrC-ms} 
& Stacking & 0.857 & \underbar{0.824} & 0.815 & \textbf{0.903} & 0.962 & \textbf{0.949} & \textbf{0.933} && 0.838 & 0.822 & \textbf{0.901} & 0.909 & 0.964 & 0.943 & \underbar{0.939} \\
& Concat. & 0.766 & 0.718 & 0.617 & 0.766 & 0.927 & 0.907 & 0.789 && 0.838 & \underbar{0.845} & 0.859 & 0.915 & 0.967 & \textbf{0.956} & \textbf{0.941} \\
& CA & 0.721 & 0.742 & 0.817 & 0.866 & 0.927 & 0.878 & 0.879 && 0.805 & 0.827 & 0.866 & 0.896 & 0.961 & \underbar{0.953} & 0.932 \\
\midrule
\multirow{3}{*}{S-ms} 
& Stacking & 0.853 & 0.819 & 0.857 & 0.872 & \textbf{0.974} & \underbar{0.935} & \underbar{0.929} && 0.832 & 0.808 & 0.861 & 0.881 & 0.976 & 0.938 & 0.889 \\
& Concat. & 0.735 & 0.715 & \underbar{0.881} & 0.741 & 0.963 & 0.878 & 0.812 && 0.836 & 0.801 & 0.779 & 0.865 & \textbf{0.978} & 0.949 & 0.915 \\
& CA & 0.753 & 0.734 & \textbf{0.886} & 0.833 & \underbar{0.965} & 0.902 & 0.856 && 0.843 & 0.769 & \underbar{0.881} & 0.799 & \underbar{0.977} & 0.932 & 0.904 \\
\midrule
\multirow{3}{*}{SDS-ms} 
& Stacking & 0.858 & 0.771 & 0.828 & 0.843 & \underbar{0.965} & 0.899 & 0.889 && 0.847 & 0.731 & 0.858 & 0.842 & 0.962 & 0.893 & 0.844 \\
& Concat. & 0.819 & 0.680 & 0.743 & 0.742 & 0.907 & 0.777 & 0.722 && \underbar{0.850} & 0.727 & 0.854 & 0.850 & 0.964 & 0.904 & 0.864 \\
& CA & 0.730 & 0.669 & 0.747 & 0.720 & 0.932 & 0.851 & 0.665 && 0.844 & 0.701 & 0.827 & 0.801 & 0.958 & 0.880 & 0.837 \\

\bottomrule
\end{tabular}
\end{table*}

\begin{table*}
\centering
\caption{Class-wise AUC for multi-scale input configurations with ResNeXt and ViT backbones under cross-domain (CD) evaluation. Best and second-best scores in each column are indicated in \textbf{bold} and \underline{underlined}, respectively.}
\label{supp-d:tab_ms_class_auc_cd}
\scriptsize
\setlength{\tabcolsep}{5pt}
\begin{tabular}{lc ccccccc c ccccccc}
\toprule
\multirow{2}{*}{\textbf{Input}} & \multirow{2}{*}{\textbf{Fusion}} & \multicolumn{7}{c}{\textbf{ResNeXt}} && \multicolumn{7}{c}{\textbf{ViT}} \\
\cmidrule{3-9} \cmidrule{11-17}
&& \textbf{af1} & \textbf{Qal} & \textbf{Qaf} & \textbf{Qat} & \textbf{Qc} & \textbf{Qca} & \textbf{Qr} && \textbf{af1} & \textbf{Qal} & \textbf{Qaf} & \textbf{Qat} & \textbf{Qc} & \textbf{Qca} & \textbf{Qr} \\
\midrule
\multirow{3}{*}{EP-ms} 
& Stacking & 0.819 & \underbar{0.741} & 0.503 & 0.773 & 0.504 & 0.606 & 0.613 && 0.852 & 0.693 & 0.720 & 0.745 & 0.845 & 0.918 & 0.821 \\
& Concat. & 0.859 & 0.736 & 0.586 & 0.763 & 0.735 & 0.816 & 0.832 && \underbar{0.874} & \textbf{0.722} & 0.709 & 0.719 & 0.826 & 0.920 & 0.859 \\
& CA & 0.860 & \textbf{0.805} & 0.775 & 0.777 & 0.646 & 0.868 & 0.816 && 0.853 & 0.631 & 0.651 & 0.570 & 0.861 & 0.917 & 0.884 \\
\midrule
\multirow{3}{*}{PlC-ms} 
& Stacking & 0.813 & 0.711 & 0.709 & 0.694 & 0.878 & 0.901 & 0.829 && 0.768 & 0.653 & 0.783 & 0.608 & 0.844 & 0.889 & 0.832 \\
& Concat. & 0.521 & 0.662 & 0.760 & 0.627 & 0.762 & 0.802 & 0.691 && 0.693 & 0.655 & 0.831 & \underbar{0.773} & 0.864 & 0.897 & 0.814 \\
& CA & 0.594 & 0.555 & \textbf{0.881} & 0.708 & 0.615 & 0.821 & 0.622 && 0.684 & 0.561 & \underbar{0.841} & 0.727 & 0.806 & 0.856 & 0.775 \\
\midrule
\multirow{3}{*}{PrC-ms} 
& Stacking & \textbf{0.887} & 0.738 & 0.835 & 0.844 & 0.951 & \textbf{0.948} & \textbf{0.897} && 0.848 & 0.708 & \textbf{0.875} & 0.665 & 0.971 & \textbf{0.960} & \underbar{0.886} \\
& Concat. & 0.750 & 0.684 & 0.279 & 0.216 & 0.844 & 0.907 & 0.629 && 0.845 & \underbar{0.719} & 0.683 & \textbf{0.778} & 0.969 & \underbar{0.956} & \textbf{0.909} \\
& CA & 0.774 & 0.631 & 0.664 & 0.658 & 0.909 & 0.856 & \textbf{0.897} && 0.791 & 0.626 & 0.643 & 0.551 & 0.962 & 0.950 & 0.860 \\
\midrule
\multirow{3}{*}{S-ms} 
& Stacking & 0.864 & 0.723 & 0.592 & 0.777 & \underbar{0.962} & \underbar{0.926} & 0.780 && 0.850 & 0.650 & 0.716 & 0.719 & 0.974 & 0.954 & 0.869 \\
& Concat. & 0.696 & 0.689 & 0.710 & 0.525 & \textbf{0.965} & 0.917 & 0.811 && 0.791 & 0.654 & 0.589 & 0.517 & \textbf{0.982} & \textbf{0.960} & 0.892 \\
& CA & 0.729 & 0.699 & 0.772 & 0.695 & 0.952 & 0.924 & \underbar{0.893} && 0.814 & 0.602 & 0.674 & 0.516 & \underbar{0.981} & 0.954 & 0.872 \\
\midrule
\multirow{3}{*}{SDS-ms} 
& Stacking & \underbar{0.880} & 0.690 & 0.630 & \textbf{0.846} & 0.938 & 0.897 & 0.869 && 0.868 & 0.669 & 0.572 & 0.727 & 0.968 & 0.918 & 0.874 \\
& Concat. & 0.829 & 0.617 & 0.769 & \underbar{0.845} & 0.882 & 0.741 & 0.545 && \textbf{0.875} & 0.681 & 0.540 & 0.658 & 0.966 & 0.908 & 0.876 \\
& CA & 0.733 & 0.532 & \underbar{0.845} & 0.832 & 0.940 & 0.871 & 0.441 && 0.847 & 0.629 & 0.514 & 0.709 & 0.966 & 0.909 & 0.844 \\
\bottomrule
\end{tabular}
\end{table*}


\begin{table*}[h]
\centering
\caption{Macro-averaged F1 and mAP for multimodal input configurations with ResNeXt and ViT backbones. Performance was evaluated in-domain (ID) and cross-domain (CD), with cross-domain degradation computed as $\Delta=\mathrm{ID}-\mathrm{CD}$. ID$+$CD was additionally reported for F1 as a measure of combined performance across domains. Best and second-best scores in each column are indicated in \textbf{bold} and \underline{underlined}, respectively.}
\label{supp-d:tab_mm_f1_map}
\scriptsize
\setlength{\tabcolsep}{5pt}
\begin{tabular}{ll cccc c cccc c ccc c ccc}
    \toprule
    \multirow{2}{*}{\textbf{Input}} & \multirow{2}{*}{\textbf{Fusion}} & \multicolumn{4}{c}{\textbf{F1 (ResNeXt)}} && \multicolumn{4}{c}{\textbf{F1 (ViT)}} && \multicolumn{3}{c}{\textbf{mAP (ResNeXt)}}  && \multicolumn{3}{c}{\textbf{mAP (ViT)}}\\
    \cmidrule{3-6} \cmidrule{8-11} \cmidrule{13-15} \cmidrule{17-19}
     && \textbf{ID} & \textbf{CD} & \textbf{$\Delta$} & \textbf{ID$+$CD} && \textbf{ID} & \textbf{CD} & \textbf{$\Delta$} & \textbf{ID$+$CD} && \textbf{ID} & \textbf{CD} & \textbf{$\Delta$} && \textbf{ID} & \textbf{CD} & \textbf{$\Delta$} \\
    \midrule
    \multirow{3}{1.65cm}{DEM+RGB} & Stacking & 0.642 & 0.519 & 0.123 & 1.160 && 0.662 & 0.507 & 0.155 & 1.169 && 0.666 & 0.516 & 0.150 && 0.686 & 0.539 & 0.147 \\ & Concat. & 0.595 & 0.525 & 0.070 & 1.119 && 0.651 & 0.446 & 0.205 & 1.097 && 0.602 & 0.534 & 0.068 && 0.707 & 0.513 & 0.194 \\ & CA & 0.586 & 0.484 & 0.102 & 1.071 && 0.669 & 0.425 & 0.244 & 1.094 && 0.621 & 0.545 & 0.076 && 0.711 & 0.505 & 0.205 \\
    \midrule

    \multirow{3}{1.65cm}{DEM+NHD+\newline OSM+RGB} & Stacking & 0.649 & 0.484 & 0.166 & 1.133 && 0.632 & 0.465 & 0.167 & 1.096 && 0.682 & 0.501 & 0.181 && 0.688 & 0.514 & 0.174 \\ & Concat. & 0.583 & 0.466 & 0.116 & 1.049 && 0.652 & 0.513 & 0.139 & 1.165 && 0.586 & 0.543 & 0.044 && 0.698 & 0.546 & 0.152 \\ & CA & 0.584 & 0.548 & 0.036 & 1.131 && 0.673 & 0.465 & 0.208 & 1.138 && 0.603 & 0.565 & 0.038 && 0.697 & 0.523 & 0.174 \\
\bottomrule
\end{tabular}
\end{table*}

\begin{table*}[h]
\centering
\caption{Macro-averaged precision and recall for multimodal input configurations with ResNeXt and ViT backbones. Performance was evaluated in-domain (ID) and cross-domain (CD), with cross-domain degradation computed as $\Delta=\mathrm{ID}-\mathrm{CD}$. Best and second-best scores in each column are indicated in \textbf{bold} and \underline{underlined}, respectively.}
\label{supp-d:tab_mm_p_r}
\scriptsize
\setlength{\tabcolsep}{5pt}
\begin{tabular}{ll ccc c ccc c ccc c ccc}
    \toprule
    \multirow{2}{*}{\textbf{Input}} & \multirow{2}{*}{\textbf{Fusion}} & \multicolumn{3}{c}{\textbf{Precision (ResNeXt)}} && \multicolumn{3}{c}{\textbf{Precision (ViT)}} && \multicolumn{3}{c}{\textbf{Recall (ResNeXt)}}  && \multicolumn{3}{c}{\textbf{Recall (ViT)}}\\
    \cmidrule{3-5} \cmidrule{7-9} \cmidrule{11-13} \cmidrule{15-17}
     && \textbf{ID} & \textbf{CD} & \textbf{$\Delta$} && \textbf{ID} & \textbf{CD} & \textbf{$\Delta$} && \textbf{ID} & \textbf{CD} & \textbf{$\Delta$} && \textbf{ID} & \textbf{CD} & \textbf{$\Delta$} \\
    \midrule
    \multirow{3}{*}{DEM+RGB} & Stacking & 0.598 & 0.452 & 0.146 && 0.606 & 0.476 & 0.129 && \underbar{0.772} & 0.628 & 0.143 && 0.761 & 0.548 & 0.213 \\ & Concat. & 0.562 & 0.528 & 0.034 && 0.626 & 0.504 & 0.122 && 0.634 & 0.537 & 0.097 && 0.684 & 0.446 & 0.237 \\ & CA & 0.546 & 0.488 & 0.058 && 0.628 & 0.441 & 0.187 && 0.640 & 0.623 & 0.017 && 0.738 & 0.503 & 0.235 \\
    \midrule

    \multirow{3}{*}{DEM+NHD+OSM+RGB} & Stacking & 0.613 & 0.448 & 0.165 && 0.614 & 0.439 & 0.175 && 0.718 & 0.537 & 0.181 && 0.654 & 0.540 & 0.115 \\ & Concat. & 0.546 & 0.396 & 0.150 && 0.626 & 0.448 & 0.178 && 0.640 & 0.693 & \underbar{-0.053} && 0.699 & 0.618 & 0.080 \\ & CA & 0.541 & 0.496 & 0.044 && 0.634 & 0.474 & 0.161 && 0.672 & 0.678 & -0.006 && 0.743 & 0.512 & 0.231 \\
\bottomrule
\end{tabular}
\end{table*}

\begin{table*}
\centering
\caption{Class-wise F1 for multimodal input configurations with ResNeXt and ViT backbones under in-domain (ID) evaluation. Best and second-best scores in each column are indicated in \textbf{bold} and \underline{underlined}, respectively.}
\label{supp-d:tab_mm_class_f1_id}
\scriptsize
\setlength{\tabcolsep}{4pt}
\begin{tabular}{lc ccccccc c ccccccc}
\toprule
\multirow{2}{*}{\textbf{Input}} & \multirow{2}{*}{\textbf{Fusion}} & \multicolumn{7}{c}{\textbf{ResNeXt}} && \multicolumn{7}{c}{\textbf{ViT}} \\
\cmidrule{3-9} \cmidrule{11-17}
&& \textbf{af1} & \textbf{Qal} & \textbf{Qaf} & \textbf{Qat} & \textbf{Qc} & \textbf{Qca} & \textbf{Qr} && \textbf{af1} & \textbf{Qal} & \textbf{Qaf} & \textbf{Qat} & \textbf{Qc} & \textbf{Qca} & \textbf{Qr} \\
\midrule
\multirow{3}{*}{DEM+RGB} & Stacking & 0.714 & 0.779 & 0.087 & \textbf{0.413} & 0.823 & 0.707 & 0.968 && 0.708 & 0.795 & \textbf{0.242} & 0.310 & 0.858 & 0.753 & 0.966 \\
& Concat. & 0.668 & 0.771 & 0.000 & 0.276 & 0.797 & 0.685 & 0.964 && 0.712 & 0.801 & 0.000 & \underbar{0.403} & 0.884 & 0.787 & 0.971 \\
& CA & 0.662 & 0.778 & 0.000 & 0.219 & 0.819 & 0.662 & 0.965 && 0.714 & 0.806 & 0.136 & \textbf{0.434} & 0.881 & 0.746 & 0.969 \\
\midrule
\multirow{3}{*}{DEM+NHD+OSM+RGB} & Stacking & 0.715 & 0.771 & 0.163 & 0.351 & 0.859 & 0.719 & 0.967 && 0.675 & 0.798 & 0.108 & 0.240 & 0.868 & 0.766 & 0.965 \\
& Concat. & 0.677 & 0.771 & 0.089 & 0.200 & 0.797 & 0.579 & 0.966 && 0.723 & 0.797 & 0.140 & 0.288 & 0.877 & 0.769 & 0.971 \\
& CA & 0.656 & 0.775 & 0.114 & 0.200 & 0.773 & 0.601 & 0.967 && 0.699 & 0.789 & \underbar{0.235} & 0.387 & 0.874 & 0.760 & 0.968 \\
\bottomrule
\end{tabular}
\end{table*}

\begin{table*}
\centering
\caption{Class-wise F1 for multimodal input configurations with ResNeXt and ViT backbones under cross-domain (CD) evaluation. Best and second-best scores in each column are indicated in \textbf{bold} and \underline{underlined}, respectively.}
\label{supp-d:tab_mm_class_f1_cd}
\scriptsize
\setlength{\tabcolsep}{4pt}
\begin{tabular}{lc ccccccc c ccccccc}
\toprule
\multirow{2}{*}{\textbf{Input}} & \multirow{2}{*}{\textbf{Fusion}} & \multicolumn{7}{c}{\textbf{ResNeXt}} && \multicolumn{7}{c}{\textbf{ViT}} \\
\cmidrule{3-9} \cmidrule{11-17}
&& \textbf{af1} & \textbf{Qal} & \textbf{Qaf} & \textbf{Qat} & \textbf{Qc} & \textbf{Qca} & \textbf{Qr} && \textbf{af1} & \textbf{Qal} & \textbf{Qaf} & \textbf{Qat} & \textbf{Qc} & \textbf{Qca} & \textbf{Qr} \\
\midrule
\multirow{3}{*}{DEM+RGB} & Stacking & 0.584 & 0.692 & 0.000 & 0.181 & 0.652 & 0.531 & \underbar{0.992} && 0.514 & 0.598 & 0.000 & 0.000 & 0.771 & 0.672 & \underbar{0.992} \\
& Concat. & 0.649 & 0.696 & 0.000 & 0.287 & 0.585 & 0.464 & 0.991 && 0.418 & 0.323 & 0.000 & 0.024 & 0.705 & 0.659 & \underbar{0.992} \\
& CA & 0.517 & 0.606 & 0.000 & 0.327 & 0.530 & 0.417 & \underbar{0.992} && 0.458 & 0.169 & 0.000 & 0.000 & 0.790 & 0.565 & \underbar{0.992} \\
\midrule
\multirow{3}{*}{DEM+NHD+OSM+RGB} & Stacking & 0.584 & 0.685 & 0.000 & 0.027 & 0.641 & 0.460 & 0.988 && 0.371 & 0.646 & 0.000 & 0.000 & 0.622 & 0.621 & \underbar{0.992} \\
& Concat. & 0.547 & 0.710 & 0.000 & 0.176 & 0.479 & 0.361 & 0.991 && 0.601 & 0.628 & 0.000 & 0.030 & 0.711 & 0.629 & \underbar{0.992} \\
& CA & 0.561 & 0.709 & 0.000 & 0.419 & 0.588 & 0.566 & 0.991 && 0.624 & 0.681 & 0.000 & 0.000 & 0.744 & 0.214 & \underbar{0.992} \\
\bottomrule
\end{tabular}
\end{table*}

\clearpage

\begin{table*}
\centering
\caption{Class-wise AUC for multimodal input configurations with ResNeXt and ViT backbones under in-domain (ID) evaluation. Best and second-best scores in each column are indicated in \textbf{bold} and \underline{underlined}, respectively.}
\label{supp-d:tab_mm_class_auc_id}
\scriptsize
\setlength{\tabcolsep}{4pt}
\begin{tabular}{lc ccccccc c ccccccc}
\toprule
\multirow{2}{*}{\textbf{Input}} & \multirow{2}{*}{\textbf{Fusion}} & \multicolumn{7}{c}{\textbf{ResNeXt}} && \multicolumn{7}{c}{\textbf{ViT}} \\
\cmidrule{3-9} \cmidrule{11-17}
&& \textbf{af1} & \textbf{Qal} & \textbf{Qaf} & \textbf{Qat} & \textbf{Qc} & \textbf{Qca} & \textbf{Qr} && \textbf{af1} & \textbf{Qal} & \textbf{Qaf} & \textbf{Qat} & \textbf{Qc} & \textbf{Qca} & \textbf{Qr} \\
\midrule
\multirow{3}{*}{DEM+RGB} & Stacking & 0.850 & 0.752 & 0.861 & 0.888 & 0.913 & 0.882 & 0.918 && 0.849 & 0.786 & 0.907 & 0.896 & 0.939 & 0.920 & 0.931 \\
& Concat. & 0.785 & 0.702 & 0.533 & 0.704 & 0.881 & 0.865 & 0.726 && 0.852 & 0.832 & 0.868 & 0.919 & 0.963 & 0.939 & 0.942 \\
& CA & 0.803 & 0.700 & 0.776 & 0.861 & 0.909 & 0.835 & 0.867 && 0.854 & 0.822 & \underbar{0.925} & 0.919 & 0.966 & 0.919 & 0.939 \\
\midrule
\multirow{3}{*}{DEM+NHD+OSM+RGB} & Stacking & 0.855 & 0.759 & 0.832 & 0.886 & 0.934 & 0.895 & 0.924 && 0.828 & 0.800 & \textbf{0.933} & 0.891 & 0.952 & 0.921 & 0.931 \\
& Concat. & 0.813 & 0.644 & 0.737 & 0.781 & 0.887 & 0.802 & 0.818 && 0.864 & 0.820 & 0.866 & 0.909 & 0.960 & 0.929 & \underbar{0.947} \\
& CA & 0.803 & 0.693 & 0.740 & 0.797 & 0.889 & 0.806 & 0.842 && 0.850 & 0.802 & 0.887 & 0.921 & 0.958 & 0.925 & 0.930 \\
\bottomrule
\end{tabular}
\end{table*}

\begin{table*}
\centering
\caption{Class-wise AUC for multimodal input configurations with ResNeXt and ViT backbones under cross-domain (CD) evaluation. Best and second-best scores in each column are indicated in \textbf{bold} and \underline{underlined}, respectively.}
\label{supp-d:tab_mm_class_auc_cd}
\scriptsize
\setlength{\tabcolsep}{4pt}
\begin{tabular}{lc ccccccc c ccccccc}
\toprule
\multirow{2}{*}{\textbf{Input}} & \multirow{2}{*}{\textbf{Fusion}} & \multicolumn{7}{c}{\textbf{ResNeXt}} && \multicolumn{7}{c}{\textbf{ViT}} \\
\cmidrule{3-9} \cmidrule{11-17}
&& \textbf{af1} & \textbf{Qal} & \textbf{Qaf} & \textbf{Qat} & \textbf{Qc} & \textbf{Qca} & \textbf{Qr} && \textbf{af1} & \textbf{Qal} & \textbf{Qaf} & \textbf{Qat} & \textbf{Qc} & \textbf{Qca} & \textbf{Qr} \\
\midrule
\multirow{3}{*}{DEM+RGB} & Stacking & 0.803 & 0.580 & 0.418 & 0.615 & 0.874 & 0.850 & 0.720 && 0.742 & 0.546 & 0.400 & 0.357 & 0.953 & 0.922 & 0.722 \\
& Concat. & 0.828 & 0.697 & 0.248 & 0.728 & 0.847 & 0.832 & 0.814 && 0.659 & 0.557 & 0.524 & 0.425 & 0.889 & 0.922 & 0.746 \\
& CA & 0.733 & 0.538 & 0.560 & 0.781 & 0.888 & 0.857 & 0.692 && 0.634 & 0.534 & 0.786 & 0.436 & 0.941 & 0.886 & 0.792 \\
\midrule
\multirow{3}{*}{DEM+NHD+OSM+RGB} & Stacking & 0.772 & 0.603 & 0.712 & 0.541 & 0.857 & 0.775 & 0.617 && 0.624 & 0.486 & 0.334 & 0.549 & 0.905 & 0.900 & 0.818 \\
& Concat. & 0.776 & 0.633 & \textbf{0.779} & 0.752 & 0.866 & 0.897 & 0.653 && 0.803 & 0.599 & 0.502 & 0.492 & 0.916 & 0.903 & 0.692 \\
& CA & 0.777 & 0.601 & 0.501 & 0.705 & 0.924 & 0.838 & 0.583 && 0.819 & 0.607 & 0.604 & 0.552 & 0.913 & 0.821 & 0.730 \\
\bottomrule
\end{tabular}
\end{table*}

\clearpage

\begin{table*}[htb]
\centering
\caption{Mean IoU (mIoU) and macro-averaged Dice scores for semantic segmentation with DeepLabV3+ and Segformer architectures. Performance was evaluated in-domain (ID) and cross-domain (CD), with cross-domain degradation computed as $\Delta=\mathrm{ID}-\mathrm{CD}$. Best and second-best scores in each column are indicated in \textbf{bold} and \underline{underlined}, respectively.}
\label{supp-d:seg_iou_dice}
\scriptsize
\begin{tabular}{l ccc c ccc c ccc c ccc}
\toprule
\multirow{2}{*}{\textbf{Input}} & \multicolumn{3}{c}{\textbf{mIoU (DeepLabV3+)}} && \multicolumn{3}{c}{\textbf{mIoU (Segformer)}} && \multicolumn{3}{c}{\textbf{Dice (DeepLabV3+)}}  && \multicolumn{3}{c}{\textbf{Dice (Segformer)}}\\
\cmidrule{2-4} \cmidrule{6-8} \cmidrule{10-12} \cmidrule{14-16}
& \textbf{ID} & \textbf{CD} & \textbf{$\Delta$} && \textbf{ID} & \textbf{CD} & \textbf{$\Delta$} && \textbf{ID} & \textbf{CD} & \textbf{$\Delta$} && \textbf{ID} & \textbf{CD} & \textbf{$\Delta$} \\
\midrule
DEM & \underbar{0.486} & 0.356 & 0.130 && \underbar{0.512} & 0.412 & 0.100 && \underbar{0.548} & 0.415 & 0.133 && \underbar{0.576} & 0.474 & 0.101 \\
EP-5 & 0.448 & 0.378 & 0.070 && 0.460 & 0.400 & 0.059 && 0.513 & 0.423 & 0.089 && 0.525 & 0.447 & 0.079 \\
EP-51 & 0.453 & 0.378 & 0.075 && 0.485 & 0.403 & 0.083 && 0.521 & 0.426 & 0.095 && 0.552 & 0.451 & 0.102 \\
S-5 & 0.461 & 0.384 & 0.078 && 0.484 & 0.430 & 0.054 && 0.522 & 0.442 & 0.080 && 0.546 & 0.491 & 0.055 \\
S-50 & 0.424 & \underbar{0.394} & \textbf{0.031} && 0.432 & 0.407 & \textbf{0.025} && 0.482 & 0.442 & \textbf{0.040} && 0.488 & 0.455 & \textbf{0.034} \\
EP-ms & 0.463 & 0.376 & 0.087 && 0.467 & 0.407 & 0.060 && 0.529 & 0.426 & 0.103 && 0.534 & 0.453 & 0.081 \\
S-ms & 0.485 & 0.\textbf{406} & 0.079 && 0.497 & \underbar{0.441} & 0.057 && 0.550 & \underbar{0.467} & 0.083 && 0.560 & \underbar{0.502} & 0.058 \\
DEM+RGB & 0.401 & 0.344 & \underbar{0.057} && 0.415 & 0.374 & \underbar{0.040} && 0.459 & 0.393 & \underbar{0.066} && 0.476 & 0.433 & \underbar{0.043} \\
EP-ms+PrC-ms+S-ms & \textbf{0.500} & \textbf{0.406} & 0.094 && \textbf{0.520} & \textbf{0.450} & 0.070 && \textbf{0.566} & \textbf{0.468} & 0.098 && \textbf{0.587} & \textbf{0.513} & 0.073 \\
\bottomrule
\end{tabular}
\end{table*}

\begin{table*}[htb]
\centering
\caption{Macro-averaged precision and recall for semantic segmentation with DeepLabV3+ and Segformer architectures. Performance was evaluated in-domain (ID) and cross-domain (CD), with cross-domain degradation computed as $\Delta=\mathrm{ID}-\mathrm{CD}$. Best and second-best scores in each column are indicated in \textbf{bold} and \underline{underlined}, respectively.}
\label{supp-d:seg_p_r}
\scriptsize
\begin{tabular}{l ccc c ccc c ccc c ccc}
\toprule
\multirow{2}{*}{\textbf{Input}} & \multicolumn{3}{c}{\textbf{Precision (DeepLabV3+)}} && \multicolumn{3}{c}{\textbf{Precision (Segformer)}} && \multicolumn{3}{c}{\textbf{Recall (DeepLabV3+)}}  && \multicolumn{3}{c}{\textbf{Recall (Segformer)}}\\
\cmidrule{2-4} \cmidrule{6-8} \cmidrule{10-12} \cmidrule{14-16}
& \textbf{ID} & \textbf{CD} & \textbf{$\Delta$} && \textbf{ID} & \textbf{CD} & \textbf{$\Delta$} && \textbf{ID} & \textbf{CD} & \textbf{$\Delta$} && \textbf{ID} & \textbf{CD} & \textbf{$\Delta$} \\
\midrule
DEM & \underbar{0.616} & 0.498 & 0.118 && \underbar{0.632} & 0.535 & 0.096 && 0.551 & 0.441 & 0.110 && \underbar{0.585} & 0.496 & 0.089 \\
EP-5 & 0.582 & 0.486 & 0.096 && 0.594 & 0.494 & 0.099 && 0.528 & 0.461 & 0.066 && 0.536 & 0.484 & 0.052 \\
EP-51 & 0.581 & 0.471 & 0.111 && 0.616 & 0.497 & 0.119 && 0.533 & 0.465 & 0.068 && 0.561 & 0.490 & 0.071 \\
S-5 & 0.592 & 0.508 & \underbar{0.084} && 0.607 & 0.541 & 0.066 && 0.529 & 0.454 & 0.075 && 0.555 & 0.511 & \underbar{0.044} \\
S-50 & 0.532 & 0.490 & 0.042 && 0.530 & 0.500 & \underbar{0.030} && 0.505 & 0.461 & \textbf{0.044} && 0.512 & 0.474 & \textbf{0.037} \\
EP-ms & 0.586 & 0.458 & 0.128 && 0.601 & 0.491 & 0.111 && 0.543 & 0.473 & 0.070 && 0.544 & 0.493 & 0.051 \\
S-ms & 0.609 & \underbar{0.524} & 0.085 && 0.624 & \underbar{0.553} & 0.071 && \underbar{0.561} & \underbar{0.486} & 0.075 && 0.570 & \underbar{0.516} & 0.054 \\
DEM+RGB & 0.520 & 0.453 & \textbf{0.068} && 0.533 & 0.509 & \textbf{0.024} && 0.482 & 0.425 & \underbar{0.057} && 0.496 & 0.458 & \textbf{0.037} \\
EP-ms+PrC-ms+S-ms & \textbf{0.631} & \textbf{0.529} & 0.102 && \textbf{0.654} & \textbf{0.578} & 0.076 && \textbf{0.572} & \textbf{0.489} & 0.083 && \textbf{0.594} & \textbf{0.533} & 0.061 \\
\bottomrule
\end{tabular}
\end{table*}

\clearpage

\begin{table*}
\centering
\caption{Class-wise IoU for semantic segmentation with DeepLabV3+ and Segformer architectures under in-domain (ID) evaluation. Best and second-best scores in each column are indicated in \textbf{bold} and \underline{underlined}, respectively.}
\label{supp-d:seg_class_iou_id}
\scriptsize
\setlength{\tabcolsep}{5pt}
\begin{tabular}{l ccccccc c ccccccc}
\toprule
\multirow{2}{*}{\textbf{Input}} & \multicolumn{7}{c}{\textbf{DeepLabV3+}} && \multicolumn{7}{c}{\textbf{Segformer}} \\
\cmidrule{2-8} \cmidrule{10-16}
& \textbf{af1} & \textbf{Qal} & \textbf{Qaf} & \textbf{Qat} & \textbf{Qc} & \textbf{Qca} & \textbf{Qr} && \textbf{af1} & \textbf{Qal} & \textbf{Qaf} & \textbf{Qat} & \textbf{Qc} & \textbf{Qca} & \textbf{Qr} \\
\midrule
DEM & 0.240 & 0.343 & 0.000 & 0.057 & 0.563 & 0.316 & \underbar{0.750} && \textbf{0.296} & \textbf{0.407} & 0.000 & \textbf{0.076} & 0.590 & \underbar{0.339} & 0.766 \\
EP-5 & 0.239 & 0.328 & 0.000 & 0.017 & 0.469 & 0.209 & 0.720 && 0.266 & 0.356 & 0.000 & 0.055 & 0.494 & 0.277 & 0.727 \\
EP-51 & \underbar{0.262} & \underbar{0.367} & 0.000 & \textbf{0.078} & 0.456 & 0.257 & 0.728 && \underbar{0.286} & 0.384 & 0.000 & 0.058 & 0.514 & 0.280 & 0.750 \\
S-5 & 0.227 & 0.303 & 0.000 & 0.048 & \underbar{0.580} & 0.296 & 0.732 && 0.256 & 0.314 & 0.000 & 0.055 & \textbf{0.606} & 0.319 & 0.761 \\
S-50 & 0.045 & 0.243 & 0.000 & 0.043 & 0.547 & 0.299 & 0.711 && 0.037 & 0.249 & 0.000 & 0.052 & 0.555 & 0.323 & 0.717 \\
EP-ms & 0.246 & \textbf{0.381} & 0.000 & \underbar{0.074} & 0.497 & 0.259 & 0.722 && 0.262 & 0.371 & 0.000 & 0.048 & 0.498 & 0.267 & 0.733 \\
S-ms & 0.222 & 0.338 & 0.000 & 0.058 & \textbf{0.584} & \underbar{0.317} & 0.753 && 0.265 & 0.346 & 0.000 & 0.052 & \underbar{0.599} & 0.332 & \underbar{0.767} \\
DEM+RGB & 0.203 & 0.144 & 0.000 & 0.000 & 0.476 & 0.218 & 0.709 && 0.234 & 0.177 & 0.000 & 0.005 & 0.489 & 0.232 & 0.720 \\
EP-ms+PrC-ms+S-ms & \textbf{0.264} & \textbf{0.381} & 0.000 & 0.068 & 0.551 & \textbf{0.326} & \textbf{0.774} && 0.276 & \underbar{0.405} & 0.000 & \underbar{0.073} & 0.577 & \textbf{0.369} & \textbf{0.776} \\
\bottomrule
\end{tabular}
\end{table*}

\begin{table*}
\centering
\caption{Class-wise IoU for semantic segmentation with DeepLabV3+ and Segformer architectures under cross-domain (CD) evaluation. Best and second-best scores in each column are indicated in \textbf{bold} and \underline{underlined}, respectively.}
\label{supp-d:seg_class_iou_cd}
\scriptsize
\setlength{\tabcolsep}{5pt}
\begin{tabular}{l ccccccc c ccccccc}
\toprule
\multirow{2}{*}{\textbf{Input}} & \multicolumn{7}{c}{\textbf{DeepLabV3+}} && \multicolumn{7}{c}{\textbf{Segformer}} \\
\cmidrule{2-8} \cmidrule{10-16}
& \textbf{af1} & \textbf{Qal} & \textbf{Qaf} & \textbf{Qat} & \textbf{Qc} & \textbf{Qca} & \textbf{Qr} && \textbf{af1} & \textbf{Qal} & \textbf{Qaf} & \textbf{Qat} & \textbf{Qc} & \textbf{Qca} & \textbf{Qr} \\
\midrule
DEM & 0.126 & 0.189 & 0.000 & 0.000 & 0.264 & 0.099 & 0.714 && \textbf{0.201} & 0.286 & 0.000 & 0.000 & 0.338 & 0.123 & 0.755 \\
EP-5 & 0.105 & 0.260 & 0.000 & 0.027 & 0.023 & 0.007 & \underbar{0.774} && 0.147 & 0.306 & 0.000 & \underbar{0.078} & 0.022 & 0.001 & 0.783 \\
EP-51 & \textbf{0.179} & 0.283 & 0.000 & 0.037 & 0.017 & 0.008 & 0.763 && 0.174 & 0.307 & 0.000 & 0.066 & 0.016 & 0.017 & 0.780 \\
S-5 & 0.138 & 0.231 & 0.000 & \textbf{0.044} & \underbar{0.365} & 0.126 & 0.749 && 0.173 & 0.285 & 0.000 & 0.067 & 0.405 & \underbar{0.167} & 0.783 \\
S-50 & 0.006 & 0.174 & 0.000 & 0.013 & \textbf{0.376} & 0.138 & 0.766 && 0.006 & 0.185 & 0.000 & 0.013 & \underbar{0.374} & 0.162 & 0.776 \\
EP-ms & 0.112 & \textbf{0.315} & 0.000 & \underbar{0.042} & 0.030 & 0.008 & 0.757 && 0.112 & \textbf{0.326} & 0.000 & \textbf{0.084} & 0.003 & 0.000 & 0.785 \\
S-ms & 0.149 & 0.233 & 0.000 & 0.020 & 0.359 & \underbar{0.158} & 0.773 && \underbar{0.185} & 0.265 & 0.000 & 0.054 & \textbf{0.412} & 0.180 & \underbar{0.804} \\
DEM+RGB & 0.067 & 0.049 & 0.000 & 0.000 & 0.302 & 0.068 & 0.749 && 0.151 & 0.114 & 0.000 & 0.000 & 0.287 & 0.165 & 0.758 \\
EP-ms+PrC-ms+S-ms & \underbar{0.150} & \underbar{0.311} & 0.000 & 0.040 & 0.256 & \textbf{0.167} & \textbf{0.784} && \underbar{0.185} & \underbar{0.313} & 0.000 & 0.046 & 0.310 & \textbf{0.206} & \textbf{0.808} \\
\bottomrule
\end{tabular}
\end{table*}

\clearpage

\begin{table*}
\centering
\caption{Class-wise Dice scores for semantic segmentation with DeepLabV3+ and Segformer architectures under in-domain (ID) evaluation. Best and second-best scores in each column are indicated in \textbf{bold} and \underline{underlined}, respectively.}
\label{supp-d:seg_class_dice_id}
\scriptsize
\setlength{\tabcolsep}{5pt}
\begin{tabular}{l ccccccc c ccccccc}
\toprule
\multirow{2}{*}{\textbf{Input}} & \multicolumn{7}{c}{\textbf{DeepLabV3+}} && \multicolumn{7}{c}{\textbf{Segformer}} \\
\cmidrule{2-8} \cmidrule{10-16}
& \textbf{af1} & \textbf{Qal} & \textbf{Qaf} & \textbf{Qat} & \textbf{Qc} & \textbf{Qca} & \textbf{Qr} && \textbf{af1} & \textbf{Qal} & \textbf{Qaf} & \textbf{Qat} & \textbf{Qc} & \textbf{Qca} & \textbf{Qr} \\
\midrule
DEM & 0.295 & 0.415 & 0.000 & 0.079 & 0.630 & \underbar{0.392} & 0.802 && \textbf{0.357} & \textbf{0.484} & 0.000 & \textbf{0.103} & 0.657 & \underbar{0.414} & 0.819 \\
EP-5 & 0.294 & 0.399 & 0.000 & 0.026 & 0.544 & 0.272 & 0.782 && 0.329 & 0.428 & 0.000 & 0.074 & 0.568 & 0.349 & 0.786 \\
EP-51 & \underbar{0.324} & 0.442 & 0.000 & \textbf{0.106} & 0.532 & 0.331 & 0.790 && 0.349 & 0.459 & 0.000 & 0.083 & 0.588 & 0.356 & 0.808 \\
S-5 & 0.279 & 0.370 & 0.000 & 0.071 & \underbar{0.646} & 0.366 & 0.788 && 0.314 & 0.383 & 0.000 & 0.075 & \textbf{0.675} & 0.392 & 0.814 \\
S-50 & 0.058 & 0.301 & 0.000 & 0.061 & 0.628 & 0.373 & 0.771 && 0.049 & 0.303 & 0.000 & 0.070 & 0.633 & 0.399 & 0.776 \\
EP-ms & 0.305 & \underbar{0.455} & 0.000 & \underbar{0.101} & 0.571 & 0.332 & 0.783 && 0.322 & 0.446 & 0.000 & 0.067 & 0.575 & 0.339 & 0.794 \\
S-ms & 0.277 & 0.414 & 0.000 & 0.077 & \textbf{0.654} & \underbar{0.392} & \underbar{0.809} && 0.326 & 0.419 & 0.000 & 0.070 & \underbar{0.667} & 0.407 & \underbar{0.820} \\
DEM+RGB & 0.252 & 0.189 & 0.000 & 0.000 & 0.550 & 0.288 & 0.770 && 0.293 & 0.232 & 0.000 & 0.007 & 0.564 & 0.303 & 0.779 \\
EP-ms+PrC-ms+S-ms & \textbf{0.328} & \textbf{0.458} & 0.000 & 0.093 & 0.623 & \textbf{0.405} & \textbf{0.828} && \underbar{0.340} & \underbar{0.483} & 0.000 & \underbar{0.099} & 0.648 & \textbf{0.450} & \textbf{0.829} \\
\bottomrule
\end{tabular}
\end{table*}

\begin{table*}
\centering
\caption{Class-wise Dice scores for semantic segmentation with DeepLabV3+ and Segformer architectures under cross-domain (CD) evaluation. Best and second-best scores in each column are indicated in \textbf{bold} and \underline{underlined}, respectively.}
\label{supp-d:seg_class_dice_cd}
\scriptsize
\setlength{\tabcolsep}{5pt}
\begin{tabular}{l ccccccc c ccccccc}
\toprule
\multirow{2}{*}{\textbf{Input}} & \multicolumn{7}{c}{\textbf{DeepLabV3+}} && \multicolumn{7}{c}{\textbf{Segformer}} \\
\cmidrule{2-8} \cmidrule{10-16}
& \textbf{af1} & \textbf{Qal} & \textbf{Qaf} & \textbf{Qat} & \textbf{Qc} & \textbf{Qca} & \textbf{Qr} && \textbf{af1} & \textbf{Qal} & \textbf{Qaf} & \textbf{Qat} & \textbf{Qc} & \textbf{Qca} & \textbf{Qr} \\
\midrule
DEM & 0.174 & 0.248 & 0.000 & 0.000 & 0.332 & 0.143 & 0.779 && \textbf{0.265} & 0.352 & 0.000 & 0.000 & 0.408 & 0.169 & 0.819 \\
EP-5 & 0.148 & 0.322 & 0.000 & 0.041 & 0.034 & 0.010 & 0.831 && 0.198 & 0.371 & 0.000 & \underbar{0.109} & 0.031 & 0.002 & 0.837 \\
EP-51 & \textbf{0.243} & 0.347 & 0.000 & 0.055 & 0.025 & 0.013 & 0.821 && 0.235 & 0.372 & 0.000 & 0.089 & 0.023 & 0.025 & 0.835 \\
S-5 & 0.192 & 0.287 & 0.000 & \textbf{0.066} & \underbar{0.435} & 0.168 & 0.813 && 0.235 & 0.348 & 0.000 & 0.088 & \underbar{0.480} & 0.225 & 0.841 \\
S-50 & 0.010 & 0.214 & 0.000 & 0.022 & \textbf{0.460} & 0.188 & 0.824 && 0.010 & 0.224 & 0.000 & 0.020 & 0.457 & 0.217 & 0.833 \\
EP-ms & 0.154 & \textbf{0.387} & 0.000 & \underbar{0.062} & 0.040 & 0.013 & 0.820 && 0.150 & \textbf{0.394} & 0.000 & \textbf{0.120} & 0.006 & 0.000 & 0.839 \\
S-ms & \underbar{0.206} & 0.291 & 0.000 & 0.031 & 0.428 & \underbar{0.212} & \underbar{0.839} && \underbar{0.249} & 0.325 & 0.000 & 0.074 & \textbf{0.486} & \underbar{0.239} & \underbar{0.863} \\
DEM+RGB & 0.092 & 0.075 & 0.000 & 0.000 & 0.386 & 0.099 & 0.816 && 0.207 & 0.163 & 0.000 & 0.000 & 0.359 & 0.228 & 0.822 \\
EP-ms+PrC-ms+S-ms & 0.203 & \underbar{0.377} & 0.000 & 0.055 & 0.340 & \textbf{0.227} & \textbf{0.849} && 0.247 & \underbar{0.377} & 0.000 & 0.066 & 0.395 & \textbf{0.278} & \textbf{0.865} \\
\bottomrule
\end{tabular}
\end{table*}

\clearpage

\end{document}